\documentclass[12pt]{article}
\usepackage[utf8]{inputenc}
\usepackage{cite}
\usepackage{amsfonts}
\usepackage{amsmath}
\usepackage{amsthm}
\usepackage{enumitem}
\usepackage{amssymb}
\usepackage{caption}
\usepackage{graphicx}
\usepackage{mathrsfs}
\usepackage{multirow}
\usepackage{xcolor}
\usepackage{setspace}
\usepackage{booktabs}
\usepackage{latexsym, mathdots, array, extarrows}
\usepackage{algpseudocode,float}
\usepackage{lipsum}
\usepackage{pifont}
\usepackage{diagbox}

\usepackage{amsmath,amsfonts,amsthm,amssymb}
\usepackage{bm,dsfont}
\usepackage{booktabs}
\usepackage{graphicx}
\usepackage{array}
\usepackage{xcolor}
\usepackage{sidecap}
\usepackage{listings}
\usepackage{times}
\usepackage{url}
\usepackage{hyperref}
\usepackage{subfigure}
\usepackage{makecell} 

\usepackage{algorithm}      % For the algorithm float environment
\usepackage{algpseudocode}  % For the algorithmic pseudocode syntax

\usepackage{cleveref}

\usepackage{titletoc}

\usepackage{geometry}
\newtheorem{theorem}{Theorem}
\newtheorem{proposition}{Proposition}
\newtheorem{lemma}{Lemma}

\newcommand{\R}{\mathbb{R}}
\newcommand{\E}{\mathbb{E}}
\newcommand{\diag}{\mathrm{diag}}

\allowdisplaybreaks[4]

\title{On the Global Convergence of Stochastic Adam for Physics-Informed Neural Networks
}

\author{Yihang Gao\thanks{Department of Mathematics, Hong Kong Baptist University. Email: gaoyihang@hkbu.edu.hk}}

\date{}

\begin{document}

\maketitle

\begin{abstract}
The Adam optimizer is widely used to train neural networks with highly nonlinear and nonconvex objective functions. Remarkably, Adam often drives the empirical training loss nearly to zero. This phenomenon has been partially explained by neural tangent kernel analyses of gradient descent (GD) and stochastic gradient descent (SGD). However, analyzing Adam presents additional challenges. Its first-moment momentum couples stochastic gradients across iterations, while its adaptive second-moment preconditioner depends on the same stochastic gradient used in the update. In this paper, we establish linear convergence guarantees for Adam in training sufficiently wide two-layer neural networks with either full-batch or minibatch gradients. For full-batch Adam, we construct a Lyapunov function to control the effect of historical momentum. For minibatch Adam, our analysis builds on a virtual-iterate representation, a conditional analysis of the dependence between stochastic gradients and adaptive preconditioners, and uniform concentration bounds for the second-moment estimates. We prove geometric decay of the empirical loss for full-batch Adam and in restricted expectation on a high-probability event for minibatch Adam. Beyond regression, we extend our convergence analysis to physics-informed neural networks (PINNs) for solving the Poisson equation, where differential operators and separate sampling of interior and boundary points introduce additional challenges. We establish global convergence guarantees for both deterministic and stochastic Adam under explicit conditions on network width and Adam's parameters, characterizing the effects of minibatch size on convergence. These results help bridge the gap between Adam's empirical success and its theoretical understanding, providing sharper convergence guarantees for PINNs than existing analyses.

\end{abstract}

\section{Introduction}

The Adam optimizer \cite{kingma2015adam} has been widely used to train deep neural networks in various applications, including computer vision~\cite{karras2019style}, natural language processing~\cite{vaswani2017attention}, and scientific machine learning \cite{wang2022and}. Despite its empirical success, the theoretical understanding of Adam remains incomplete. In practice, Adam can often reduce the nonnegative training loss to nearly zero~\cite{kingma2015adam,cattaneo2024on,deresse2025deep}, suggesting that it can approach a global minimizer even for highly nonlinear and nonconvex objectives. However, classical optimization theory generally guarantees convergence to stationary points rather than global minimizers for nonconvex problems~\cite{ghadimi2013stochastic,chen2018on}. This gap between empirical observations and existing theoretical guarantees motivates a closer study of the global convergence of Adam in neural network training.

One influential approach to understanding this phenomenon is based on the neural tangent kernel (NTK) framework \cite{jacot2018neural}, which describes the training dynamics of sufficiently wide neural networks through their tangent kernels. Based on this framework, several studies have established global convergence guarantees for gradient descent (GD)~\cite{du2018gradient,du2019gradient,oymak2019overparameterized} and stochastic gradient descent (SGD)~\cite{allen2019convergence,zou2019improved,xu2024overparametrized,vaswani2019fast,liu2022loss,gess2026exponential} under suitable overparameterization conditions. These results show that sufficiently wide neural networks can maintain favorable local regularity properties throughout training, leading to convergence to global minima. However, most existing finite-width analyses focus on GD and SGD, and their techniques do not directly extend to Adam. Unlike GD and SGD, Adam uses exponentially weighted moving averages of gradients and squared gradients, which introduce dependence across iterations and adaptive preconditioning. In addition, the stochastic gradient is correlated with the adaptive preconditioner, making the conditional unbiasedness arguments used for SGD no longer directly applicable. These challenges motivate a separate convergence analysis for Adam.

The convergence properties of Adam and related adaptive optimization methods have also received considerable attention \cite{reddi2018on,chen2018on,dereich2025ode}. Existing studies have established convergence guarantees for convex and nonconvex problems, including convergence to stationary points~\cite{barakat2021convergence,li2023convergence} and linear convergence under suitable Polyak--\L{}ojasiewicz (PL) conditions~\cite{chakrabarti2024methodology}. Although these results provide theoretical insights, they do not fully explain why Adam can reduce the training loss to nearly zero when training neural networks. In particular, convergence to a stationary point does not imply global optimality, while the global PL condition is generally difficult to verify for highly nonlinear neural network objectives. Several studies have also established global convergence guarantees for Adam in specific neural network models~\cite{zou2023understanding,tang2026understanding}. However, the global convergence of standard Adam in training finite-width neural networks, particularly under stochastic sampling, remains less understood.

Beyond standard regression problems, physics-informed neural networks (PINNs)~\cite{raissi2019physics} have become a popular approach for solving partial differential equations~\cite{karniadakis2021physics}. Unlike standard regression, PINN training involves differential operators acting on neural network outputs, together with boundary or initial conditions. These features introduce additional challenges in convergence analysis. Recent studies have extended NTK-based convergence theory to PINNs, establishing global convergence guarantees for GD~\cite{gao2023gradient}, SGD~\cite{jin2026convergence}, natural gradient descent~\cite{xu2026fast}, and implicit gradient descent~\cite{li2023implicit,xu2027}. Related convergence results have also been established for physics-informed Kolmogorov--Arnold networks~\cite{gao2025convergence}. However, these studies mainly focus on optimization methods other than Adam. Although the training dynamics of Adam for PINNs have also been studied~\cite{farhani2025simple}, global convergence guarantees for Adam remain limited, particularly under stochastic sampling.

To address these challenges, we study the global convergence of Adam in training two-layer neural networks for regression and physics-informed learning. We establish linear convergence guarantees under suitable conditions on the momentum parameters, damping parameter, step size, network width, and minibatch size. Our results provide a theoretical explanation for the empirical effectiveness of Adam in these settings. The main contributions are summarized as follows:
\begin{enumerate}
    \item We first study deterministic Adam for neural network regression. We establish global linear convergence for any fixed choice of momentum parameters, provided that the network is sufficiently wide and the other algorithmic parameters satisfy suitable conditions. Our proof uses a Lyapunov function to control the effect of historical momentum and establishes NTK stability throughout training.

    \item We extend the convergence analysis to stochastic Adam with minibatch sampling for regression. In this setting, additional conditions on the algorithmic parameters are needed to control the stochastic errors introduced by momentum and adaptive preconditioning. We develop a virtual-iterate analysis to handle historical momentum and a conditional analysis to control the bias arising from the dependence between stochastic gradients and adaptive preconditioners. Our results establish geometric decay of the empirical loss in restricted expectation on a high-probability event, under explicit conditions on network width and Adam's parameters. These conditions and the convergence rate also characterize the dependence on minibatch size.

    \item We further extend our analysis to physics-informed neural networks (PINNs) for solving the Poisson equation. The presence of differential operators and separate sampling of interior and boundary points introduces additional difficulties in controlling the training trajectory. By extending our convergence analysis to handle these difficulties, we establish analogous linear convergence guarantees with explicit width requirements and conditions on Adam's parameters, including the dependence on the PDE and boundary minibatch sizes.
\end{enumerate}

\noindent\textbf{Notation.}
Throughout this paper, we use $\|\cdot\|$ to denote the Euclidean norm for vectors and the spectral norm for matrices. 
For any event $\mathcal{A}$, the indicator function $\bm{1}_{\mathcal{A}}$ equals one if $\mathcal A$ occurs and zero otherwise, and $\mathcal{A}^{c}$ denotes its complement.
Given a positive integer $n$, $[n]$ is defined as the set $\{1,2,\cdots,n\}$.
We write $a \lesssim b$ (respectively, $a \gtrsim b$) if there exists a universal constant $C>0$ such that $a \leq Cb$ (respectively, $a \geq Cb$). We write $a \asymp b$ if both $a \lesssim b$ and $a \gtrsim b$ hold. 
We use $\widetilde{\Omega}$ and $\widetilde{\Theta}$ analogously to $\gtrsim$ and $\asymp$, respectively, with polylogarithmic factors suppressed.

\section{Convergence Analysis of Adam for Regression}

Consider a regression problem with distinct training samples $x_1,\ldots,x_n\in\R^d$ and corresponding targets $y_1,\ldots,y_n\in\R$. We assume that $\|x_i\|\leq 1$ and $|y_i|\leq 1$ for all $i=1,\ldots,n$. The regression problem is formulated as
\begin{equation}
\label{eq_main_prob}
    \min_{\theta} L(\theta)
    = \frac{1}{2n}\sum_{i=1}^{n}(f_\theta(x_i)-y_i)^2,
\end{equation}
where $f_\theta$ denotes a neural network parameterized by $\theta$, which contains all trainable parameters.

Throughout this paper, we consider a two-layer feedforward neural network of the form
\begin{equation}
\label{eq:reg-network}
    f_\theta(x)
    = \frac{1}{\sqrt{m}}\sum_{r=1}^m a_r \sigma(w_r^T\widetilde{x}),
    \qquad
    \theta=(a_1,w_1,\ldots,a_m,w_m),
\end{equation}
where $\widetilde{x}=(x,1)/\sqrt{2}$ is the augmented input vector that includes a bias term, $a_r\in\R$ and $w_r\in\R^{d+1}$ are trainable parameters, and $\sigma(\cdot)=\tanh(\cdot)$ is the hyperbolic tangent activation function. The network parameters are independently initialized as
\begin{equation}
\label{eq_init}
    w_r^{(0)}\sim\mathcal{N}(0,I_{d+1}),
    \qquad
    a_r^{(0)}\sim\mathrm{Uniform}(\{-1,1\}),
\end{equation}
for $r=1,\ldots,m$. Following previous studies~\cite{du2018gradient,jin2026convergence,gao2023gradient}, we adopt Rademacher initialization for the output weights $a_r$ for analytical convenience. Our analysis can also be extended to more general initialization distributions.

We study the convergence of Adam applied to Problem~\eqref{eq_main_prob}. Let $\theta^{(t)}=(a_1^{(t)},w_1^{(t)},\ldots,a_m^{(t)},w_m^{(t)})$ denote the network parameters at iteration $t$. At each iteration, Adam~\cite{kingma2015adam} computes a gradient estimate $g^{(t)}$ of $\nabla_\theta L(\theta^{(t)})$. Given momentum parameters $\beta_1,\beta_2\in[0,1)$, a damping parameter $\epsilon>0$, and a stepsize $\eta>0$, the moment estimates and their bias-corrected counterparts are computed as
\begin{equation}
\label{eq:adam-moment}
\begin{split}
    &s^{(t)}=\beta_1s^{(t-1)}+(1-\beta_1)g^{(t)},\quad
    v^{(t)}=\beta_2v^{(t-1)}+(1-\beta_2)g^{(t)\odot 2},\\
    &\widehat{s}^{(t)}=\frac{s^{(t)}}{1-\beta_1^{t+1}},\quad
    \widehat{v}^{(t)}=\frac{v^{(t)}}{1-\beta_2^{t+1}},\quad
    D^{(t)}=\diag\left(\frac{1}{\sqrt{\widehat{v}^{(t)}}+\epsilon}\right),
\end{split}
\end{equation}
where $s^{(-1)}=v^{(-1)}=0$. Here, $s^{(t)}$ and $v^{(t)}$ are the first- and second-moment estimates, respectively, and $\widehat{s}^{(t)}$ and $\widehat{v}^{(t)}$ are their bias-corrected counterparts. The matrix $D^{(t)}$ is the adaptive preconditioner. All powers, square roots, and divisions involving vectors are performed coordinatewise. The network parameters are then updated by
\begin{equation}
\label{eq:adam-update}
    \theta^{(t+1)}=\theta^{(t)}-\eta D^{(t)}\widehat{s}^{(t)}.
\end{equation}
For a network of width $m$, we choose the damping parameter and stepsize as
\begin{equation}
\label{eq:adam-scaling}
    \epsilon=\frac{\bar{\epsilon}}{\sqrt{m}}>0,
    \qquad
    \eta=\bar{\eta}\epsilon>0.
\end{equation}
Our main results provide guidelines for choosing $\bar{\epsilon}$ and $\bar{\eta}$.

Throughout our analysis, we assume that the Adam parameters $\beta_1$, $\beta_2$, $\epsilon$, and $\eta$ are fixed across iterations, although they may depend on the training data, network width, and other problem parameters. For deterministic Adam, our convergence results allow arbitrary fixed $\beta_1,\beta_2\in[0,1)$. For stochastic Adam with minibatch sampling, we allow any fixed $\beta_1\in[0,1)$, while $\beta_2\in(0,1)$ must be sufficiently close to one. The precise conditions on these parameters are specified in the main theorems.
Unlike GD and SGD, Adam uses both first-moment momentum and second-moment-based adaptive preconditioning, as shown in \eqref{eq:adam-moment}. The resulting dependence on historical gradients and adaptive preconditioners introduces additional challenges to the convergence analysis. Consequently, existing techniques for GD and SGD cannot be directly applied.

The convergence analysis of Adam relies on the properties of the loss function $L(\theta)$. To study these properties, we define the normalized residuals
\begin{equation*}
    R_i(\theta)=\frac{f_\theta(x_i)-y_i}{\sqrt{n}},
    \qquad
    L(\theta)=\frac{1}{2}\|R(\theta)\|^2,
\end{equation*}
where $R(\theta)=(R_1(\theta),\ldots,R_n(\theta))^\top\in\R^n$ is the normalized residual vector. We further define
\begin{equation}
\label{eq:residual-form}
    J(\theta)=\nabla R(\theta),
    \qquad
    K(\theta)=J(\theta)J(\theta)^\top,
\end{equation}
where $J(\theta)\in\R^{n\times m(d+2)}$ is the Jacobian matrix of $R(\theta)$ and $K(\theta)\in\R^{n\times n}$ is the associated Gram matrix. Since $\nabla L(\theta)=J(\theta)^\top R(\theta)$, we have
\begin{equation}
\label{eq:kernel-pl}
    \|\nabla L(\theta)\|^2
    =R(\theta)^\top K(\theta)R(\theta)
    \geq 2\lambda_{\min}(K(\theta))L(\theta).
\end{equation}
This inequality relates the positive definiteness of the Gram matrix to the Polyak--\L{}ojasiewicz (PL) inequality~\cite{karimi2016linear,oymak2019overparameterized}. In particular, if $\lambda_{\min}(K(\theta))$ remains uniformly bounded away from zero along the optimization trajectory, then the PL inequality holds along the trajectory. This property is important for establishing linear convergence of Adam. Our analysis therefore focuses on maintaining a positive lower bound on the minimum eigenvalue of the Gram matrix throughout training.

To this end, we define the limiting Gram matrix associated with the random initialization~\eqref{eq_init} as
\begin{equation}
\label{eq:limiting-kernel}
    K_\infty=\E_{\theta^{(0)}}[K(\theta^{(0)})],
    \qquad
    \lambda_0=\lambda_{\min}(K_\infty).
\end{equation}
Although $K(\theta)$ changes during training, we will show that it remains close to its initial value, which can be controlled through $K_\infty$. The following lemma establishes the positive definiteness of $K_\infty$.

\begin{lemma}[Positive definiteness of $K_\infty$]
\label{lem:reg-positivity}
    The limiting Gram matrix $K_\infty$ is positive definite, i.e., $\lambda_0>0$.
\end{lemma}

The proof is provided in Appendix~\ref{app:gram_pd} and uses the full support of the Gaussian distribution and properties of the $\tanh$ activation function. Similar positive definiteness results have been established in the NTK literature~\cite{du2018gradient,carvalho2025positivity,du2019gradient}. However, extending the convergence analysis from GD and SGD to Adam requires new techniques to handle momentum and adaptive preconditioning, which constitute the main theoretical contributions of this work.

\subsection{Convergence of Deterministic Adam}

We first study the convergence of deterministic Adam, where the exact gradient $g^{(t)}=\nabla_\theta L(\theta^{(t)})$ is used in \eqref{eq:adam-moment} at each iteration. This setting allows us to study the effects of momentum and adaptive preconditioning without the randomness introduced by minibatch sampling. Nevertheless, the convergence analysis remains challenging due to the dependence on historical gradients and the time-varying preconditioner. Our first main result establishes linear convergence of deterministic Adam under suitable conditions on the network width and algorithmic parameters.

\begin{theorem}[Regression with full gradients]
\label{thm:reg-det}
Let $\delta\in(0,1)$. Consider Adam defined by \eqref{eq:adam-moment}--\eqref{eq:adam-update} with full gradients $g^{(t)}=\nabla L(\theta^{(t)})$ and arbitrary fixed momentum parameters $\beta_1,\beta_2\in[0,1)$. Suppose that
\begin{equation}
\label{eq:reg-det-parameters}
    \bar{\epsilon}\gtrsim\sqrt{\log\frac{2}{\delta}},
    \qquad
    0<\bar\eta\lesssim 1-\beta_1,
\end{equation}
and that the network width satisfies
\begin{equation}
\label{eq:reg-det-width}
    m\gtrsim\frac{1}{\lambda_0}\log\frac{n}{\delta}
    +\frac{1}{\lambda_0^2}\log\frac{2}{\delta}.
\end{equation}
Then, with probability at least $1-\delta$ over the random initialization~\eqref{eq_init}, the Adam iterates satisfy
\begin{equation}
\label{eq:reg-det-rate}
    L(\theta^{(t)})
    \leq (1-c\bar\eta\lambda_0)^t L(\theta^{(0)}),
    \qquad t\geq 0,
\end{equation}
where $c>0$ is a universal constant. Moreover, the parameter sequence $\{\theta^{(t)}\}$ converges to a finite limit $\theta^{\star}$ satisfying $L(\theta^{\star})=0$.
\end{theorem}

\begin{proof}[Proof sketch]
The complete proof is provided in Appendix~\ref{app:reg-det-proof}. The proof consists of three main steps: concentration at initialization, stability of the Gram matrix, and a Lyapunov-based descent analysis.

First, under the network width condition~\eqref{eq:reg-det-width}, concentration inequalities imply that
\[
K(\theta^{(0)})\succeq \frac34\lambda_0 I,
\qquad
L_0:=L(\theta^{(0)})\lesssim \log\frac2\delta,
\]
with probability at least $1-\delta$.
As long as the output coefficients remain bounded, the residual Jacobian $J(\theta)$ is Lipschitz continuous with constant $C/\sqrt{m}$. Consequently,
\[
K(\theta)\succeq \frac{\lambda_0}{4}I
\]
for all $\theta$ satisfying
\[
\|\theta-\theta^{(0)}\|
\leq \mathcal R:=c_0\sqrt{m\lambda_0},
\]
provided that the output coefficients remain bounded, where $c_0>0$ is a sufficiently small universal constant.
By~\eqref{eq:kernel-pl}, this yields a uniform PL inequality in this neighborhood.

Next, we control the difference between the bias-corrected first-moment estimate $\widehat{s}^{(t)}$ and the current gradient $g^{(t)}$. Define the normalized preconditioner $P^{(t)}=\epsilon D^{(t)}$. As long as the output weights remain bounded and the loss does not exceed $L_0$, the coordinatewise gradient bounds and the damping condition imply that $\frac12 I\preceq P^{(t)}\preceq I.$
Combining the descent inequality for the loss function with the recurrence relation for $\widehat{s}^{(t)}-g^{(t)}$, we construct the Lyapunov function
\[
\mathcal E^{(t)}
:=L(\theta^{(t)})
+\frac{4\bar\eta}{1-\beta_1}
\|\widehat{s}^{(t)}-g^{(t)}\|^2,
\]
which satisfies
\[
\mathcal E^{(t+1)}
\leq (1-c\bar\eta\lambda_0)\mathcal E^{(t)}.
\]
This contraction follows from the local lower bound on the Gram matrix and the smoothness of the loss function along each update segment.

To ensure that the preceding conditions hold throughout training, we establish two bounds on the cumulative parameter movement:
\[
\begin{aligned}
\sum_{t=0}^{T}
\|\theta^{(t+1)}-\theta^{(t)}\|
&\lesssim \sqrt{\frac{L_0}{\lambda_0}},\\
\sum_{t=0}^{T}\max_{1\le r\le m}
\|\theta_r^{(t+1)}-\theta_r^{(t)}\|
&\lesssim \frac{\sqrt{L_0}}{\lambda_0\sqrt m},
\end{aligned}
\]
where $\theta_r=(a_r,w_r)$ denotes the parameters of the $r$-th neuron. The first bound follows from the Lyapunov descent analysis, while the second uses the exact bias-corrected moment weights. When $m\gtrsim L_0/\lambda_0^2$, these bounds ensure that the iterates remain in the region where the Gram matrix is stable, and the output weights remain bounded. The bound on cumulative neuron-wise movement also ensures that each update segment stays in the region where the smoothness conditions hold. These properties allow us to close the induction and establish the contraction for all iterations.

Finally, the bias correction gives $\widehat{s}^{(0)}=g^{(0)}$, so $\mathcal E^{(0)}=L_0$. Combining this identity with the contraction of $\mathcal E^{(t)}$ and the bound $L(\theta^{(t)})\leq\mathcal E^{(t)}$ yields the claimed linear convergence rate.
\end{proof}

\paragraph{Discussion.}
Theorem~\ref{thm:reg-det} establishes linear convergence of deterministic Adam for arbitrary fixed momentum parameters $\beta_1,\beta_2\in[0,1)$, provided that the damping parameter, stepsize, and network width satisfy the conditions~\eqref{eq:reg-det-parameters}--\eqref{eq:reg-det-width}. The first-moment parameter $\beta_1$ affects the admissible normalized stepsize through $\bar\eta\lesssim 1-\beta_1$. Thus, a larger $\beta_1$ requires a smaller stepsize under our conditions. We emphasize that our result guarantees convergence with fixed momentum parameters but does not establish any acceleration due to momentum.

The network width requirement in \eqref{eq:reg-det-width} consists of two terms with different roles. The term $\lambda_0^{-1}\log(n/\delta)$ ensures a positive lower bound on the minimum eigenvalue of the Gram matrix at initialization, while $\lambda_0^{-2}\log(2/\delta)$ controls parameter movement and maintains this lower bound throughout training. For fixed training data, $\lambda_0$ is independent of the network width $m$, and the normalized stepsize $\bar\eta$ can also be chosen independently of $m$. Therefore, once the width requirement is satisfied, the contraction factor $1-c\bar\eta\lambda_0$ does not depend on $m$.

The actual stepsize $\eta=\bar\eta\bar\epsilon/\sqrt{m}$ decreases as the network width $m$ increases, while the contraction factor does not explicitly depend on $m$. To explain this, recall the normalized preconditioner $P^{(t)}=\epsilon D^{(t)}$, which allows us to rewrite the Adam update as
\[
\theta^{(t+1)}
=\theta^{(t)}-\bar\eta P^{(t)}\widehat{s}^{(t)}.
\]
Under the gradient bounds and the damping condition, the eigenvalues of $P^{(t)}$ remain uniformly bounded above and away from zero throughout training. Thus, the width-dependent scaling of $\eta$ compensates for the adaptive preconditioning, and the convergence rate is determined by the normalized stepsize $\bar\eta$ rather than $\eta$.

\subsection{Convergence of Stochastic Adam}

In practice, computing the full gradient at each iteration can be expensive for large training datasets. Therefore, Adam is commonly implemented using stochastic gradients computed from minibatches. In this section, we extend our analysis to stochastic Adam and establish linear convergence guarantees under suitable conditions on the algorithmic parameters and network width.

At each iteration $t$, we sample a minibatch $B^{(t)}\subset[n]$ of size $b$ uniformly without replacement, independently across iterations. The stochastic gradient is given by
\begin{equation}
\label{eq:minibatch_gradients}
    g^{(t)}=\frac{1}{b}\sum_{i\in B^{(t)}}
    \big(f_{\theta^{(t)}}(x_i)-y_i\big)
    \nabla_\theta f_{\theta^{(t)}}(x_i).
\end{equation}
The stochastic gradient is conditionally unbiased, satisfying
$\E[g^{(t)}\mid\mathcal F_t]=\nabla_\theta L(\theta^{(t)})$,
where $\mathcal F_t$ is the sigma-algebra generated by the initialization and all minibatches sampled before iteration $t$.
Compared with deterministic Adam, the stochastic setting introduces additional difficulties due to gradient noise and the dependence between stochastic gradients, momentum, and adaptive preconditioners. To characterize the effect of minibatch sampling, we define
\begin{equation}
\label{eq:reg-rho}
    \rho_b=1+\frac{n-b}{b(n-1)\lambda_0},
\end{equation}
for $n\geq 2$. This quantity captures the effects of minibatch sampling and the minimum eigenvalue of the limiting Gram matrix.

\begin{theorem}[Regression with minibatches]
\label{thm:reg-stoch}
Let $\delta\in(0,1)$, $n\geq 2$, and $1\leq b\leq n$. Consider Adam defined by
\eqref{eq:adam-moment}--\eqref{eq:adam-update} with minibatch gradients
\eqref{eq:minibatch_gradients} and any fixed $\beta_1\in[0,1)$.
Let $\rho_b$ be defined in \eqref{eq:reg-rho}, and suppose that
\begin{equation}
\label{eq:reg-stoch-damping}
    \bar\epsilon\gtrsim
    \sqrt{\frac{1}{\delta}\log\frac{2}{\delta}}.
\end{equation}
Choose fixed $\beta_2\in(0,1)$ and $\bar\eta>0$ such that
\begin{equation}
\label{eq:reg-stoch-parameters}
    0<1-\beta_2\lesssim
    \frac{1-\beta_1}{
    \rho_b\log\frac{\rho_b}{(1-\beta_1)\lambda_0\delta}},
    \qquad
    \bar\eta\asymp 1-\beta_2.
\end{equation}
Suppose further that the network width satisfies
\begin{equation}
\label{eq:reg-stoch-width}
    m\gtrsim
    \frac{1}{\lambda_0}\log\frac{n}{\delta}
    +\frac{1}{\delta\lambda_0^2}\log\frac{2}{\delta}.
\end{equation}
Then, with probability at least $1-\delta/2$ over the random initialization~\eqref{eq_init}, there exists an event $\mathcal{G}$ associated with minibatch sampling such that
\begin{equation}
\label{eq:reg-stoch-rate}
    \mathbb P(\mathcal{G}\mid\theta^{(0)})\geq 1-\delta/2,
    \qquad
    \E\left[
    L(\theta^{(t)})\mathbf{1}_{\mathcal{G}}
    \mid\theta^{(0)}
    \right]
    \leq C(1-c\bar\eta\lambda_0)^tL(\theta^{(0)}),
    \quad t\geq 0,
\end{equation}
where $C,c>0$ are universal constants. In particular, the joint probability of successful initialization and minibatch sampling is at least $1-\delta$.

Moreover, on the high-probability event $\mathcal G$, the parameter sequence $\{\theta^{(t)}\}$ converges almost surely to a global minimizer $\theta^{\star}$ satisfying $L(\theta^{\star})=0$.
\end{theorem}

\begin{proof}[Proof sketch]
The complete proof is provided in Appendix~\ref{app:reg-stoch-proof}.
The proof consists of three main steps: concentration at initialization, a conditional descent analysis, and a stopping argument to control the stochastic trajectory.
Let $L_0=L(\theta^{(0)})$, $\gamma=1-\beta_1$, and $\nu=1-\beta_2$.

First, as in the proof of Theorem~\ref{thm:reg-det}, the initialization bounds
\[
K(\theta^{(0)})\succeq\frac34\lambda_0 I,
\qquad
L_0\lesssim\log\frac2\delta
\]
hold with probability at least $1-\delta/2$.
We condition on an initialization satisfying these bounds.
To handle the stochastic updates, we introduce an auxiliary network that is globally smooth and agrees with the original network whenever $|a_r|\leq 2$ for all $r$.
The regularity of the auxiliary network ensures that
\[
K(\theta)\succeq\frac{\lambda_0}{4}I
\]
whenever
\[
\|\theta-\theta^{(0)}\|
\leq\mathcal R:=c_0\sqrt{m\lambda_0}.
\]
We analyze Adam on this auxiliary network and show that its iterates agree with those of the original network on a suitable high-probability event.

Recall the filtration $\mathcal F_t$ defined above, and write
$\E_t[\cdot]=\E[\cdot\mid\mathcal F_t]$.
Using the exact covariance formula for uniform minibatch sampling, we obtain
\begin{equation}
\label{eq_cov_grad}
    \E_t\|g^{(t)}-\nabla L(\theta^{(t)})\|^2
    \lesssim
    (\rho_b-1)\|\nabla L(\theta^{(t)})\|^2
\end{equation}
within the region of Gram matrix stability.
To control the adaptive preconditioner, we bound each coordinate of the bias-corrected second-moment estimate $\widehat v^{(t)}$ by a scalar moving average of the sampled squared residuals.
A uniform-in-time concentration bound for this stopped process then leads to the logarithmic restriction on $\nu$ in \eqref{eq:reg-stoch-parameters}.

Next, we derive a conditional descent estimate that accounts for the dependence between the current stochastic gradient and the adaptive preconditioner.
Recall that $P^{(t)}=\epsilon D^{(t)}$.
Using the exact bias-correction formulas, we introduce the virtual iterate
\begin{equation}
\label{eq5}
    Z^{(t)}
=\theta^{(t)}
-\bar\eta k^{(t)}P^{(t-1)}\widehat{s}^{(t-1)},
\qquad
0\leq k^{(t)}\leq\frac{\beta_1}{\gamma},
\end{equation}
where the deterministic coefficients $k^{(t)}$ are chosen to eliminate the contribution of historical first-moment estimates from the virtual update.
This allows us to separate the virtual update into a main descent term involving a preconditioner determined by past gradients and a remainder term controlled by the second moments of the stochastic gradients.

Combining the descent estimate with the first-moment recurrence, we construct the Lyapunov function
\[
\mathcal E^{(t)}
:=L(Z^{(t)})
+C_0\frac{\bar\eta\nu}{\gamma^2}
\|\widehat{s}^{(t-1)}\|^2,
\]
which satisfies
\[
\E_t\mathcal E^{(t+1)}
\leq(1-c\bar\eta\lambda_0)\mathcal E^{(t)},
\qquad
L(\theta^{(t)})\leq2\mathcal E^{(t)},
\]
for $t\geq T_*\asymp\nu^{-1}$, provided that the local Gram matrix, loss, and historical second-moment bounds hold. The first $T_*$ iterations are treated separately, yielding
\[
\E[\mathcal E^{(T_*)}\mid\theta^{(0)}]
\lesssim L_0.
\]
Since $\bar\eta T_*\lesssim1$, this initial phase contributes only a universal constant factor to the final convergence bound.

To ensure that the local conditions hold throughout training, we introduce a stopping time for the auxiliary process, defined as the first iteration at which any of the prescribed energy, second-moment, or parameter movement bounds fails. The descent estimate controls the predictable part of the virtual iterate movement, while martingale maximal inequalities control the stochastic fluctuations. In addition, the exact bias-corrected moment weights allow us to control the cumulative movement of all neurons simultaneously. Under the network width and algorithmic parameter conditions, these estimates yield an event $\mathcal G$ with conditional probability at least $1-\delta/2$. On this event, the iterates remain within the region of Gram matrix stability, and $|a_r^{(t)}|\leq 3/2$ for all $r$ and $t$.

Applying the conditional contraction estimate before the stopping time yields geometric decay of the loss in restricted expectation on $\mathcal G$. On this event, the auxiliary and original Adam iterates coincide, including their gradients, moment estimates, and parameter updates. Thus, the convergence bound also holds for the original Adam iterates:
\[
\E\left[
L(\theta^{(t)})\mathbf{1}_{\mathcal G}
\mid\theta^{(0)}
\right]
\leq C(1-c\bar\eta\lambda_0)^tL_0,
\qquad t\geq 0.
\]
Combining the probability bounds for initialization and minibatch sampling completes the proof.
\end{proof}

\paragraph{Discussion.}
The parameter $\rho_b$ reflects the effects of minibatch sampling and the minimum eigenvalue of the limiting Gram matrix. As shown in \eqref{eq_cov_grad}, $\rho_b-1$ controls the variance of the stochastic gradient relative to the squared norm of the full gradient. In particular, $\rho_b$ decreases as the batch size $b$ increases, with $\rho_1=1+\lambda_0^{-1}$ and $\rho_n=1$. Moreover, the stochastic gradient variance decreases as the training loss approaches zero. This allows Adam to converge to zero training loss without requiring its parameters to decay over iterations.

The restriction on $\beta_2$ is needed to control the fluctuations of the adaptive preconditioner and its dependence on the current stochastic gradient. Although $g^{(t)}$ is conditionally unbiased, the same minibatch is also used to compute $\widehat v^{(t)}$. Thus, conditional unbiasedness alone is insufficient to establish descent for the preconditioned update. In particular, the Lyapunov descent analysis requires
\[
\frac{(1-\beta_2)\rho_b}{1-\beta_1}
\]
to be sufficiently small. The additional logarithmic factor in \eqref{eq:reg-stoch-parameters} comes from controlling the second-moment estimates uniformly over all iterations. The condition $\bar\eta\asymp 1-\beta_2$ allows us to choose the normalized stepsize and the second-moment parameter jointly. In particular, a larger minibatch size reduces $\rho_b$, allowing a larger value of $1-\beta_2$ and hence a larger normalized stepsize under our theoretical conditions. 
Moreover, \eqref{eq:reg-stoch-parameters} and $\rho_b>1$ imply a higher requirement on the second-moment parameter $\beta_2$ than on the first-moment parameter $\beta_1$. This is consistent with the default Adam settings $\beta_1=0.9$ and $\beta_2=0.999$~\cite{kingma2015adam}.

Compared with Theorem~\ref{thm:reg-det}, the stochastic convergence result requires stronger conditions on the Adam parameters and a larger network width. The additional dependence on the confidence parameter $\delta$ arises from controlling the stochastic training trajectory with high probability. In particular, the stopping argument uses an energy threshold of order $S\asymp\frac{1}{\delta}\log\frac{2}{\delta},$ which ensures that the probability of exceeding this threshold is of order $\delta$. The condition $\bar\epsilon^2\gtrsim S$ then provides the required control of the adaptive preconditioner. Moreover, the parameter movement bounds imply that the probability of leaving the region of Gram matrix stability is at most $\frac{CL_0}{m\lambda_0^2}.$
Together with $L_0\lesssim\log(2/\delta)$ on the initialization event, this yields the additional width requirement $\delta^{-1}\lambda_0^{-2}\log(2/\delta)$. The initialization requirement $\lambda_0^{-1}\log(n/\delta)$ remains unchanged.

The stochastic analysis differs from the deterministic case in two main aspects: controlling the stochastic training trajectory and handling the dependence between stochastic gradients and adaptive preconditioners. In the full-gradient setting, this dependence does not involve sampling randomness, allowing arbitrary fixed $\beta_2\in[0,1)$. Thus, when $b=n$, Theorem~\ref{thm:reg-det} applies directly without the additional restrictions of the stochastic theorem. Moreover, for fixed training data and confidence level, the Adam parameters can be chosen independently of the network width $m$. Hence, the contraction factor does not explicitly depend on $m$, despite the width-dependent scaling of the damping parameter and stepsize.

The convergence bound in \eqref{eq:reg-stoch-rate} controls the expected loss restricted to a single high-probability event $\mathcal G$, on which the iterates remain within the region of Gram matrix stability. This gives geometric decay in the restricted expectation. A similar localized convergence guarantee based on a high-probability event appears in the nonlinear least-squares analysis of SGD in \cite[Theorem~3.1]{oymak2019overparameterized}. Our theorem also provides separate probability guarantees for initialization and minibatch sampling, accounting for both sources of randomness.

Finally, we next compare our convergence guarantees with those of GD and SGD. Under the same local regularity conditions, the loss function satisfies a local PL inequality with a constant of order $\lambda_0$. Moreover, the stochastic gradient satisfies
\[
\E_t\|g^{(t)}\|^2
\lesssim\rho_b\|\nabla L(\theta^{(t)})\|^2.
\]
A standard SGD descent argument with stepsize $h\asymp\rho_b^{-1}$ then yields a conditional contraction of the loss with a contraction gap of order $\lambda_0/\rho_b$, as long as the iterates remain in the region where the local conditions hold. A related result under global PL and strong-growth conditions is given in \cite[Theorem~4]{vaswani2019fast}. 
Compared with previous results for GD and SGD~\cite{du2018gradient,allen2019convergence,zou2019improved,xu2024overparametrized,vaswani2019fast}, our analysis provides sharper width requirements and improved convergence-rate bounds. Compared with existing convergence analyses of Adam~\cite{dereich2025ode,chakrabarti2024methodology}, which often assume the PL condition, our results establish explicit conditions on the network width and optimizer parameters, while also characterizing the effect of minibatch size on convergence.

\section{Convergence Analysis of Adam for PINNs}
\label{sec:pinn}

We now extend our convergence analysis of Adam from regression to physics-informed neural networks (PINNs)~\cite{raissi2019physics}. Unlike standard regression, PINN training involves differential operators acting on neural network outputs, which introduces additional challenges in analyzing the residual Jacobian and the associated Gram matrix. Following existing convergence analyses of GD and SGD for PINNs~\cite{gao2023gradient,xu2026fast,jin2026convergence}, we consider the Poisson equation with Dirichlet boundary conditions:
\begin{equation}
\label{eq:pinn-problem}
    \Delta u=f \quad \text{in }\Omega,
    \qquad
    u=\varphi \quad \text{on }\partial\Omega,
\end{equation}
where $\Omega\subset\R^d$ is a bounded domain with $d\geq1$, $f$ is the source term, and $\varphi$ denotes the boundary data. Although we focus on the Poisson equation, our analysis can also be extended to other PDEs under suitable assumptions.

Let $x_1,\ldots,x_{n_{\rm PDE}}\in\Omega$ and $z_1,\ldots,z_{n_{\rm BC}}\in\partial\Omega$ denote the fixed interior and boundary collocation points, respectively, where $n_{\rm PDE},n_{\rm BC}\geq1$. We assume that all collocation points are pairwise distinct and satisfy
\[
    \|x_i\|,\|z_j\|\leq1,
    \qquad
    |f(x_i)|,|\varphi(z_j)|\leq1.
\]
We use the same two-layer neural network architecture and initialization as in \eqref{eq:reg-network} and \eqref{eq_init}, respectively, and denote its output by $u_\theta(x)$.
For a boundary penalty parameter $\alpha_{\rm BC}>0$, we consider the empirical PINN loss
\begin{equation}
\label{eq:pinn-loss}
    L(\theta)
    =\frac{1}{2n_{\rm PDE}}
    \sum_{i=1}^{n_{\rm PDE}}
    \bigl(\Delta u_\theta(x_i)-f(x_i)\bigr)^2
    +\frac{\alpha_{\rm BC}}{2n_{\rm BC}}
    \sum_{j=1}^{n_{\rm BC}}
    \bigl(u_\theta(z_j)-\varphi(z_j)\bigr)^2.
\end{equation}
Here, we regard $\alpha_{\rm BC}$ as a fixed universal constant.

We use the same notation $L$, $R$, $J$, $K$, $K_\infty$, and $\lambda_0$ as in the regression analysis, now defined for the PINN loss. Specifically, let $N=n_{\rm PDE}+n_{\rm BC}$ and define the normalized residual vector
\begin{equation}
\label{eq:pinn-residual}
R(\theta)=
\begin{pmatrix}
\bigl((\Delta u_\theta(x_i)-f(x_i))/\sqrt{n_{\rm PDE}}\bigr)_i\\
\bigl(\sqrt{\alpha_{\rm BC}/n_{\rm BC}}\,
(u_\theta(z_j)-\varphi(z_j))\bigr)_j
\end{pmatrix}
\in\R^N,
\qquad
L(\theta)=\frac12\|R(\theta)\|^2.
\end{equation}
As before, we define
\[
J(\theta)=\nabla R(\theta),
\qquad
K(\theta)=J(\theta)J(\theta)^\top,
\]
and
\[
K_\infty=\E_{\theta^{(0)}}[K(\theta^{(0)})],
\qquad
\lambda_0=\lambda_{\min}(K_\infty).
\]
The gradient inequality~\eqref{eq:kernel-pl} therefore also holds for the PINN loss. The following lemma establishes the positive definiteness of the limiting Gram matrix $K_\infty$ in this setting.

\begin{lemma}[Positive definiteness of the PINN Gram matrix]
\label{lem:pinn-positivity}
The limiting PINN Gram matrix $K_\infty$ is positive definite, i.e., $\lambda_0>0$.
\end{lemma}

The proof is provided in Appendix~\ref{app:pinn-gram-proof}. The positive definiteness follows from the properties of the activation function and the distinctness of the collocation points, without imposing an additional positive definiteness assumption on $K_\infty$. Similar results on the positive definiteness of PINN Gram matrix can also be found in~\cite{jin2026differential,jin2026convergence,gao2023gradient,xu2026fast}.

\subsection{Convergence of Deterministic Adam}

We first study the convergence of deterministic Adam for training PINNs. At each iteration, we use the exact gradient $g^{(t)}=\nabla L(\theta^{(t)})$ of the PINN loss~\eqref{eq:pinn-loss} in the Adam updates~\eqref{eq:adam-moment}--\eqref{eq:adam-update}, with the damping parameter and the stepsize~\eqref{eq:adam-scaling}. Compared with regression, the PDE residual involves derivatives of the network output with respect to its inputs, introducing additional terms in the residual Jacobian and its regularity estimates. This requires more careful control of the network parameters and the Gram matrix throughout training.

\begin{theorem}[Poisson PINN with full gradients]
\label{thm:pinn-det}
Let $\delta\in(0,1)$ and consider Adam with full gradients $g^{(t)}=\nabla L(\theta^{(t)})$ and arbitrary fixed momentum parameters $\beta_1,\beta_2\in[0,1)$. Suppose that
\begin{equation}
\label{eq:pinn-det-parameters}
\bar\epsilon\gtrsim
d\sqrt{
\left(
d^2+\log^2\frac{md^2}{\delta\lambda_0}
\right)
\log\frac{2}{\delta}
},
\qquad
0<\bar\eta\lesssim\frac{1-\beta_1}{d^2},
\end{equation}
and that the network width satisfies
\begin{equation}
\label{eq:pinn-det-width}
m\gtrsim
\left(
d^2+\log^2\frac{md^2}{\delta\lambda_0}
\right)
\left(
\frac{1}{\lambda_0}\log\frac{N}{\delta}
+\frac{d^2}{\lambda_0^2}\log\frac{2}{\delta}
\right).
\end{equation}
Then, with probability at least $1-\delta$ over the random initialization~\eqref{eq_init}, the Adam iterates satisfy
\begin{equation}
\label{eq:pinn-det-rate}
L(\theta^{(t)})
\leq
(1-c\bar\eta\lambda_0)^tL(\theta^{(0)}),
\qquad t\geq0,
\end{equation}
where $c>0$ is a universal constant.
Moreover, the parameter sequence $\{\theta^{(t)}\}$ converges to a finite limit $\theta^{\star}$ satisfying $L(\theta^{\star})=0$.
\end{theorem}

\begin{proof}[Proof sketch]
The complete proof is provided in Appendix~\ref{app:pinn-det-proof}.
The proof follows the same framework as in Theorem~\ref{thm:reg-det}, with additional estimates to handle the PDE residuals.
For convenience, define
\begin{equation}
\label{eq_l_m}
    \ell_m
    =d^2+\log^2\left(
    \frac{C_*md^2}{\delta\lambda_0}
    \right),
\end{equation}
where $C_*>1$ is a universal constant.

First, under the network width condition~\eqref{eq:pinn-det-width}, concentration inequalities imply that
\[
K(\theta^{(0)})\succeq\frac{3\lambda_0}{4}I,
\qquad
L_0:=L(\theta^{(0)})
\lesssim d^2\log\frac{2}{\delta},
\]
with probability at least $1-\delta$.
Unlike regression, the PDE residuals involve derivatives of the neural network, which require additional control of the hidden weights.
Specifically, when $|a_r|\leq2$ and $\|w_r-w_r^{(0)}\|\leq1$ for all $r$, the residual Jacobian is Lipschitz continuous with constant $C\sqrt{\ell_m/m}$.
Combining this estimate with the initialization bound and a singular-value perturbation argument, we obtain
\[
K(\theta)\succeq\frac{\lambda_0}{4}I
\]
whenever
\[
\|\theta-\theta^{(0)}\|
\leq\mathcal R
:=c_0\sqrt{\frac{m\lambda_0}{\ell_m}}.
\]
This yields a local PL inequality, as in the regression analysis.

Next, we use the same Lyapunov-based descent analysis as in the proof of Theorem~\ref{thm:reg-det}. Under the damping condition~\eqref{eq:pinn-det-parameters} and the local bounds on the network parameters and loss, the normalized preconditioner $P^{(t)}=\epsilon D^{(t)}$ satisfies $\frac12 I\preceq P^{(t)}\preceq I.$
Using the same Lyapunov function as in the regression analysis,
\[
\mathcal E^{(t)}
:=L(\theta^{(t)})
+\frac{4\bar\eta}{1-\beta_1}
\|\widehat s^{(t)}-g^{(t)}\|^2,
\]
we obtain
\[
\mathcal E^{(t+1)}
\leq(1-c\bar\eta\lambda_0)\mathcal E^{(t)}.
\]
The main difference from regression is the loss Hessian bound, which requires additional estimates for the PDE residuals. These estimates ensure that the descent inequality holds along each update segment.

As in the regression analysis, we establish two bounds on the cumulative parameter movement:
\[
\begin{aligned}
\sum_{t=0}^{T}
\|\theta^{(t+1)}-\theta^{(t)}\|
&\lesssim
\sqrt{\frac{L_0}{\lambda_0}},\\
\sum_{t=0}^{T}
\max_{1\leq r\leq m}
\|\theta_r^{(t+1)}-\theta_r^{(t)}\|
&\lesssim
\frac{1}{\lambda_0}
\sqrt{\frac{\ell_mL_0}{m}},
\end{aligned}
\]
where $\theta_r=(a_r,w_r)$ denotes the parameters of the $r$-th neuron. The first bound follows from the Lyapunov descent analysis, while the second uses the exact bias-corrected moment weights and the neuron-wise gradient bounds.

Under the width condition $m\gtrsim\ell_mL_0/\lambda_0^2$, these bounds allow us to close the induction argument. Specifically, the neuron-wise movement bound ensures that each update segment stays within the region where the network regularity estimates hold. This provides the required Hessian bound for the Lyapunov descent analysis. Meanwhile, the bound on total parameter movement ensures that the iterates remain within the region of Gram matrix stability. Together, these bounds establish the Lyapunov contraction for all iterations.

Finally, exact bias correction gives $\mathcal E^{(0)}=L_0$. Combining the Lyapunov contraction with $L(\theta^{(t)})\leq\mathcal E^{(t)}$ yields the stated linear convergence rate. Furthermore, the finite cumulative parameter movement implies that the parameter iterates converge to a limit.
\end{proof}

\paragraph{Discussion.}
As in Theorem~\ref{thm:reg-det}, our convergence guarantee allows arbitrary fixed momentum parameters $\beta_1,\beta_2\in[0,1)$, without requiring any additional relationship between them. Compared with regression, the main difference arises from the second-order spatial derivatives in the PDE residuals. These derivatives introduce additional dependence on the input dimension $d$ in the regularity estimates. Moreover, controlling the maximum over the Gaussian-initialized neurons introduces logarithmic factors, captured by $\ell_m$, in the network width and damping requirements. By separating these two types of estimates, we obtain an admissible normalized stepsize without additional logarithmic dependence on the network width $m$.

The two terms in the width requirement~\eqref{eq:pinn-det-width} serve different purposes. The first ensures a positive lower bound on the Gram matrix at initialization, while the second controls parameter movement to maintain this bound throughout training. Although the width requirement involves $m$ on both sides, its dependence on $m$ on the right-hand side is only logarithmic. Thus, the condition holds for all sufficiently large $m$ when the other problem parameters are fixed. Unlike in regression, the required lower bound on $\bar\epsilon$ also depends logarithmically on $m$. Choosing $\bar\epsilon$ at this order and keeping $\bar\eta$ fixed, the scaling~\eqref{eq:adam-scaling} gives damping and stepsize parameters of order $m^{-1/2}$ up to logarithmic factors.

Finally, Theorem~\ref{thm:pinn-det} establishes linear convergence of Adam for the empirical PINN loss evaluated at the collocation points. However, convergence of the empirical loss does not necessarily imply that the neural network converges to the true PDE solution throughout $\Omega$. Such a guarantee requires additional generalization and PDE stability analysis, which are beyond the scope of this paper. We leave this question for future work.

\subsection{Convergence of Stochastic Adam}
\label{subsec:pinn-stoch}

We now study stochastic Adam applied to the empirical PINN loss~\eqref{eq:pinn-loss}. At each iteration $t$, we independently sample two minibatches
$B_{\rm PDE}^{(t)}\subset[n_{\rm PDE}]$ and
$B_{\rm BC}^{(t)}\subset[n_{\rm BC}]$
of sizes $b_{\rm PDE}$ and $b_{\rm BC}$, respectively, uniformly without replacement. The minibatches are sampled independently across iterations and independently of the initialization, while the collocation points remain fixed. The stochastic gradient is given by
\begin{equation}
\label{eq:pinn-minibatch-gradients}
\begin{aligned}
g^{(t)}
&=\frac1{b_{\rm PDE}}
\sum_{i\in B_{\rm PDE}^{(t)}}
\bigl(\Delta u_{\theta^{(t)}}(x_i)-f(x_i)\bigr)
\nabla_\theta\Delta u_{\theta^{(t)}}(x_i)\\
&\quad+\frac{\alpha_{\rm BC}}{b_{\rm BC}}
\sum_{j\in B_{\rm BC}^{(t)}}
\bigl(u_{\theta^{(t)}}(z_j)-\varphi(z_j)\bigr)
\nabla_\theta u_{\theta^{(t)}}(z_j).
\end{aligned}
\end{equation}
By uniform sampling, $g^{(t)}$ is a conditionally unbiased estimator of the full gradient $\nabla L(\theta^{(t)})$.

To characterize the effects of minibatch sampling on the PDE and boundary residuals, we define
\begin{equation}
\label{eq:pinn-rho}
\rho_b
=1+\frac1{\lambda_0}
\max\left\{
d^2\frac{n_{\rm PDE}-b_{\rm PDE}}
{b_{\rm PDE}(n_{\rm PDE}-1)},
\;
\alpha_{\rm BC}\frac{n_{\rm BC}-b_{\rm BC}}
{b_{\rm BC}(n_{\rm BC}-1)}
\right\},
\end{equation}
where $b=(b_{\rm PDE},b_{\rm BC})$ denotes the pair of minibatch sizes. The factor $d^2$ arises from the derivatives in the PDE residuals, while the corresponding bounds for the boundary residuals does not rely on $d$.

\begin{theorem}[Poisson PINN with minibatches]
\label{thm:pinn-stoch}
Let $\delta\in(0,1)$ and $n_{\rm PDE}, n_{\rm BC} \geq 2$,
$1\leq b_{\rm PDE}\leq n_{\rm PDE}$,
$1\leq b_{\rm BC}\leq n_{\rm BC}$.
Consider Adam with the minibatch gradients~\eqref{eq:pinn-minibatch-gradients}, the damping parameter and stepsize~\eqref{eq:adam-scaling}, and any fixed $\beta_1\in[0,1)$.
Suppose that
\begin{equation}
\label{eq:pinn-stoch-damping}
\bar\epsilon\gtrsim
d\sqrt{
\frac1\delta
\left(
d^2+\log^2\frac{md^2}{\delta\lambda_0}
\right)
\log\frac2\delta
}.
\end{equation}
Choose fixed $\beta_2\in(0,1)$ and $\bar\eta>0$ such that
\begin{equation}
\label{eq:pinn-stoch-parameters}
0<1-\beta_2\lesssim
\frac{1-\beta_1}{
\rho_b
\log\left(
\frac{\rho_b d^2}
{(1-\beta_1)\lambda_0\delta}
\right)},
\qquad
\bar\eta\asymp\frac{1-\beta_2}{d^2}.
\end{equation}
Suppose further that the network width satisfies
\begin{equation}
\label{eq:pinn-stoch-width}
m\gtrsim
\left(
d^2+\log^2\frac{md^2}{\delta\lambda_0}
\right)
\left(
\frac1{\lambda_0}\log\frac N\delta
+\frac{d^2}{\delta\lambda_0^2}\log\frac2\delta
\right).
\end{equation}
Then, with probability at least $1-\delta/2$ over the
random initialization~\eqref{eq_init}, there exists an event
$\mathcal G$ associated with minibatch sampling such that
\begin{equation}
\label{eq:pinn-stoch-rate}
\begin{gathered}
\mathbb P(\mathcal G\mid\theta^{(0)})
\geq1-\delta/2,\\
\E\left[
L(\theta^{(t)})\mathbf1_{\mathcal G}
\mid\theta^{(0)}
\right]
\leq
C(1-c\bar\eta\lambda_0)^tL(\theta^{(0)}),
\qquad t\geq0.
\end{gathered}
\end{equation}
Here, $c,C>0$ are universal constants. In particular, the joint probability of successful initialization and minibatch sampling is at least $1-\delta$.

Moreover, on the high-probability event $\mathcal G$, the parameter sequence $\{\theta^{(t)}\}$ converges almost surely to a global minimizer $\theta^{\star}$ satisfying $L(\theta^{\star})=0$.
\end{theorem}

\begin{proof}[Proof sketch]
The complete proof is provided in Appendix~\ref{app:pinn-stoch-proof}. We follow the stochastic Adam analysis of Theorem~\ref{thm:reg-stoch}, with additional regularity estimates for the PDE residuals.

As in the deterministic PINN analysis, the initialization bounds give
\[
K(\theta^{(0)})\succeq\frac{3\lambda_0}{4}I,
\qquad
L_0:=L(\theta^{(0)})
\lesssim d^2\log\frac{2}{\delta}.
\]
We introduce an auxiliary network with smooth cutoffs on the output weights and hidden-weight movement from initialization. The auxiliary network agrees with the original PINN whenever
$|a_r|\leq2$ and
$\|w_r-w_r^{(0)}\|\leq1$ for all $r$.
Its residual map is locally Lipschitz, and the associated Gram matrix remains uniformly positive definite whenever
\[
\|\theta-\theta^{(0)}\|
\leq\mathcal R
:=c_0\sqrt{\frac{m\lambda_0}{\ell_m}},
\]
where $\ell_m$ is defined in~\eqref{eq_l_m}.
We analyze the auxiliary process and show that its iterates agree with those of the original PINN on a suitable high-probability event.

Next, the independence of the PDE and boundary minibatches allows us to bound their conditional variances separately. As in the stochastic regression analysis, we obtain
\[
\E_t\|g^{(t)}-G^{(t)}\|^2
\lesssim(\rho_b-1)\|G^{(t)}\|^2,
\]
where $G^{(t)}:=\nabla L(\theta^{(t)})$ and $\E_t$ denotes conditional expectation given the sampling history.
To control the adaptive preconditioner, let $U^{(t)}$ be the bias-corrected moving average of the sampled, weighted squared residuals. Using a common scalar bound, we obtain
\[
\widehat v_j^{(t)}
\lesssim \frac{\ell_m}{m} U^{(t)}
\quad\text{for all }j,
\qquad
\sum_j\widehat v_j^{(t)}
\lesssim d^2U^{(t)}.
\]
The first bound controls individual coordinates of the adaptive preconditioner, while the second controls the sum of the second-moment estimates. In particular, the latter depends on $d^2$ rather than $\ell_m$, avoiding additional logarithmic factors in the subsequent descent analysis.

We use the same virtual iterate $Z^{(t)}$ as in~\eqref{eq5} and construct the Lyapunov function
\[
\mathcal E^{(t)}
=L(Z^{(t)})
+C_0\frac{\bar\eta\nu}{\gamma^2}
\|\widehat s^{(t-1)}\|^2,
\]
where $\gamma=1-\beta_1$ and $\nu=1-\beta_2$.
Compared with regression, the PINN analysis requires additional control of the residuals along the segment connecting the actual and virtual iterates, as well as all possible virtual update segments. Under the network width condition, we establish a uniform Hessian bound of order $d^2$ along these segments. Combining this bound with the conditional descent analysis from the regression setting, we obtain
\[
\E_t[\mathcal E^{(t+1)}]
\leq(1-c\bar\eta\lambda_0)\mathcal E^{(t)},
\]
provided that the required energy, second-moment, and parameter movement bounds hold.

Finally, we apply the stopping-time argument from the stochastic regression analysis, using the PINN-specific regularity bounds to control the training trajectory. As before, the energy descent estimate and martingale maximal inequalities control the cumulative parameter movement, while the bias-corrected moment weights provide neuron-wise bounds. Together, these estimates yield an event $\mathcal G$ with conditional probability at least $1-\delta/2$, on which the auxiliary and original Adam iterates coincide for all iterations. Applying the conditional Lyapunov contraction before the stopping time then results in the convergence bound~\eqref{eq:pinn-stoch-rate}.
\end{proof}

\paragraph{Discussion.}
The factor $\rho_b$ characterizes the relative variance of the stochastic gradient arising from the PDE and boundary minibatches. As the minibatch sizes increase, the corresponding sampling variance decreases and vanishes under full-batch sampling. The factor $d^2$ in the PDE contribution arises from the derivative bounds for the Laplacian features. For the boundary contribution, only one factor of $\alpha_{\rm BC}$ remains, since the other is absorbed into the weighted loss when bounding the variance. When both minibatches are sampled in full, Theorem~\ref{thm:pinn-det} applies with arbitrary fixed $\beta_2\in[0,1)$. The stochastic theorem also allows one group to be sampled in full while the other is sampled stochastically.

The conditions in~\eqref{eq:pinn-stoch-parameters} serve different purposes. As in stochastic regression, choosing $1-\beta_2$ sufficiently small controls the dependence between the current stochastic gradient and the adaptive preconditioner, as well as the accumulation of second-moment estimates. The logarithmic factor arises from controlling these estimates uniformly over iterations. The condition $\bar\eta\asymp(1-\beta_2)/d^2$ further accounts for the additional $d^2$ factor in the PINN smoothness bounds. These parameter conditions do not introduce any additional dependence on the network width $m$.

Compared with the deterministic result in Theorem~\ref{thm:pinn-det}, the stochastic analysis requires a threshold of order $d^2\delta^{-1}\log(2/\delta)$ to control the stochastic training trajectory. This leads to an additional factor of $\delta^{-1/2}$ in the damping requirement~\eqref{eq:pinn-stoch-damping} and $\delta^{-1}$ in the parameter movement term of the width requirement~\eqref{eq:pinn-stoch-width}. As in the deterministic PINN analysis, controlling the maximum over the Gaussian-initialized neurons introduces logarithmic dependence on $m$ in the damping and width requirements through $\ell_m$. In contrast, the averaged feature bounds allow us to control the stepsize and variance estimates without additional logarithmic dependence on $m$. Consequently, the conditions on $\beta_2$ and $\bar\eta$ in~\eqref{eq:pinn-stoch-parameters} do not explicitly depend on the network width.
As in Theorem~\ref{thm:reg-stoch}, the convergence bound~\eqref{eq:pinn-stoch-rate} holds in expectation restricted to a single high-probability event $\mathcal G$ for all iterations. The probability statement accounts separately for initialization and minibatch sampling.

Table~\ref{tab:pinn-optimizer-comparison} compares the convergence guarantees for GD, SGD, and Adam in training PINNs. By adapting our analysis to GD and SGD, we obtain sharper width requirements and improved convergence rates compared with existing results~\cite{gao2023gradient,jin2026convergence}. In particular, the existing SGD analysis~\cite{jin2026convergence} does not explicitly characterize how the minibatch size affects the stepsize conditions and convergence rates. Our GD and SGD results, established in Appendix~\ref{app:pinn-gd-sgd}, make explicit the dependence on the input dimension and minibatch size. Moreover, our analysis shows that Adam achieves comparable width requirements and convergence rates to GD and SGD, up to logarithmic factors in the stochastic setting, despite the additional challenges introduced by momentum and adaptive preconditioning.

\begin{table}[t!]
\centering
\small
\renewcommand{\arraystretch}{1.5}
\begin{tabular*}{\textwidth}{
  @{\extracolsep{\fill}}lccc@{}
}
\toprule
Method
& Sufficient width
& Stepsize
& Convergence rate \\
\midrule

GD~\cite[Theorem 4.5]{gao2023gradient}
& $\widetilde{\Omega}
   \bigl(N^2\delta^{-3}\lambda_0^{-4}\bigr)$
& $h\asymp\lambda_0$
& $\lambda_0^2$ \\

SGD~\cite[Theorem 3.1]{jin2026convergence}
& $\widetilde{\Omega}
   \bigl(d^6\lambda_0^{-2}
   +d^{12}\delta^{-2}\lambda_0^{-4}\bigr)$
& $h=\widetilde{\Theta}
   \bigl(\lambda_0/d^{9/2}\bigr)$
& $\widetilde{\Theta}
   \bigl(\lambda_0^2/d^{9/2}\bigr)$ \\

\midrule

GD (Proposition~\ref{prop:pinn-gd-sgd})
& $\widetilde{\Omega}
   \bigl(d^2\lambda_0^{-1}
   +d^4\lambda_0^{-2}\bigr)$
& $h\asymp d^{-2}$
& $\lambda_0/d^2$ \\

SGD (Proposition~\ref{prop:pinn-gd-sgd})
& $\widetilde{\Omega}
   \bigl(d^2\lambda_0^{-1}
   +d^4\delta^{-1}\lambda_0^{-2}\bigr)$
& $h\asymp(d^2\rho_b)^{-1}$
& $\lambda_0/(d^2\rho_b)$ \\

Full Adam (Theorem~\ref{thm:pinn-det})
& $\widetilde{\Omega}
   \bigl(d^2\lambda_0^{-1}
   +d^4\lambda_0^{-2}\bigr)$
& $\bar\eta\asymp(1-\beta_1)/d^2$
& $(1-\beta_1)\lambda_0/d^2$ \\

Minibatch Adam (Theorem~\ref{thm:pinn-stoch})
& $\widetilde{\Omega}
   \bigl(d^2\lambda_0^{-1}
   +d^4\delta^{-1}\lambda_0^{-2}\bigr)$
& $\bar\eta=\widetilde{\Theta}
   \bigl((1-\beta_1)/(d^2\rho_b)\bigr)$
& $\widetilde{\Theta}
   \bigl((1-\beta_1)\lambda_0/(d^2\rho_b)\bigr)$ \\
\bottomrule
\end{tabular*}
\caption{
Comparison of network width requirements, stepsize, and convergence rate for GD, SGD, and Adam in training PINNs for the Poisson equation. The convergence rates are expressed as contraction gaps. The notation $\widetilde{\Omega}$ and $\widetilde{\Theta}$ hides logarithmic factors. Here, $\rho_b$ is the minibatch sampling factor defined in \eqref{eq:pinn-rho}, $h$ is the stepsize for GD and SGD, and $\bar\eta=\eta/\epsilon$ is the normalized stepsize for Adam.
The analysis of~\cite{gao2023gradient} treats the input dimension $d$ as fixed and does not explicitly track its dependence. The analysis of~\cite{jin2026convergence} treats the stochastic gradient variance bound as a constant and does not explicitly characterize the dependence on minibatch size.
}
\label{tab:pinn-optimizer-comparison}
\end{table}

\section{Conclusion}

In this paper, we study the global convergence of Adam in training two-layer neural networks for regression and physics-informed learning. Using the neural tangent kernel framework, we establish linear convergence guarantees for sufficiently wide networks with both full-batch and minibatch gradients in Adam. Our analysis addresses the dependence introduced by historical momentum and adaptive preconditioning, which prevents a direct application of existing GD and SGD analyses. In particular, we develop Lyapunov-based descent arguments, a virtual-iterate construction, and stochastic trajectory control to establish convergence without modifying the standard Adam updates. We further extend these results to physics-informed neural networks (PINNs) for the Poisson equation, where differential residuals require additional regularity estimates and more restrictive conditions on the network width and algorithmic parameters. These results provide a theoretical explanation for the convergence behavior of Adam in overparameterized neural networks.

Several directions remain for future research. First, an interesting direction is to study the global convergence of quasi-Newton methods, particularly L-BFGS~\cite{liu1989limited}, for training PINNs. In practice, L-BFGS is widely used in scientific machine learning~\cite{raissi2019physics}, often following an initial training stage with Adam~\cite{rathore2024challenges}. Unlike GD, SGD, and Adam, L-BFGS uses curvature information through approximate inverse-Hessian updates. Recent convergence results for natural gradient descent in PINNs suggest that incorporating additional geometric information may improve convergence rates~\cite{xu2026fast}. It would therefore be interesting to investigate whether similar improvements can be established for L-BFGS.

Second, although we establish global convergence guarantees for Adam, our theoretical results do not show an advantage over SGD in terms of convergence rates or network width requirements. This differs from empirical observations in some deep learning applications, where Adam-type methods often outperform SGD. Understanding this difference is an important direction for future research. In particular, it would be interesting to study how data distributions, gradient heterogeneity, and network architectures affect the convergence of Adam. Such an analysis may help identify conditions under which Adam achieves faster convergence than SGD.

Finally, our analysis is limited to two-layer neural networks and requires specific conditions on the damping parameter, momentum parameters, and stepsize, particularly in the stochastic setting. Relaxing these conditions and establishing stronger stochastic convergence guarantees are important directions for future research. Another direction is to extend the analysis to deeper neural networks and more general PDEs, where the NTK properties and optimization dynamics are more difficult to analyze. These extensions may provide a broader theoretical understanding of optimization methods for training neural networks.

\newpage 
\bibliographystyle{plain}
\bibliography{main}

\newpage

\appendix

\startcontents[appendices]

\section*{Appendix Contents}
\printcontents[appendices]{}{1}[2]{}

\clearpage

\section{Auxiliary Lemmas}
\label{app:auxiliary-lemmas}

We collect several concentration inequalities used in our convergence analysis of Adam. For completeness, we state the results needed in our proofs and refer to the cited references for their proofs.

\begin{lemma}[Matrix Chernoff inequality {\cite[Theorem~5.1.1]{tropp2015introduction}}]
\label{lem:aux-matrix-chernoff}
Let $X_1,\ldots,X_m$ be independent symmetric random matrices in $\R^{n\times n}$. Suppose that, for some $B>0$,
\[
    0\preceq X_r\preceq BI,
    \qquad r=1,\ldots,m,
\]
almost surely. Define
\[
    \mu_{\min}
    =\lambda_{\min}\left(
    \sum_{r=1}^m\E[X_r]
    \right)>0.
\]
Then, for any $\alpha\in(0,1)$,
\begin{equation}
\label{eq:aux-matrix-chernoff}
    \mathbb P\left(
    \lambda_{\min}\left(\sum_{r=1}^m X_r\right)
    \leq(1-\alpha)\mu_{\min}
    \right)
    \leq n\exp\left(
    -\frac{\alpha^2\mu_{\min}}{2B}
    \right).
\end{equation}
\end{lemma}

\begin{lemma}[Bounded-differences inequality {\cite{mcdiarmid1989method}}]
\label{lem:aux-bounded-differences}
Let $Z_1,\ldots,Z_m$ be independent random variables taking values in $\mathcal Z_1,\ldots,\mathcal Z_m$, respectively. Let $F:\prod_{r=1}^m\mathcal Z_r\to\R$ be a measurable function satisfying $\E|F(Z_1,\ldots,Z_m)|<\infty$. Suppose there exist constants $b_1,\ldots,b_m\geq0$, with $\sum_{r=1}^m b_r^2>0$, such that
\[
    \left|
    F(z_1,\ldots,z_r,\ldots,z_m)
    -F(z_1,\ldots,z_r',\ldots,z_m)
    \right|
    \leq b_r
\]
for every $r$ and all admissible arguments. Writing $Z=(Z_1,\ldots,Z_m)$, we have, for any $u>0$,
\begin{equation}
\label{eq:aux-bounded-differences}
    \mathbb P\left(
    F(Z)-\E[F(Z)]\geq u
    \right)
    \leq
    \exp\left(
    -\frac{2u^2}{\sum_{r=1}^m b_r^2}
    \right).
\end{equation}
\end{lemma}

\begin{lemma}[Conditional Bernstein inequality {\cite[Theorem~1.1]{tropp2011freedman}}]
\label{lem:aux-martingale-bernstein}
Let $(\mathcal F_s)_{s=0}^{N+1}$ be a filtration, and let $X_s$ be $\mathcal F_{s+1}$-measurable with $\E[X_s\mid\mathcal F_s]=0$ for $s=0,\ldots,N$. Suppose there exist deterministic constants $B,V>0$ such that
\[
|X_s|\leq B \quad\text{a.s.},\qquad
\sum_{s=0}^N\E[X_s^2\mid\mathcal F_s]\leq V
\quad\text{a.s.}
\]
Then, for any $u>0$,
\begin{equation}
\label{eq:aux-martingale-bernstein}
\mathbb P\left(
\sum_{s=0}^N X_s\geq u
\right)
\leq
\exp\left(
-\frac{u^2}{2(V+Bu/3)}
\right).
\end{equation}
\end{lemma}

\begin{proof}
The lemma is a direct result of Freedman's inequality; see \cite[Theorem~1.1]{tropp2011freedman}. We include a short proof for completeness.

For any $0<h<3/B$, expanding the exponential and using the conditional centering of $X_s$, together with $k!\geq 2\cdot 3^{k-2}$ for $k\geq 2$, yields
\[
\E[e^{hX_s}\mid\mathcal F_s]
\leq
1+\frac{h^2\E[X_s^2\mid\mathcal F_s]}
{2(1-hB/3)}
\leq
\exp\left(
\frac{h^2\E[X_s^2\mid\mathcal F_s]}
{2(1-hB/3)}
\right),
\]
where we have used
\[
\left|\E[X_s^k\mid\mathcal F_s]\right|
\leq B^{k-2}\E[X_s^2\mid\mathcal F_s],
\qquad k\geq 2.
\]
Consequently, the process
\[
M_t=
\exp\left(
h\sum_{s=0}^{t-1}X_s
-\frac{h^2}{2(1-hB/3)}
\sum_{s=0}^{t-1}\E[X_s^2\mid\mathcal F_s]
\right),
\qquad t=0,\ldots,N+1,
\]
is a nonnegative supermartingale with $M_0=1$.
Applying Markov's inequality and the assumed bound on the accumulated conditional variance gives
\[
\mathbb P\left(
\sum_{s=0}^N X_s\geq u
\right)
\leq
\exp\left(
-hu+\frac{h^2V}{2(1-hB/3)}
\right).
\]
Choosing $h=u/(V+Bu/3)$ completes the proof.
\end{proof}

\begin{lemma}[Maximal inequalities~\cite{williams1991}]
\label{lem:aux-doob}
Let $(\mathcal F_t)_{t\geq0}$ be a filtration.
\begin{enumerate}
\item Let $(X_t)_{t\geq t_0}$ be a nonnegative supermartingale. Then, for any $a>0$,
\[
\mathbb P\left(
\sup_{t\geq t_0}X_t\geq a
\right)
\leq \frac{\E[X_{t_0}]}{a}.
\]

\item Let $(M_t)_{t=0}^N$ be a square-integrable Euclidean-valued martingale with $M_0=0$. Then,
\[
\E\left[\max_{0\leq t\leq N}\|M_t\|^2\right]
\leq 4\E\|M_N\|^2
=4\sum_{t=0}^{N-1}\E\|M_{t+1}-M_t\|^2.
\]
If $\sum_{t=0}^{\infty}\E\|M_{t+1}-M_t\|^2<\infty$, then
\[
\E\left[\sup_{t\geq0}\|M_t\|^2\right]
\leq4\sum_{t=0}^{\infty}
\E\|M_{t+1}-M_t\|^2.
\]
\end{enumerate}
\end{lemma}

\begin{proof}
For the first assertion, let $\tau$ be the first time $X_t$ reaches the threshold $a$, truncated at a fixed horizon $N$. By the nonnegativity and supermartingale property of $X_t$, we have
\[
a\,\mathbb P\left(
\max_{t_0\leq t\leq N}X_t\geq a
\right)
\leq \E[X_{t_0}].
\]
Letting $N\to\infty$ and, if necessary, replacing $a$ by $a-\varepsilon$ and then letting $\varepsilon\downarrow0$ proves the first assertion.

For the second assertion, let $Z_t=\|M_t\|$, which is a nonnegative submartingale. Define
\[
Z^*=\max_{0\leq t\leq N}Z_t.
\]
Applying the stopping-time argument at the first crossing of $a$ gives
\[
a\,\mathbb P(Z^*\geq a)
\leq \E\left[Z_N\mathbf{1}_{\{Z^*\geq a\}}\right].
\]
Integrating over $a>0$ and applying the Cauchy--Schwarz inequality, we obtain
\[
\begin{aligned}
\E[(Z^*)^2]
&\leq 2\E[Z_NZ^*]\\
&\leq 2\big(\E[Z_N^2]\big)^{1/2}
       \big(\E[(Z^*)^2]\big)^{1/2}.
\end{aligned}
\]
Consequently,
\[
\E[(Z^*)^2]\leq 4\E[Z_N^2].
\]
The equality involving the sum of increment variances follows from the orthogonality of martingale increments. The infinite-horizon result then follows by monotone convergence under the stated summability condition.
\end{proof}

\section{Proof of Theorem~\ref{thm:reg-det}}
\label{app:reg-det-proof}

Recall that $\theta_r=(a_r,w_r)\in\R^{d+2}$ denote the parameters of the $r$-th neuron. For convenience, define
\begin{equation}
\label{eq:det-app-normalization}
    \gamma=1-\beta_1,
    \qquad
    P^{(t)}=\epsilon D^{(t)}
    =\operatorname{diag}\left(
    \frac{\epsilon}{\sqrt{\widehat v^{(t)}}+\epsilon}
    \right).
\end{equation}
By definition, $0\prec P^{(t)}\preceq I$, and the Adam update can be rewritten as
\[
\theta^{(t+1)}
=\theta^{(t)}-\bar\eta P^{(t)}\widehat s^{(t)}.
\]
Throughout the proof, $C,c>0$ denote universal constants that may change from line to line.

We first establish several lemmas on initialization, loss function regularity, and the convergence of Adam. We then combine these results to prove Theorem~\ref{thm:reg-det} by induction.

\subsection{Positive Definiteness of the Limiting Gram Matrix}
\label{app:gram_pd}

Let $s_{ir}=w_r^\top\widetilde{x}_i$ and $\theta_r=(a_r,w_r)$. Direct differentiation gives
\begin{align}
\nabla_{\theta_r}f_\theta(x_i)
&=\frac{1}{\sqrt{m}}
\begin{pmatrix}
\sigma(s_{ir})\\
a_r\sigma'(s_{ir})\widetilde{x}_i
\end{pmatrix},
\label{eq:reg-feature}\\
\nabla_{\theta_r}^2f_\theta(x_i)
&=\frac{1}{\sqrt{m}}
\begin{pmatrix}
0 & \sigma'(s_{ir})\widetilde{x}_i^\top\\
\sigma'(s_{ir})\widetilde{x}_i &
a_r\sigma''(s_{ir})\widetilde{x}_i\widetilde{x}_i^\top
\end{pmatrix}.
\label{eq:reg-feature-hessian}
\end{align}
Consequently, the Gram matrix $K(\theta)=J(\theta)J(\theta)^\top$ can be expressed as
\begin{equation}
\label{eq:reg-kernel-summand}
K(\theta)_{ij}
=\frac{1}{nm}\sum_{r=1}^m
\left[
\sigma(s_{ir})\sigma(s_{jr})
+a_r^2\sigma'(s_{ir})\sigma'(s_{jr})
\widetilde{x}_i^\top\widetilde{x}_j
\right].
\end{equation}
Both terms define positive semidefinite matrices. Taking expectations with respect to the initialization~\eqref{eq_init}, we obtain
\begin{align}
(K_\infty^{(a)})_{ij}
&=\frac{1}{n}\E_w\left[
\sigma(w^\top\widetilde{x}_i)
\sigma(w^\top\widetilde{x}_j)
\right],\nonumber\\
(K_\infty^{(w)})_{ij}
&=\frac{\widetilde{x}_i^\top\widetilde{x}_j}{n}
\E_w\left[
\sigma'(w^\top\widetilde{x}_i)
\sigma'(w^\top\widetilde{x}_j)
\right],\nonumber\\
K_\infty
&=K_\infty^{(a)}+K_\infty^{(w)}
\succeq K_\infty^{(w)},
\qquad
\lambda_0=\lambda_{\min}(K_\infty).
\label{eq:reg-limiting-kernel}
\end{align}

\begin{lemma}
\label{app:lem-reg-positivity}
Both $K_\infty^{(w)}$ and $K_\infty$ are positive definite.
\end{lemma}

\begin{proof}
Since $K_\infty\succeq K_\infty^{(w)}$, it suffices to show that $K_\infty^{(w)}$ is positive definite. For any $c\in\R^n$, we have
\[
c^\top K_\infty^{(w)}c
=
\frac{1}{n}\E_w\left[
\left\|
\sum_{i=1}^n
c_i\sigma'(w^\top\widetilde{x}_i)\widetilde{x}_i
\right\|^2
\right].
\]
Suppose that this quadratic form is zero. Since the integrand is nonnegative and continuous in $w$, and the Gaussian distribution has full support on $\R^{d+1}$, it follows that
\[
\sum_{i=1}^n
c_i\sigma'(w^\top\widetilde{x}_i)\widetilde{x}_i=0,
\qquad
\text{for all }w\in\R^{d+1}.
\]

We next show that $c_k=0$ for each $k=1,\ldots,n$. Since the training inputs are pairwise distinct, there exists $v\in\R^d$ such that $v^\top x_1,\ldots,v^\top x_n$ are pairwise distinct. Indeed, any vector outside the finite union of proper hyperplanes
\[
\bigcup_{i<j}
\left\{
v\in\R^d:
v^\top(x_i-x_j)=0
\right\}
\]
satisfies this property.
Fix any $k\in [n]$ and define
\[
w_A=(Av,-Av^\top x_k)\in\R^{d+1},
\qquad A>0.
\]
By the definition $\widetilde{x}_i=(x_i,1)/\sqrt{2}$, we have
\[
w_A^\top\widetilde{x}_i
=
\frac{A}{\sqrt{2}}v^\top(x_i-x_k).
\]
For $i=k$, this expression equals zero, while for $i\neq k$, it tends to either $+\infty$ or $-\infty$ as $A\to\infty$. Since $\sigma=\tanh$ satisfies $\sigma'(0)=1$ and $\sigma'(s)\to0$ as $|s|\to\infty$, substituting $w=w_A$ into the preceding identity and letting $A\to\infty$ yields
\[
c_k\widetilde{x}_k=0.
\]
The last coordinate of $\widetilde x_k$ is $1/\sqrt2$, so $c_k=0$. Since $k$ was arbitrary, we conclude that $c=0$.
\end{proof}

\subsection{Key Lemmas}

\begin{lemma}[Initialization]
\label{lem:det-app-init}
There exist universal constants $C_{\rm init},C_0>0$ such that, for any $\delta\in(0,1)$, if
\[
m\geq \frac{C_{\rm init}}{\lambda_0}\log\frac{2n}{\delta},
\]
then, with probability at least $1-\delta$ over the random initialization~\eqref{eq_init}, we have
\begin{equation}
\label{eq:det-app-init}
K(\theta^{(0)})\succeq \frac34\lambda_0 I,
\qquad
L(\theta^{(0)})\leq S_0,
\qquad
S_0=C_0\log\frac2\delta.
\end{equation}
\end{lemma}

\begin{proof}
For any $w\in\R^{d+1}$, define the feature vectors
\[
\phi_i(w)=
\begin{pmatrix}
\sigma(w^\top\widetilde{x}_i)\\
\sigma'(w^\top\widetilde{x}_i)\widetilde{x}_i
\end{pmatrix},
\qquad i=1,\ldots,n,
\]
and the associated Gram matrix $H(w)\in\R^{n\times n}$ with entries
\[
H(w)_{ij}=\frac1n\phi_i(w)^\top\phi_j(w).
\]
Since $(a_r^{(0)})^2=1$, it follows that
\[
K(\theta^{(0)})
=\frac1m\sum_{r=1}^m H(w_r^{(0)}),
\qquad
\E[H(w_r^{(0)})]=K_\infty.
\]
The matrices $H(w_r^{(0)})$ are independent and positive semidefinite. Let $u=\tanh^2(s)\in[0,1]$. Since $\|\widetilde{x}_i\|\leq1$, we have
\[
\tanh^2(s)+\tanh'(s)^2\|\widetilde{x}_i\|^2
\leq u+(1-u)^2\leq1.
\]
Consequently, $\|\phi_i(w)\|^2\leq1$ for all $i=1,\ldots,n$, which implies
\[
\operatorname{trace}(H(w))\leq1,
\qquad
0\preceq H(w)\preceq I.
\]
Applying Lemma~\ref{lem:aux-matrix-chernoff} with $X_r=H(w_r^{(0)})$, $B=1$, $\mu_{\min}=m\lambda_0$, and $\alpha=1/4$ gives
\[
\mathbb P\left(
\lambda_{\min}(K(\theta^{(0)}))
<\frac34\lambda_0
\right)
\leq n\exp\left(-\frac{m\lambda_0}{32}\right).
\]
Under the width condition on $m$, this probability is at most $\delta/2$.

Next, we bound the initial loss. Define
\[
F^{(0)}
=\frac1{\sqrt n}
\big(f_{\theta^{(0)}}(x_1),\ldots,
f_{\theta^{(0)}}(x_n)\big)^\top
=\frac1{\sqrt m}\sum_{r=1}^m
a_r^{(0)}\Phi(w_r^{(0)}),
\]
where
\[
\Phi(w)=\frac1{\sqrt n}
\big(\sigma(w^\top\widetilde{x}_i)\big)_{i=1}^n.
\]
Since $\|\Phi(w)\|\leq1$ and the output weights $a_r^{(0)}$ are independent and centered, we obtain
\[
\E\|F^{(0)}\|^2\leq1,
\qquad
\E\|F^{(0)}\|\leq1.
\]
Moreover, replacing any one of the independent pairs $(a_r^{(0)},w_r^{(0)})$ changes $\|F^{(0)}\|$ by at most $2/\sqrt m$. Applying Lemma~\ref{lem:aux-bounded-differences} with $b_r=2/\sqrt m$ and $u=\sqrt{2\log(2/\delta)}$ gives
\[
\mathbb P\left(
\|F^{(0)}\|>
1+\sqrt{2\log(2/\delta)}
\right)
\leq\frac\delta2.
\]
Since $\|(y_i/\sqrt n)_{i=1}^n\|\leq1$, on the complementary event we have
\[
L(\theta^{(0)})
\leq\frac12
\left(2+\sqrt{2\log(2/\delta)}\right)^2
\leq C_0\log\frac2\delta.
\]
Combining the two probability bounds by a union bound completes the proof.
\end{proof}

\begin{lemma}[Regularity and Gram matrix stability]
\label{lem:det-app-geometry}
Define
\[
\mathcal S=\{\theta:\ |a_r|\leq 2,\ 1\leq r\leq m\}.
\]
There exist universal constants $H\geq 1$, $C_J,C_g>0$, and $c_0>0$ such that, for any $\theta,\vartheta\in\mathcal S$,
\begin{align}
&\|J(\theta)\|\leq 2,
\qquad
\|J(\theta)-J(\vartheta)\|
\leq\frac{C_J}{\sqrt m}\|\theta-\vartheta\|,
\label{eq:det-app-jacobian}\\
&\|\nabla^2L(\theta)\|\leq H,
\qquad
\|\nabla_{\theta_r}L(\theta)\|
\leq C_g\sqrt{\frac{L(\theta)}{m}},
\quad 1\leq r\leq m.
\label{eq:det-app-smoothness}
\end{align}
On the initialization event~\eqref{eq:det-app-init}, let
\[
\mathcal R=c_0\sqrt{m\lambda_0}.
\]
Then, every $\theta\in\mathcal S$ satisfying
$\|\theta-\theta^{(0)}\|\leq\mathcal R$ obeys
\begin{equation}
\label{eq:det-app-local-kernel}
K(\theta)\succeq\frac{\lambda_0}{4}I,
\qquad
\|\nabla L(\theta)\|^2
\geq\frac{\lambda_0}{2}L(\theta).
\end{equation}
\end{lemma}

\begin{proof}
Recall from \eqref{eq:reg-feature} and \eqref{eq:reg-feature-hessian} that, with $s_{ir}=w_r^\top\widetilde{x}_i$,
\begin{align*}
\nabla_{\theta_r}f_\theta(x_i)
&=\frac{1}{\sqrt m}
\begin{pmatrix}
\sigma(s_{ir})\\
a_r\sigma'(s_{ir})\widetilde{x}_i
\end{pmatrix},\\
\nabla_{\theta_r}^2f_\theta(x_i)
&=\frac{1}{\sqrt m}
\begin{pmatrix}
0 & \sigma'(s_{ir})\widetilde{x}_i^\top\\
\sigma'(s_{ir})\widetilde{x}_i &
a_r\sigma''(s_{ir})
\widetilde{x}_i\widetilde{x}_i^\top
\end{pmatrix}.
\end{align*}
Moreover, the mixed second derivatives between different neurons vanish. Since $|\sigma|\leq 1$, $|\sigma'|\leq 1$, $|\sigma''|\leq 2$, and $\|\widetilde{x}_i\|\leq 1$, we have, for any $\theta\in\mathcal S$,
\[
\|\nabla_{\theta_r}f_\theta(x_i)\|
\leq\frac{2}{\sqrt m},
\qquad
\|\nabla_{\theta_r}^2f_\theta(x_i)\|
\leq\frac{C}{\sqrt m}.
\]
To verify the first bound, let $u=\tanh^2(s)\in[0,1]$, we use
\[
\sigma(s)^2
+a_r^2\sigma'(s)^2\|\widetilde{x}_i\|^2
\leq u+4(1-u)^2\leq 4.
\]
Summing the squared first-derivative bounds over all samples and neurons gives
\[
\|J(\theta)\|
\leq\|J(\theta)\|_{\rm F}\leq 2.
\]
Since $\mathcal S$ is convex, integrating the second-derivative bound along the line segment joining $\theta$ and $\vartheta$ gives
\[
\|J_{i,r}(\theta)-J_{i,r}(\vartheta)\|
\leq\frac{C}{\sqrt{nm}}
\|\theta_r-\vartheta_r\|,
\]
where $J_{i,r}$ denotes the block of row $i$ of $J$ corresponding to neuron $r$. Squaring and summing over $i$ and $r$ establishes the second inequality in \eqref{eq:det-app-jacobian}.

Next, we establish the bounds in \eqref{eq:det-app-smoothness}. By the Cauchy--Schwarz inequality,
\[
\begin{aligned}
\|\nabla_{\theta_r}L(\theta)\|
&\leq\frac{2}{\sqrt m}
\frac{1}{n}\sum_{i=1}^n
|f_\theta(x_i)-y_i|\\
&\leq 2\sqrt{\frac{2L(\theta)}{m}}.
\end{aligned}
\]
For the Hessian of the loss function, we have
\[
\nabla^2L(\theta)
=J(\theta)^\top J(\theta)
+\sum_{i=1}^n R_i(\theta)\nabla^2R_i(\theta).
\]
The second term is block diagonal with respect to the neuron parameters, and each block has spectral norm bounded by
\[
\frac{C}{\sqrt{nm}}
\sum_{i=1}^n|R_i(\theta)|
\leq\frac{C}{\sqrt m}\|R(\theta)\|.
\]
Consequently,
\[
\|\nabla^2L(\theta)\|
\leq 4+\frac{C}{\sqrt m}\|R(\theta)\|
\leq H,
\]
where the last inequality follows from
$|f_\theta(x_i)|\leq 2\sqrt m$ and
$\|R(\theta)\|\leq 2\sqrt m+1$.
Thus, the loss function is uniformly smooth on $\mathcal S$, with a smoothness constant that does not depend on the network width or the hidden weights.

Finally, we establish the lower bound on the Gram matrix. For any unit vector $z\in\R^n$, the initialization bound~\eqref{eq:det-app-init} and the Jacobian Lipschitz estimate~\eqref{eq:det-app-jacobian} imply
\[
\begin{aligned}
\|J(\theta)^\top z\|
&\geq
\|J(\theta^{(0)})^\top z\|
-\|J(\theta)-J(\theta^{(0)})\|\\
&\geq
\sqrt{\frac{3\lambda_0}{4}}
-\frac{C_J}{\sqrt m}
\|\theta-\theta^{(0)}\|.
\end{aligned}
\]
Choose $c_0>0$ sufficiently small such that $C_Jc_0\leq 1/4$. Then, whenever $\|\theta-\theta^{(0)}\|\leq\mathcal R$, we obtain
\[
\|J(\theta)^\top z\|
\geq\frac{\sqrt{\lambda_0}}{2}.
\]
Since this holds for every unit vector $z$, it follows that
\[
K(\theta)=J(\theta)J(\theta)^\top
\succeq\frac{\lambda_0}{4}I.
\]
The gradient inequality in \eqref{eq:det-app-local-kernel} then follows from $\nabla L(\theta)=J(\theta)^\top R(\theta)$ and $\|R(\theta)\|^2=2L(\theta)$.
\end{proof}

\begin{lemma}[Moment bounds and parameter movement]
\label{lem:det-app-moments}
Suppose that $\theta^{(0)},\ldots,\theta^{(T)}\in\mathcal S$.
\begin{enumerate}
\item If $L(\theta^{(k)})\leq S$ for all $0\leq k\leq T$, then
\begin{equation}
\label{eq:det-app-metric}
P^{(t)}\succeq
\left(1+\frac{C_g\sqrt S}{\bar\epsilon}\right)^{-1}I,
\qquad 0\leq t\leq T.
\end{equation}
In particular, if $\bar\epsilon\geq C_g\sqrt S$, then $P^{(t)}\succeq I/2$ for all $0\leq t\leq T$ and any $\beta_2\in[0,1)$.

\item Suppose further that
\[
L(\theta^{(k)})\leq q^kL(\theta^{(0)}),
\qquad 0\leq k\leq T,
\]
where $q\in(0,1)$ satisfies $1-q\leq\gamma/2$. Then
\begin{align}
\|\widehat s_r^{(t)}\|
&\leq 2C_g
\sqrt{\frac{L(\theta^{(0)})}{m}}\,q^{t/2},
\qquad 0\leq t\leq T,
\label{eq:det-app-first-moment}\\
\sum_{t=0}^T
\max_{1\leq r\leq m}
\|\theta_r^{(t+1)}-\theta_r^{(t)}\|
&\leq
\frac{4C_g\bar\eta}{1-q}
\sqrt{\frac{L(\theta^{(0)})}{m}}.
\label{eq:det-app-neuron-path}
\end{align}
\end{enumerate}
\end{lemma}

\begin{proof}
By the bias-correction formulas, the first- and second-moment estimates can be written as convex combinations of the historical gradients and their coordinatewise squares, respectively:
\begin{align}
\widehat s^{(t)}
&=\sum_{k=0}^t
\frac{\gamma\beta_1^{t-k}}
{1-\beta_1^{t+1}}\,g^{(k)},
\label{eq:det-app-s-history}\\
\widehat v^{(t)}
&=\sum_{k=0}^t
\frac{(1-\beta_2)\beta_2^{t-k}}
{1-\beta_2^{t+1}}\,g^{(k)\odot2}.
\label{eq:det-app-v-history}
\end{align}
When either momentum parameter is zero, the corresponding bias-corrected moment reduces to the current gradient or its coordinatewise square.

By the block-gradient bound in \eqref{eq:det-app-smoothness}, every coordinate of $g^{(k)}$ is bounded in absolute value by $C_g\sqrt{S/m}$. Since the weights in \eqref{eq:det-app-v-history} are nonnegative and sum to one, we obtain
\[
\widehat v_j^{(t)}
\leq\frac{C_g^2S}{m}
\]
for every coordinate $j$ and all $0\leq t\leq T$. Substituting this bound into \eqref{eq:det-app-normalization} and using $\epsilon=\bar\epsilon/\sqrt m$ proves \eqref{eq:det-app-metric}.

For the second assertion, combining the block-gradient bound with \eqref{eq:det-app-s-history} yields
\[
\|\widehat s_r^{(t)}\|
\leq C_g\sqrt{\frac{L(\theta^{(0)})}{m}}
\sum_{k=0}^t
\frac{\gamma\beta_1^{t-k}}
{1-\beta_1^{t+1}}q^{k/2}.
\]
For $\beta_1>0$, let $z=\beta_1/\sqrt q$. Since $1-q\leq\gamma/2$, we have $z<1$ and $1-z\geq\gamma/2$. Indeed, setting $a=\sqrt q$, we obtain
\[
1-a\leq1-q\leq\frac{\gamma}{2},
\qquad
1-z=\frac{a-\beta_1}{a}\geq\frac{\gamma}{2}.
\]
Consequently, the finite geometric sum satisfies
\[
\begin{aligned}
\sum_{k=0}^t
\frac{\gamma\beta_1^{t-k}}
{1-\beta_1^{t+1}}q^{k/2}
&=q^{t/2}\frac{\gamma}{1-z}
\frac{1-z^{t+1}}{1-\beta_1^{t+1}}\\
&\leq 2q^{t/2},
\end{aligned}
\]
where the last inequality follows from $z\geq\beta_1$. The same bound holds directly when $\beta_1=0$, proving \eqref{eq:det-app-first-moment}.

Finally, since $0\prec P^{(t)}\preceq I$, the Adam update gives
\[
\max_{1\leq r\leq m}
\|\theta_r^{(t+1)}-\theta_r^{(t)}\|
\leq 2C_g\bar\eta
\sqrt{\frac{L(\theta^{(0)})}{m}}\,q^{t/2}.
\]
Summing over $t=0,\ldots,T$ and using
$1-\sqrt q\geq(1-q)/2$
establishes \eqref{eq:det-app-neuron-path}.
\end{proof}

\begin{lemma}[Lyapunov descent and parameter movement]
\label{lem:det-app-energy}
Let $H$ be defined in Lemma~\ref{lem:det-app-geometry}, and suppose that
\[
0<\bar\eta\leq\frac{\gamma}{8H}.
\]
Define
\begin{equation}
\label{eq:det-app-energy}
e^{(t)}=\widehat s^{(t)}-g^{(t)},
\qquad
\mathcal E^{(t)}
=L(\theta^{(t)})
+\frac{4\bar\eta}{\gamma}\|e^{(t)}\|^2,
\qquad
Q^{(t)}=(g^{(t)})^\top P^{(t)}g^{(t)}.
\end{equation}
Suppose that, at iteration $t$,
\[
[\theta^{(t)},\theta^{(t+1)}]\subset\mathcal S,
\qquad
K(\theta^{(t)})\succeq\frac{\lambda_0}{4}I,
\qquad
\frac12I\preceq P^{(t)}\preceq I,
\]
where $[\theta^{(t)},\theta^{(t+1)}]$ denotes the line segment joining $\theta^{(t)}$ and $\theta^{(t+1)}$. Then, with $q=1-\bar\eta\lambda_0/32\in(0,1)$, we have
\begin{align}
\Delta^{(t)}
:=\mathcal E^{(t)}-\mathcal E^{(t+1)}
&\geq\frac{\bar\eta}{4}Q^{(t)}
+\bar\eta\|e^{(t)}\|^2,
\label{eq:det-app-dissipation}\\
\mathcal E^{(t+1)}
&\leq q\mathcal E^{(t)},
\label{eq:det-app-contraction}\\
\|\theta^{(t+1)}-\theta^{(t)}\|^2
&\leq8\bar\eta\Delta^{(t)}.
\label{eq:det-app-speed}
\end{align}
Moreover, if these conditions hold for all $0\leq t\leq T$, then
\begin{equation}
\label{eq:det-app-full-path}
\sum_{t=0}^T
\|\theta^{(t+1)}-\theta^{(t)}\|
\leq32\sqrt{\frac{L(\theta^{(0)})}{\lambda_0}}.
\end{equation}
\end{lemma}

\begin{proof}
We first establish the Lyapunov descent analysis. The proof uses the smoothness of $L$ along the update segment, while the lower bound on the Gram matrix is needed only at the current iterate $\theta^{(t)}$.

Since $(P^{(t)})^2\preceq P^{(t)}\preceq I$, the Adam update yields
\begin{equation}
\label{eq:det-app-step-bound}
\|\theta^{(t+1)}-\theta^{(t)}\|^2
\leq2\bar\eta^2
\bigl(Q^{(t)}+\|e^{(t)}\|^2\bigr).
\end{equation}
Moreover, the weighted Cauchy--Schwarz and Young's inequalities give
\[
|(g^{(t)})^\top P^{(t)}e^{(t)}|
\leq\frac14Q^{(t)}+\|e^{(t)}\|^2.
\]
Applying the smoothness inequality along the update segment and using $H\bar\eta\leq1/8$, we obtain
\begin{align}
L(\theta^{(t+1)})-L(\theta^{(t)})
&\leq-\bar\eta Q^{(t)}
-\bar\eta(g^{(t)})^\top P^{(t)}e^{(t)}
+\frac H2\|\theta^{(t+1)}-\theta^{(t)}\|^2
\nonumber\\
&\leq-\frac{\bar\eta}{2}Q^{(t)}
+2\bar\eta\|e^{(t)}\|^2.
\label{eq:det-app-loss-step}
\end{align}

Next, we control the tracking error $e^{(t)}$. The exact bias-correction formula gives
\[
\widehat s^{(t+1)}
=d_{t+1}\widehat s^{(t)}
+(1-d_{t+1})g^{(t+1)},
\qquad
d_{t+1}
=\frac{\beta_1(1-\beta_1^{t+1})}
{1-\beta_1^{t+2}}
\leq\beta_1.
\]
Consequently,
\[
e^{(t+1)}
=d_{t+1}
\bigl(e^{(t)}+g^{(t)}-g^{(t+1)}\bigr).
\]
For $\beta_1>0$, applying
$\|u+v\|^2\leq\|u\|^2/\beta_1+\|v\|^2/\gamma$
yields
\begin{align}
\|e^{(t+1)}\|^2
&\leq\beta_1\|e^{(t)}\|^2
+\frac{\beta_1^2}{\gamma}
\|g^{(t+1)}-g^{(t)}\|^2
\nonumber\\
&\leq\beta_1\|e^{(t)}\|^2
+\frac{2\beta_1^2H^2\bar\eta^2}{\gamma}
\bigl(Q^{(t)}+\|e^{(t)}\|^2\bigr),
\label{eq:det-app-tracking}
\end{align}
where the second inequality follows from the smoothness of $L$ along the update segment and \eqref{eq:det-app-step-bound}. When $\beta_1=0$, we have $e^{(t)}=0$ for all $t$, and the subsequent estimates hold directly.

Multiplying \eqref{eq:det-app-tracking} by $4\bar\eta/\gamma$ and combining it with \eqref{eq:det-app-loss-step}, we obtain
\[
\mathcal E^{(t+1)}-\mathcal E^{(t)}
\leq-\frac{3\bar\eta}{8}Q^{(t)}
-\frac{15\bar\eta}{8}\|e^{(t)}\|^2,
\]
where we use $8(H\bar\eta/\gamma)^2\leq1/8$.
This establishes \eqref{eq:det-app-dissipation}. Furthermore, the lower bounds on the Gram matrix and the normalized preconditioner imply
\[
Q^{(t)}
\geq\frac12\|g^{(t)}\|^2
\geq\frac{\lambda_0}{4}L(\theta^{(t)}).
\]
It follows that
\[
\begin{aligned}
\Delta^{(t)}
&\geq\frac{\bar\eta\lambda_0}{16}L(\theta^{(t)})
+\bar\eta\|e^{(t)}\|^2\\
&\geq\frac{\bar\eta\lambda_0}{32}\mathcal E^{(t)},
\end{aligned}
\]
where the last inequality uses $\lambda_0\leq1$, $H\geq1$, and $\bar\eta\leq\gamma/(8H)$. This proves the contraction estimate \eqref{eq:det-app-contraction}. Combining \eqref{eq:det-app-step-bound} with \eqref{eq:det-app-dissipation} further yields \eqref{eq:det-app-speed}.

Finally, we establish the bound on the cumulative parameter movement. 
% The key is to avoid introducing an additional factor of $\lambda_0^{-1/2}$ when summing the step lengths. 
To avoid an additional factor of $\lambda_0^{-1/2}$, we use a telescoping argument based on the square root of the Lyapunov function.
Since $1-q=\bar\eta\lambda_0/32$, the contraction estimate implies
\[
\Delta^{(t)}
\geq(1-q)\mathcal E^{(t)}.
\]
Whenever $\mathcal E^{(t)}>0$, we have
\[
\begin{aligned}
\sqrt{\Delta^{(t)}}
&=
\frac{\sqrt{\mathcal E^{(t)}}+
\sqrt{\mathcal E^{(t+1)}}}
{\sqrt{\Delta^{(t)}}}
\left(
\sqrt{\mathcal E^{(t)}}-
\sqrt{\mathcal E^{(t+1)}}
\right)\\
&\leq\frac{2}{\sqrt{1-q}}
\left(
\sqrt{\mathcal E^{(t)}}-
\sqrt{\mathcal E^{(t+1)}}
\right).
\end{aligned}
\]
If $\mathcal E^{(t)}=0$, the update vanishes and the same inequality holds trivially. Taking square roots in \eqref{eq:det-app-speed} and summing over $t=0,\ldots,T$ gives
\[
\begin{aligned}
\sum_{t=0}^T
\|\theta^{(t+1)}-\theta^{(t)}\|
&\leq
\sqrt{8\bar\eta}
\sum_{t=0}^T\sqrt{\Delta^{(t)}}\\
&\leq
2\sqrt{\frac{8\bar\eta}{1-q}}
\sqrt{\mathcal E^{(0)}}.
\end{aligned}
\]
By exact bias correction, $\widehat s^{(0)}=g^{(0)}$, so $\mathcal E^{(0)}=L(\theta^{(0)})$. Substituting $1-q=\bar\eta\lambda_0/32$ completes the proof of \eqref{eq:det-app-full-path}.
\end{proof}

\subsection{Completion of the Proof}

\begin{proof}[Proof of Theorem~\ref{thm:reg-det}]
We work on the initialization event established in Lemma~\ref{lem:det-app-init}. If $L_0=0$, then $g^{(0)}=0$, and all subsequent gradients and moment estimates remain zero. Thus, the conclusion holds trivially. Henceforth, assume that $L_0>0$.

Choose the universal constants in the theorem such that
\begin{equation}
\label{eq:det-app-bootstrap-parameters}
\bar\epsilon\geq C_g\sqrt{S_0},
\qquad
0<\bar\eta\leq\frac{\gamma}{8H},
\qquad
m\geq C_*\frac{S_0}{\lambda_0^2},
\end{equation}
where $C_*>0$ is a sufficiently large universal constant to be specified below. Let
\[
q=1-\frac{\bar\eta\lambda_0}{32},
\qquad
\mathcal R=c_0\sqrt{m\lambda_0},
\]
as defined in the preceding lemmas. Under these conditions, $q\in(0,1)$ and $1-q\leq\gamma/2$.

We prove by induction that, for every integer $t\geq0$,
\begin{equation}
\label{eq:det-app-induction}
\begin{gathered}
\mathcal E^{(t)}\leq q^tL_0,
\qquad
\max_{1\leq r\leq m}
\|\theta_r^{(t)}-\theta_r^{(0)}\|
\leq\frac12,\\
\|\theta^{(t)}-\theta^{(0)}\|
\leq\frac{\mathcal R}{2}.
\end{gathered}
\end{equation}
The assertions hold at $t=0$, since $\mathcal E^{(0)}=L_0$ and the parameter differences vanish. Suppose that they hold for all $0\leq t\leq T$. Then, for every $0\leq k\leq T$, we have
\[
|a_r^{(k)}|\leq\frac32,
\qquad
L(\theta^{(k)})\leq q^kL_0.
\]
In particular, $\theta^{(k)}\in\mathcal S$ for all $0\leq k\leq T$. By \eqref{eq:det-app-metric} and \eqref{eq:det-app-bootstrap-parameters}, we obtain $P^{(T)}\succeq I/2$. Moreover, the Gram matrix lower bound in \eqref{eq:det-app-local-kernel} yields
\[
K(\theta^{(T)})\succeq\frac{\lambda_0}{4}I.
\]

We next verify that the update segment from $\theta^{(T)}$ to $\theta^{(T+1)}$ lies in $\mathcal S$. By Lemma~\ref{lem:det-app-moments}, which only requires the iterates through time $T$ and the bound $P^{(t)}\preceq I$, we have
\begin{equation}
\label{eq:det-app-candidate-neurons}
\sum_{t=0}^T
\max_{1\leq r\leq m}
\|\theta_r^{(t+1)}-\theta_r^{(t)}\|
\leq\frac{128C_g}{\lambda_0}
\sqrt{\frac{L_0}{m}}
\leq\frac12,
\end{equation}
where the last inequality follows by choosing $C_*$ sufficiently large. Consequently,
\[
\max_{1\leq r\leq m}
\|\theta_r^{(T+1)}-\theta_r^{(0)}\|
\leq\frac12,
\]
which also implies $|a_r^{(T+1)}|\leq3/2$ for every $r$. Since $\mathcal S$ is convex, the entire segment $[\theta^{(T)},\theta^{(T+1)}]$ lies in $\mathcal S$. Thus, the smoothness condition required by Lemma~\ref{lem:det-app-energy} is verified before applying the descent estimate.

All the hypotheses of Lemma~\ref{lem:det-app-energy} are now satisfied at iteration $T$. Therefore,
\[
\mathcal E^{(T+1)}
\leq q\mathcal E^{(T)}
\leq q^{T+1}L_0.
\]
Applying the cumulative parameter movement bound in the same lemma over iterations $0,\ldots,T$, we further obtain
\[
\begin{aligned}
\|\theta^{(T+1)}-\theta^{(0)}\|
&\leq\sum_{t=0}^T
\|\theta^{(t+1)}-\theta^{(t)}\|\\
&\leq32\sqrt{\frac{L_0}{\lambda_0}}\\
&\leq\frac{c_0}{2}\sqrt{m\lambda_0}
=\frac{\mathcal R}{2},
\end{aligned}
\]
where the last inequality again follows by taking $C_*$ sufficiently large. Together with \eqref{eq:det-app-candidate-neurons}, this verifies all three assertions in \eqref{eq:det-app-induction} at time $T+1$, completing the induction. Consequently,
\[
L(\theta^{(t)})
\leq\mathcal E^{(t)}
\leq
\left(1-\frac{\bar\eta\lambda_0}{32}\right)^t
L(\theta^{(0)}),
\qquad t\geq0.
\]

It remains to verify that the conditions in \eqref{eq:det-app-bootstrap-parameters} follow from the assumptions of Theorem~\ref{thm:reg-det}. Lemma~\ref{lem:det-app-init} requires $m\gtrsim\lambda_0^{-1}\log(2n/\delta)$, while the induction requires
$m\gtrsim S_0/\lambda_0^2$. Since $S_0=C_0\log(2/\delta)$ and $\lambda_0\leq1$, we have
\[
\begin{aligned}
\frac1{\lambda_0}\log\frac{2n}{\delta}
&=\frac1{\lambda_0}\log\frac n\delta
+\frac{\log2}{\lambda_0}\\
&\leq
\frac1{\lambda_0}\log\frac n\delta
+\frac1{\lambda_0^2}\log\frac2\delta.
\end{aligned}
\]
Thus, both width conditions follow from \eqref{eq:reg-det-width}, up to universal constants. Similarly, the damping and stepsize conditions in \eqref{eq:det-app-bootstrap-parameters} are ensured by \eqref{eq:reg-det-parameters}.

Finally, the finite-length bound~\eqref{eq:det-app-candidate-neurons} on the trajectory implies that $\{\theta^{(t)}\}$ converges to a finite limit $\theta^{\star}$.
By the continuity of $L$ and
\eqref{eq:reg-det-rate}, we have
$L(\theta^{\star})=0$.

\end{proof}

\newpage

\section{Proof of Theorem~\ref{thm:reg-stoch}}
\label{app:reg-stoch-proof}

We use the network, loss function, initialization, and Adam iteration defined in the main text, with $\epsilon=\bar\epsilon/\sqrt{m}$ and $\eta=\bar\eta\epsilon$.

For convenience, define
\begin{equation}
\label{eq:stoch-app-abbreviations}
\gamma=1-\beta_1,\qquad
\nu=1-\beta_2,\qquad
\rho=\rho_b,\qquad
\chi_b=\frac{n-b}{b(n-1)},\qquad
\kappa=\frac{n}{b}.
\end{equation}
In particular, $\rho=1+\chi_b/\lambda_0$. Define the filtration
\[
\mathcal F_t
=\sigma\bigl(\theta^{(0)},B^{(0)},\ldots,B^{(t-1)}\bigr),
\qquad
\E_t[\cdot]=\E[\cdot\mid\mathcal F_t].
\]
Then $\theta^{(t)}$ and the moment estimates up to iteration $t-1$ are $\mathcal F_t$-measurable. Unless otherwise stated, probabilities and expectations are conditional on a fixed initialization. We use $\mathcal{G}$ to denote the final high-probability sampling event.

We will make use of the initialization and regularity estimates established in Lemmas~\ref{lem:det-app-init} and~\ref{lem:det-app-geometry}. In particular, the trace bound in the proof of Lemma~\ref{lem:det-app-init} implies $\kappa\leq\rho$. We also define
\[
T_*
=1+\left\lceil\frac{\log 2}{-\log\beta_2}\right\rceil.
\]

The proof consists of three main steps: constructing an auxiliary network with global regularity, establishing stochastic moment bounds and a conditional Lyapunov descent analysis, and controlling parameter movement through a stopping argument. We then use these results to prove the convergence guarantee for the Adam iterates.

\subsection{An Auxiliary Network with Global Regularity}
\label{subsec:stoch-auxiliary-network}

Lemma~\ref{lem:det-app-geometry} establishes the smoothness of the original loss function on $\mathcal S=\{\theta:|a_r|\leq 2,\ 1\leq r\leq m\}$. In the stochastic setting, however, the conditional descent analysis must account for all possible realizations of the next minibatch. Even when the current iterate lies in $\mathcal S$, the next update and the virtual iterates introduced later may leave this region. Consequently, the local smoothness bound alone is insufficient to justify the descent estimates. To overcome this difficulty, we construct an auxiliary network whose loss function is globally smooth and coincides with the original loss on $\mathcal S$. We first analyze Adam on this auxiliary network and then show that its iterates coincide with those of the original network on a suitable high-probability event.

Let $\psi\in C_c^\infty(\R)$ be a fixed smooth function satisfying
\[
\psi(a)=a \quad\text{for }|a|\leq 2,
\qquad
\operatorname{supp}\psi\subset[-3,3].
\]
Such a function can be constructed by multiplying $a$ by a smooth cutoff function that equals one on $[-2,2]$ and vanishes outside $[-3,3]$. We choose $\psi$ independently of the training data, network width, and optimizer parameters, so that $\psi$ and its first two derivatives are bounded by universal constants. Using the same parameter vector $\theta=(a_1,w_1,\ldots,a_m,w_m)$, we define the auxiliary network
\begin{equation}
\label{eq:stoch-app-aux-network}
f^\dagger_\theta(x)
=\frac{1}{\sqrt m}
\sum_{r=1}^m
\psi(a_r)\sigma(w_r^\top\widetilde{x}).
\end{equation}
Here, the superscript $\dagger$ denotes quantities associated with the auxiliary network. Correspondingly, we define the normalized residual vector, loss function, Jacobian, and Gram matrix by
\begin{equation}
\label{eq:stoch-app-aux-objects}
\begin{gathered}
R_i^\dagger(\theta)
=\frac{f^\dagger_\theta(x_i)-y_i}{\sqrt n},
\qquad
L^\dagger(\theta)
=\frac12\|R^\dagger(\theta)\|^2,\\
J^\dagger(\theta)=\nabla R^\dagger(\theta),
\qquad
K^\dagger(\theta)
=J^\dagger(\theta)J^\dagger(\theta)^\top.
\end{gathered}
\end{equation}

\begin{lemma}[Properties of the auxiliary network]
\label{lem:stoch-app-auxiliary}
The auxiliary network satisfies the following properties.
\begin{enumerate}
\item For any $\theta\in\mathcal S$ and input $x$, we have
\begin{equation}
\label{eq:stoch-app-aux-agreement}
f^\dagger_\theta(x)=f_\theta(x),
\qquad
\nabla_\theta f^\dagger_\theta(x)
=\nabla_\theta f_\theta(x).
\end{equation}
Consequently, $R^\dagger=R$, $L^\dagger=L$, $J^\dagger=J$, and $K^\dagger=K$ on $\mathcal S$. In particular, these identities hold at initialization.

\item There exists a universal constant $C>0$ such that, for all $\theta,\vartheta\in\R^{m(d+2)}$,
\begin{equation}
\label{eq:stoch-app-global-bounds}
\begin{gathered}
\|J^\dagger(\theta)\|\leq C,
\qquad
\|J^\dagger(\theta)-J^\dagger(\vartheta)\|
\leq\frac{C}{\sqrt m}\|\theta-\vartheta\|,\\
\|\nabla^2L^\dagger(\theta)\|\leq C,
\qquad
\max_{i,r}\|\nabla_{\theta_r}f^\dagger_\theta(x_i)\|
\leq\frac{C}{\sqrt m}.
\end{gathered}
\end{equation}

\item On the initialization event~\eqref{eq:det-app-init}, let
\[
\mathcal R=c_0\sqrt{m\lambda_0},
\]
where $c_0>0$ is a sufficiently small universal constant, chosen as in Lemma~\ref{lem:det-app-geometry}. Then, for every $\theta$ satisfying $\|\theta-\theta^{(0)}\|\leq\mathcal R$, we have
\begin{equation}
\label{eq:stoch-app-aux-kernel}
K^\dagger(\theta)\succeq\frac{\lambda_0}{4}I,
\qquad
\|\nabla L^\dagger(\theta)\|^2
\geq\frac{\lambda_0}{2}L^\dagger(\theta).
\end{equation}
\end{enumerate}
The constants depend only on the fixed activation and cutoff functions.
\end{lemma}

\begin{proof}
Since $\psi(a)=a$ and $\psi'(a)=1$ for $|a|\leq2$, direct differentiation of \eqref{eq:stoch-app-aux-network} gives \eqref{eq:stoch-app-aux-agreement}. Moreover, $|a_r^{(0)}|=1$ for all $r$, so the auxiliary and original networks agree at initialization. In particular, their Gram matrices agree at initialization, and the auxiliary network has the same limiting Gram matrix $K_\infty$ and minimum eigenvalue $\lambda_0$.

Next, we establish the global regularity bounds. The first- and second-order derivatives with respect to the parameters of neuron $r$ are
\[
\nabla_{\theta_r}f^\dagger_\theta(x_i)
=\frac{1}{\sqrt m}
\begin{pmatrix}
\psi'(a_r)\sigma(s_{ir})\\
\psi(a_r)\sigma'(s_{ir})\widetilde{x}_i
\end{pmatrix},
\]
and
\[
\nabla^2_{\theta_r}f^\dagger_\theta(x_i)
=\frac{1}{\sqrt m}
\begin{pmatrix}
\psi''(a_r)\sigma(s_{ir})
&
\psi'(a_r)\sigma'(s_{ir})\widetilde{x}_i^\top\\
\psi'(a_r)\sigma'(s_{ir})\widetilde{x}_i
&
\psi(a_r)\sigma''(s_{ir})
\widetilde{x}_i\widetilde{x}_i^\top
\end{pmatrix}.
\]
Since $\psi$, $\sigma$, and their derivatives up to second order are bounded, and $\|\widetilde{x}_i\|\leq1$, both derivatives are bounded in norm by $C/\sqrt m$ uniformly over the parameter space. Furthermore, mixed second derivatives between different neurons vanish. As in the proof of Lemma~\ref{lem:det-app-geometry}, we obtain
\[
\|J^\dagger(\theta)\|
\leq\|J^\dagger(\theta)\|_{\rm F}\leq C,
\]
and
\[
\|J^\dagger(\theta)-J^\dagger(\vartheta)\|_{\rm F}^2
\leq\frac{C}{m}
\sum_{r=1}^m\|\theta_r-\vartheta_r\|^2.
\]
These estimates establish the Jacobian bounds and the neuron-wise gradient bound in \eqref{eq:stoch-app-global-bounds}.

To establish the global smoothness of the auxiliary loss, we use
\[
\nabla^2L^\dagger(\theta)
=(J^\dagger(\theta))^\top J^\dagger(\theta)
+\sum_{i=1}^n
R_i^\dagger(\theta)\nabla^2R_i^\dagger(\theta).
\]
The second term is block diagonal with respect to the neuron parameters. By the same Cauchy--Schwarz argument as in Lemma~\ref{lem:det-app-geometry}, its operator norm is bounded by
$\frac{C}{\sqrt m}\|R^\dagger(\theta)\|.$
Since $\psi$ and $\sigma$ are bounded, we have
$|f^\dagger_\theta(x_i)|\leq C\sqrt m$ and hence
$\|R^\dagger(\theta)\|\leq C\sqrt m+1$. Consequently,
\[
\|\nabla^2L^\dagger(\theta)\|
\leq C+\frac{C}{\sqrt m}\|R^\dagger(\theta)\|
\leq C.
\]
Thus, $L^\dagger$ is globally smooth, with a smoothness constant that does not depend on the network width.

Finally, we establish the lower bound on the auxiliary Gram matrix. On the initialization event~\eqref{eq:det-app-init}, the agreement between the two networks gives
\[
K^\dagger(\theta^{(0)})
=K(\theta^{(0)})
\succeq\frac{3\lambda_0}{4}I.
\]
For any unit vector $z\in\R^n$, the Jacobian Lipschitz bound implies
\[
\begin{aligned}
\|(J^\dagger(\theta))^\top z\|
&\geq
\|(J^\dagger(\theta^{(0)}))^\top z\|
-\|J^\dagger(\theta)-J^\dagger(\theta^{(0)})\|\\
&\geq
\sqrt{\frac{3\lambda_0}{4}}
-\frac{C}{\sqrt m}\|\theta-\theta^{(0)}\|\\
&\geq\frac{\sqrt{\lambda_0}}{2},
\end{aligned}
\]
where the last inequality holds whenever
$\|\theta-\theta^{(0)}\|\leq\mathcal R$, provided that $c_0$ is sufficiently small. Since this holds for every unit vector $z$, we obtain
\[
K^\dagger(\theta)
\succeq\frac{\lambda_0}{4}I.
\]
The gradient inequality in \eqref{eq:stoch-app-aux-kernel} follows from
$\nabla L^\dagger(\theta)=(J^\dagger(\theta))^\top R^\dagger(\theta)$ and
$\|R^\dagger(\theta)\|^2=2L^\dagger(\theta)$.
\end{proof}

\paragraph{Auxiliary Adam and trajectory agreement.}
We apply Adam to the auxiliary loss $L^\dagger$, using the same optimizer parameters, initialization, and minibatch sequence as the original algorithm. Denote the auxiliary iterates and moment estimates by $\theta_\dagger^{(t)}$, $s_\dagger^{(t)}$, and $v_\dagger^{(t)}$, respectively, where $\theta_{\dagger,r}^{(t)}=(a_{\dagger,r}^{(t)},w_{\dagger,r}^{(t)})$. We initialize
\[
\theta_\dagger^{(0)}=\theta^{(0)},
\qquad
s_\dagger^{(-1)}=v_\dagger^{(-1)}=0,
\]
and compute the stochastic gradient
\[
g_\dagger^{(t)}
=\frac{1}{b}\sum_{i\in B^{(t)}}
\bigl(f^\dagger_{\theta_\dagger^{(t)}}(x_i)-y_i\bigr)
\nabla_\theta f^\dagger_{\theta_\dagger^{(t)}}(x_i).
\]
The moment estimates, bias corrections, and parameter updates follow \eqref{eq:adam-moment}--\eqref{eq:adam-update}, with the corresponding auxiliary quantities.

We now establish the agreement between the auxiliary and original Adam trajectories. For any integer $T\geq0$, if $\theta_\dagger^{(t)}\in\mathcal S$ for all $0\leq t\leq T$, then the two algorithms generate identical parameter iterates through time $T+1$, as well as identical stochastic gradients and moment estimates through time $T$. Indeed, both algorithms start from the same initialization. Whenever their iterates coincide at a point in $\mathcal S$, the agreement property \eqref{eq:stoch-app-aux-agreement} and the shared minibatch imply identical stochastic gradients. The moment recurrences and parameter updates are therefore identical, and the claim follows by induction. In the proof of Theorem~\ref{thm:reg-stoch}, we establish a high-probability event $\mathcal G$ on which $|a_{\dagger,r}^{(t)}|\leq3/2$ for all $r$ and $t$. Hence, the auxiliary and original Adam trajectories coincide for all iterations on $\mathcal G$.

\paragraph{Notation for the remaining proof.}
For notational simplicity, we omit the superscript or subscript $\dagger$ from all auxiliary quantities in the following analysis, until the final comparison with the original algorithm. 
In particular, $\theta^{(t)}$, $L(\theta^{(t)})$, and $g^{(t)}$ denote $\theta^{(t)}_{\dagger}$, $L^\dagger(\theta_\dagger^{(t)})$, and $g_\dagger^{(t)}$, respectively, and we write
\[
G^{(t)}=\nabla L^\dagger(\theta_\dagger^{(t)})
\]
for the full gradient of the auxiliary loss. The same convention applies to $f$, $R$, $J$, $K$, $s$, $v$, $\widehat s$, $\widehat v$, $D$, and the normalized preconditioner $P$. The filtration $\mathcal F_t$ remains unchanged because the two algorithms share the same initialization and minibatch sequence. Throughout the subsequent analysis, global smoothness refers to the auxiliary loss $L^\dagger$, while the lower bound on the Gram matrix is invoked only within the neighborhood specified in \eqref{eq:stoch-app-aux-kernel}.

\subsection{Key Lemmas}
\label{subsec:stoch-key-lemmas}

\begin{lemma}[Minibatch moment bounds]
\label{lem:stoch-app-sampling}
Let $L^{(t)}=L(\theta^{(t)})$  and let $g_r^{(t)}$ denote the block of the stochastic gradient $g^{(t)}$ corresponding to the parameters $\theta_r=(a_r,w_r)$ of neuron $r$.
Define
\[
z_i=(f_{\theta^{(t)}}(x_i)-y_i)
\nabla_\theta f_{\theta^{(t)}}(x_i).
\]
Then the stochastic gradient satisfies
\begin{equation}
\label{eq:stoch-app-covariance}
\E_t[g^{(t)}]=G^{(t)},
\qquad
\operatorname{Cov}_t(g^{(t)})
=\frac{\chi_b}{n}\sum_{i=1}^n
(z_i-G^{(t)})(z_i-G^{(t)})^\top.
\end{equation}
Moreover, the following moment bounds hold for all iterates:
\begin{equation}
\label{eq:stoch-app-batch-moments}
\E_t\|g^{(t)}\|^2\leq CL^{(t)},
\qquad
\max_j\E_t|g_j^{(t)}|^2
\leq\frac{CL^{(t)}}{m}.
\end{equation}
If $K(\theta^{(t)})\succeq\lambda_0 I/4$, we further have
\begin{equation}
\label{eq:stoch-app-relative-moments}
\E_t\|g^{(t)}-G^{(t)}\|^2
\leq C(\rho_b-1)\|G^{(t)}\|^2,
\qquad
\E_t\|g^{(t)}\|^2
\leq C\rho_b\|G^{(t)}\|^2.
\end{equation}

Next, define the sampled squared residual and its bias-corrected moving average by
\begin{equation}
\label{eq:stoch-app-scalar-average}
Y^{(t)}
=\frac1b\sum_{i\in B^{(t)}}
(f_{\theta^{(t)}}(x_i)-y_i)^2,
\qquad
U^{(t)}
=\sum_{s=0}^t
\frac{\nu\beta_2^{t-s}}{1-\beta_2^{t+1}}
Y^{(s)}.
\end{equation}
Then the following bounds hold pathwise:
\begin{equation}
\label{eq:stoch-app-batch-blocks}
\|g^{(t)}\|^2\leq CY^{(t)},
\qquad
\max_r\|g_r^{(t)}\|^2
\leq\frac{CY^{(t)}}{m},
\qquad
\max_j \widehat v_j^{(t)}
\leq 
\frac{C}{m}U^{(t)}.
\end{equation}
Furthermore, $Y^{(t)}$ satisfies
\begin{equation}
\label{eq:stoch-app-scalar-moments}
\E_t[Y^{(t)}]=2L^{(t)},
\qquad
0\leq Y^{(t)}\leq2\kappa L^{(t)},
\qquad
\operatorname{Var}_t(Y^{(t)})
\leq4\kappa(L^{(t)})^2.
\end{equation}
\end{lemma}

\begin{proof}
Let $I_i=\mathbf1_{\{i\in B^{(t)}\}}$ denote the minibatch sampling indicators. Since $B^{(t)}$ is sampled uniformly without replacement,
\[
\E_t[I_i]=\frac bn,
\qquad
\E_t[I_iI_j]
=\frac{b(b-1)}{n(n-1)},
\quad i\neq j.
\]
Using $g^{(t)}=b^{-1}\sum_{i=1}^n I_i z_i$, we obtain the conditional unbiasedness and covariance identities in \eqref{eq:stoch-app-covariance}.

The global derivative bounds in \eqref{eq:stoch-app-global-bounds} imply
\[
\frac1n\sum_{i=1}^n\|z_i\|^2
\leq CL^{(t)},
\]
together with the corresponding coordinatewise bound of order $L^{(t)}/m$. The covariance identity therefore gives
\[
\E_t\|g^{(t)}-G^{(t)}\|^2
\leq C\chi_b L^{(t)}.
\]
When $K(\theta^{(t)})\succeq\lambda_0 I/4$, the gradient inequality \eqref{eq:kernel-pl} gives
\[
\|G^{(t)}\|^2
\geq\frac{\lambda_0}{2}L^{(t)}.
\]
Combining these estimates and using $\rho_b=1+\chi_b/\lambda_0$ proves \eqref{eq:stoch-app-relative-moments}.

Next, Jensen's inequality applied to the minibatch average, together with the global derivative bounds, yields the first two inequalities in \eqref{eq:stoch-app-batch-blocks}. Taking conditional expectations gives \eqref{eq:stoch-app-batch-moments}. The bound on $\widehat v^{(t)}$ follows from the bias-corrected second-moment representation \eqref{eq:det-app-v-history} and the coordinatewise gradient bounds.

Finally, we establish the scalar moment estimates. Let
\[
u_i=(f_{\theta^{(t)}}(x_i)-y_i)^2\geq0,
\qquad
\mu=\frac1n\sum_{i=1}^n u_i=2L^{(t)}.
\]
By construction, $\E_t[Y^{(t)}]=\mu$ and
\[
0\leq Y^{(t)}\leq\frac nb\mu=\kappa\mu.
\]
Moreover, the covariance formula for uniform sampling without replacement gives
\[
\begin{aligned}
\operatorname{Var}_t(Y^{(t)})
&=\chi_b\left(
\frac1n\sum_{i=1}^n u_i^2-\mu^2
\right)\\
&\leq\chi_b(n-1)\mu^2\\
&\leq\kappa\mu^2,
\end{aligned}
\]
where we used
$\sum_i u_i^2\leq(\sum_i u_i)^2$.
Substituting $\mu=2L^{(t)}$ completes the proof of \eqref{eq:stoch-app-scalar-moments}.
\end{proof}

\begin{lemma}[Uniform second-moment control]
\label{lem:stoch-app-scalar-cap}
Let $0<\nu\leq 1/2$, $q=1-\Delta_{q}\in(0,1)$, and $0<\Delta_{q}\leq\nu/2$. Suppose that $S>0$ is deterministic and $I^{(s)}\in\{0,1\}$ is $\mathcal F_s$-measurable, satisfying
\[
I^{(s)}L(\theta^{(s)})
\leq Sq^{(s-T_*)_+},
\qquad s\geq0.
\]
Define the masked moving average
\[
\widetilde U^{(t)}
=\sum_{s=0}^t
\frac{\nu\beta_2^{t-s}}{1-\beta_2^{t+1}}
I^{(s)}Y^{(s)}.
\]
Then, for any $\delta_v\in(0,1/2)$, there exist universal constants $C,c>0$ such that
\begin{equation}
\label{eq:stoch-app-scalar-condition}
\nu\kappa\log\frac{C}{\Delta_{q}\delta_v}\leq c
\quad\Longrightarrow\quad
\mathbb P\left(
\sup_{t\geq T_*-1}\widetilde U^{(t)}>CS
\right)\leq\delta_v.
\end{equation}
\end{lemma}

\begin{proof}
Fix $t\geq T_*-1$ and let $j=(t-T_*)_+$. Define the moving-average coefficients
\[
\alpha_{t,s}
=\frac{\nu\beta_2^{t-s}}{1-\beta_2^{t+1}},
\qquad 0\leq s\leq t.
\]
By the definition of $T_*$, we have
$\alpha_{t,s}\leq2\nu\beta_2^{t-s}$.
Moreover, since $q=1-\Delta_{q}$ and $\beta_2=1-\nu$,
\[
1-\frac{\beta_2}{q}
=\frac{\nu-\Delta_{q}}{q}
\geq\frac{\nu}{2}.
\]
Using these inequalities and summing the corresponding geometric series, we obtain
\[
\begin{aligned}
\sum_{s=0}^t
\alpha_{t,s}q^{(s-T_*)_+}
&\leq Cq^j,\\
\max_{0\leq s\leq t}
\alpha_{t,s}q^{(s-T_*)_+}
&\leq C\nu q^j,\\
\sum_{s=0}^t
\alpha_{t,s}^2q^{2(s-T_*)_+}
&\leq C\nu q^{2j}.
\end{aligned}
\]

Next, define the martingale differences
\[
X_s
=\alpha_{t,s}I^{(s)}
\bigl(Y^{(s)}-2L^{(s)}\bigr),
\qquad 0\leq s\leq t.
\]
Since $I^{(s)}$ is $\mathcal F_s$-measurable and
$\E_s[Y^{(s)}]=2L^{(s)}$, we have
$\E_s[X_s]=0$.
By \eqref{eq:stoch-app-scalar-moments}, the assumed loss bound, and the weight estimates,
\[
|X_s|
\leq C\nu\kappa Sq^j,
\qquad
\sum_{s=0}^t\E_s[X_s^2]
\leq C\nu\kappa S^2q^{2j}.
\]
Furthermore, the predictable component of the moving average satisfies
\[
\sum_{s=0}^t
\alpha_{t,s}I^{(s)}\,2L^{(s)}
\leq CSq^j.
\]
Applying the conditional Bernstein inequality in Lemma~\ref{lem:aux-martingale-bernstein} with a sufficiently large constant multiple of $S$ as the deviation threshold yields
\[
\mathbb P\left(
\widetilde U^{(t)}>CS
\right)
\leq
\exp\left(
-\frac{cq^{-j}}{\nu\kappa}
\right).
\]

To obtain a uniform bound over all iterations, let
$A=c/(\nu\kappa)\geq1$.
Since $q=1-\Delta_{q}$, we have $q^{-j}\geq1+\Delta_{q}j$.
A union bound therefore gives
\[
\begin{aligned}
\mathbb P\left(
\sup_{t\geq T_*-1}\widetilde U^{(t)}>CS
\right)
&\leq e^{-A}
+\sum_{j=0}^{\infty}e^{-Aq^{-j}}\\
&\leq e^{-A}
+\frac{e^{-A}}{1-e^{-A\Delta_{q}}}\\
&\leq\frac{Ce^{-A}}{\Delta_{q}},
\end{aligned}
\]
where the first term accounts for $t=T_*-1$.
Under condition~\eqref{eq:stoch-app-scalar-condition}, the last expression is at most $\delta_v$, completing the proof.
\end{proof}

The following lemma provides a stochastic counterpart to Lemma~\ref{lem:det-app-energy}. We use the normalized preconditioner $P^{(t)}=\epsilon D^{(t)}$ defined in~\eqref{eq:det-app-normalization}. The bias-corrected moment representations~\eqref{eq:det-app-s-history} and~\eqref{eq:det-app-v-history} also hold for stochastic gradients.
To establish conditional descent, we introduce a virtual iterate that separates the current stochastic gradient from the historical momentum contribution. Using the exact bias-correction formulas, we define, for $t\geq1$,
\begin{equation}
\label{eq:stoch-app-virtual}
\begin{gathered}
d^{(t)}=\frac{\beta_1(1-\beta_1^t)}{1-\beta_1^{t+1}},
\qquad
\omega^{(t)}=\frac{\nu}{1-\beta_2^{t+1}},
\qquad
k^{(t)}=\sum_{s=t}^{\infty}\prod_{j=t}^s d^{(j)},\\
Z^{(t)}
=\theta^{(t)}
-\bar\eta k^{(t)}P^{(t-1)}\widehat s^{(t-1)},\\
\zeta=C_0\frac{\bar\eta\nu}{\gamma^2},
\qquad
\mathcal E^{(t)}
=L(Z^{(t)})
+\zeta\|\widehat s^{(t-1)}\|^2,
\end{gathered}
\end{equation}
where $C_0>0$ is a sufficiently large universal constant.

\begin{lemma}[Stochastic Lyapunov descent]
\label{lem:stoch-app-energy}
Suppose that $t\geq T_*$ and that, for a fixed universal constant $C_U>0$,
\[
K(\theta^{(t)})\succeq\frac{\lambda_0}{4}I,
\qquad
L(\theta^{(t)})\leq S,
\qquad
U^{(t-1)}\leq C_US.
\]
If $\bar\epsilon\geq C\sqrt S$,
$0<\nu\leq c\gamma/\rho_b$, and
$0<\bar\eta\leq c\nu$, then
\begin{equation}
\label{eq:stoch-app-drift}
\begin{aligned}
\mathcal E^{(t)}-\E_t[\mathcal E^{(t+1)}]
&\geq \mathcal D^{(t)}\\
&:=c\bar\eta\|G^{(t)}\|^2
+c\gamma\zeta\|\widehat s^{(t-1)}\|^2\\
&\geq c\bar\eta\lambda_0\mathcal E^{(t)}.
\end{aligned}
\end{equation}
Moreover, let $\Delta Z^{(t)}=Z^{(t+1)}-Z^{(t)}$. Then
\begin{align}
\|\E_t[\Delta Z^{(t)}]\|^2
&\leq C\bar\eta\mathcal D^{(t)},\nonumber\\
\E_t\left[
\|\Delta Z^{(t)}-\E_t[\Delta Z^{(t)}]\|^2
\right]
&\leq
\frac{C\bar\eta^2(\rho_b-1)}{\gamma}
\|G^{(t)}\|^2,
\label{eq:stoch-app-speeds}
\end{align}
where $\mathcal D^{(t)}$ is defined in~\eqref{eq:stoch-app-drift}.
Finally, the following bounds hold globally for the auxiliary process, without assuming the local Gram matrix and loss bounds:
\begin{equation}
\label{eq:stoch-app-comparison}
L(\theta^{(t)})\leq2\mathcal E^{(t)},
\qquad
\|\theta^{(t)}-Z^{(t)}\|^2
\leq C\mathcal E^{(t)}.
\end{equation}
\end{lemma}

\begin{proof}
Fix $\mathcal F_t$ and write $G=G^{(t)}$ and $h=\widehat s^{(t-1)}$ for brevity. Under the prescribed parameter conditions, we have $\beta_2\geq\beta_1$ and $\nu\leq1/4$. Moreover, for $t\geq T_*$,
\[
1-d^{(t)}\leq2\gamma,\qquad
\omega^{(t)}\leq2\nu,\qquad
1-\omega^{(t)}\geq\frac12.
\]
Since $d^{(t)}$ is nondecreasing and bounded above by $\beta_1$, the definition of $k^{(t)}$ yields
\[
0\leq k^{(t)}\leq\frac{\beta_1}{\gamma},
\qquad
k^{(t)}=d^{(t)}(1+k^{(t+1)}).
\]
Consequently,
\[
1\leq c^{(t)}
:=1+k^{(t+1)}-k^{(t)}
=(1+k^{(t+1)})(1-d^{(t)})
\leq2.
\]

We first establish a bound relating the historical first- and second-moment estimates. Applying the weighted Cauchy--Schwarz inequality to their bias-corrected representations gives
\[
\begin{aligned}
|h_j|^2
&\leq
\widehat v_j^{(t-1)}
\frac{\gamma^2(1-\beta_2^t)}
{\nu(1-\beta_1^t)^2}
\sum_{s=0}^{t-1}
\left(\frac{\beta_1^2}{\beta_2}\right)^s\\
&\leq
\frac{4\gamma}{\nu}
\widehat v_j^{(t-1)}.
\end{aligned}
\]
Here, we used $\beta_1^t\leq\beta_2^t\leq1/2$ and
$1-\beta_1^2/\beta_2\geq\gamma$.
Furthermore, the convex combination
\[
\widehat s^{(t)}
=d^{(t)}h+(1-d^{(t)})g^{(t)}
\]
and the minibatch moment bounds in
Lemma~\ref{lem:stoch-app-sampling} imply
\begin{equation}
\label{eq:stoch-app-moment-innovation}
\E_t\|\widehat s^{(t)}\|^2
\leq
\beta_1\|h\|^2
+C\gamma\rho_b\|G\|^2.
\end{equation}

Next, we analyze the virtual iterate update. For any gradient vector $g$, define
\[
P_t(g)=\diag\left(
\frac{\epsilon}{
\epsilon+
\sqrt{
(1-\omega^{(t)})\widehat v^{(t-1)}
+\omega^{(t)}g^{\odot2}
}}
\right),
\qquad
P^-=P_t(0).
\]
Substituting the bias-corrected moment recurrences into
\eqref{eq:stoch-app-virtual} yields the exact identity
\begin{equation}
\label{eq:stoch-app-exact-update}
\begin{aligned}
\Delta Z^{(t)}
&=\bar\eta u_t(g^{(t)}),\\
u_t(g)
&=-c^{(t)}P_t(g)g
+k^{(t)}(P^{(t-1)}-P_t(g))h.
\end{aligned}
\end{equation}
We decompose this update into a principal term involving the past-dependent preconditioner $P^-$ and a remainder:
\begin{equation}
\label{eq:stoch-app-predictable-map}
\begin{aligned}
u_t(g)
&=-c^{(t)}P^-g
+k^{(t)}(P^{(t-1)}-P^-)h
+e_t(g),\\
e_t(g)
&=(P^--P_t(g))(c^{(t)}g+k^{(t)}h).
\end{aligned}
\end{equation}

We first bound the remainder $e_t(g)$, which captures the dependence of the adaptive preconditioner on the current gradient. For each coordinate $j$, define
\[
A_j=\sqrt{(1-\omega^{(t)})\widehat v_j^{(t-1)}},
\qquad
B_j=\sqrt{A_j^2+\omega^{(t)}g_j^2}.
\]
Then
\[
(P^--P_t(g))_{jj}
=\frac{\epsilon(B_j-A_j)}
{(\epsilon+A_j)(\epsilon+B_j)},
\]
where
\[
B_j-A_j\leq\sqrt{\omega^{(t)}}|g_j|,
\qquad
A_j(B_j-A_j)\leq\omega^{(t)}g_j^2.
\]
Combining these inequalities with
$|h_j|\leq C\sqrt{\gamma/\nu}\,A_j$ gives
\[
|e_{t,j}(g)|
\leq
C\sqrt{\frac{\nu}{\gamma}}
\frac{g_j^2}{\epsilon}.
\]
By Lemma~\ref{lem:stoch-app-sampling}, we therefore obtain
\begin{equation}
\label{eq:stoch-app-normalization-bias}
\begin{aligned}
\|\E_t[e_t(g^{(t)})]\|^2
&\leq
\frac{C\nu}{\gamma\epsilon^2}
\left(\max_j\E_t|g_j^{(t)}|^2\right)
\E_t\|g^{(t)}\|^2\\
&\leq
\frac{C\nu\rho_b S}
{\gamma\bar\epsilon^2}\|G\|^2.
\end{aligned}
\end{equation}
Thus, the dependence of the preconditioner on the current minibatch is captured by a controllable remainder.

We next bound the conditional variance of the virtual update. For each coordinate $j$, define
\[
p_j(z)
=\frac{\epsilon}
{\epsilon+\sqrt{A_j^2+\omega^{(t)}z^2}}.
\]
Direct differentiation yields
\[
|(p_j(z)z)'|\leq1,
\qquad
A_j|p_j'(z)|\leq C\sqrt{\omega^{(t)}}.
\]
Together with the preceding historical moment bound, these inequalities imply that each coordinate of the mapping $u_t$ is $C/\sqrt{\gamma}$-Lipschitz. When $A_j=0$, the corresponding historical moment coordinate vanishes, and the same bound follows directly from
\[
p_j(z)z
=\frac{\epsilon z}
{\epsilon+\sqrt{\omega^{(t)}}|z|}.
\]
Applying this Lipschitz estimate to two conditionally independent copies of $g^{(t)}$, together with \eqref{eq:stoch-app-relative-moments}, yields
\begin{equation}
\label{eq:stoch-app-update-variance}
\begin{aligned}
\E_t\left[
\|u_t(g^{(t)})-\E_t[u_t(g^{(t)})]\|^2
\right]
&\leq
\frac{C}{\gamma}
\E_t\|g^{(t)}-G\|^2\\
&\leq
\frac{C(\rho_b-1)}{\gamma}\|G\|^2.
\end{aligned}
\end{equation}

We now establish the conditional descent estimate. The historical second-moment bound and the damping condition ensure that
\[
p_*I\preceq P^-\preceq I
\]
for some universal constant $p_*>0$.
Moreover,
\[
\|P^{(t-1)}-P^-\|\leq\omega^{(t)}.
\]
Indeed, letting
$\ell_j=\sqrt{\widehat v_j^{(t-1)}}/\epsilon$, we have
\[
\frac{(P^--P^{(t-1)})_{jj}}{P^-_{jj}}
=
\frac{\ell_j-\sqrt{1-\omega^{(t)}}\,\ell_j}{1+\ell_j}
\leq\omega^{(t)}.
\]
Taking conditional expectations in
\eqref{eq:stoch-app-predictable-map}, we obtain
\[
\E_t[u_t(g^{(t)})]
=-c^{(t)}P^-G+\mathfrak r^{(t)},
\]
where
\[
\|\mathfrak r^{(t)}\|
\leq\alpha_{\mathrm{err}}\|G\|
+\frac{C\nu}{\gamma}\|h\|,
\qquad
\alpha_{\mathrm{err}}^2
\leq\frac{C\nu\rho_b}{\gamma}.
\]
Under the prescribed parameter conditions, $\alpha_{\mathrm{err}}$ is sufficiently small relative to $p_*$.

By global smoothness and the definition of the virtual iterate,
\[
\|\nabla L(Z^{(t)})-G\|
\leq
\frac{C\bar\eta}{\gamma}\|h\|.
\]
Combining this estimate with the conditional mean and variance
bounds for the virtual update, and using the smallness of
$\alpha_{\mathrm{err}}$, we obtain
\[
\begin{aligned}
&\E_t[L(Z^{(t+1)})]-L(Z^{(t)})\\
&\quad\leq
-c_0\bar\eta\|G\|^2
+\frac{C\bar\eta^2\rho_b}{\gamma}\|G\|^2+
\frac{C\bar\eta(\nu+\bar\eta)}{\gamma}
\|G\|\|h\|
+\frac{C\bar\eta^2(\nu+\nu^2)}{\gamma^2}
\|h\|^2,
\end{aligned}
\]
where $c_0>0$ is a universal constant.
By Young's inequality and $\bar\eta\leq c\nu$, the mixed term
can be bounded by
\[
\begin{aligned}
\frac{C\bar\eta(\nu+\bar\eta)}{\gamma}
\|G\|\|h\|
&\leq
\frac{c_0\bar\eta}{2}\|G\|^2+
\frac{C\bar\eta\nu^2}{\gamma^2}\|h\|^2.
\end{aligned}
\]
The remaining historical-moment terms are also bounded by
$C\bar\eta\nu^2\|h\|^2/\gamma^2$.
Consequently,
\begin{equation}
\label{eq:stoch-app-virtual-descent}
\begin{aligned}
&\E_t[L(Z^{(t+1)})]-L(Z^{(t)})\\
&\quad\leq
-c_1\bar\eta\|G\|^2
+\frac{C\bar\eta^2\rho_b}{\gamma}\|G\|^2
+\frac{C\bar\eta\nu^2}{\gamma^2}\|h\|^2,
\end{aligned}
\end{equation}
where $c_1>0$ is a universal constant.

We next establish the Lyapunov descent analysis.
Recall that $\mathcal E^{(t)}=L(Z^{(t)})+\zeta\|h\|^2,$ and $\zeta=C_0\frac{\bar\eta\nu}{\gamma^2}.$
Combining \eqref{eq:stoch-app-virtual-descent} with
$\zeta$ times \eqref{eq:stoch-app-moment-innovation},
and using $\beta_1=1-\gamma$, yields
\[
\begin{aligned}
&\E_t[\mathcal E^{(t+1)}]-\mathcal E^{(t)}\\
&\quad\leq
-\bar\eta
\left(
c_1
-\frac{C\bar\eta\rho_b}{\gamma}
-\frac{CC_0\nu\rho_b}{\gamma}
\right)\|G\|^2
-\gamma\zeta
\left(
1-\frac{C\nu}{C_0\gamma}
\right)\|h\|^2.
\end{aligned}
\]
Choose $C_0$ sufficiently large and then take
the universal constants in
$\bar\eta\leq c\nu$ and
$\nu\leq c\gamma/\rho_b$
sufficiently small. Since $\rho_b\geq1$,
the two coefficients satisfy
\[
c_1
-\frac{C\bar\eta\rho_b}{\gamma}
-\frac{CC_0\nu\rho_b}{\gamma}
\geq\frac{c_1}{2},
\qquad
1-\frac{C\nu}{C_0\gamma}\geq\frac12.
\]
Therefore,
\[
\begin{aligned}
\mathcal E^{(t)}
-\E_t[\mathcal E^{(t+1)}]
&\geq
\frac{c_1\bar\eta}{2}\|G\|^2
+\frac{\gamma\zeta}{2}\|h\|^2\\
&\geq\mathcal D^{(t)},
\end{aligned}
\]
after choosing the universal constant in
$\mathcal D^{(t)}$ sufficiently small.

To establish the contraction estimate, global residual Lipschitz continuity and
$\zeta\geq C\bar\eta^2(k^{(t)})^2$
imply the comparisons in
\eqref{eq:stoch-app-comparison}.
Conversely, using the local Gram matrix lower bound, we obtain
\[
\begin{aligned}
\mathcal E^{(t)}
&\leq
2L(\theta^{(t)})+2\zeta\|h\|^2\\
&\leq
\frac{4}{\lambda_0}\|G\|^2
+2\zeta\|h\|^2.
\end{aligned}
\]
Since $\bar\eta\lambda_0\leq c\gamma$, this estimate and the definition of $\mathcal D^{(t)}$ give
\[
\mathcal D^{(t)}
\geq c\bar\eta\lambda_0\mathcal E^{(t)},
\]
completing the proof of \eqref{eq:stoch-app-drift}.

Finally, we prove the virtual iterate movement bounds.
From the conditional mean estimate, we have
\[
\begin{aligned}
\|\E_t[\Delta Z^{(t)}]\|^2
&=\bar\eta^2
\|\E_t[u_t(g^{(t)})]\|^2\\
&\leq
C\bar\eta^2
\left(
\|G\|^2+
\frac{\nu^2}{\gamma^2}\|h\|^2
\right)\\
&\leq C\bar\eta\mathcal D^{(t)}.
\end{aligned}
\]
The conditional variance bound in
\eqref{eq:stoch-app-speeds} follows directly from
\eqref{eq:stoch-app-update-variance} and
$\Delta Z^{(t)}=\bar\eta u_t(g^{(t)})$.
This completes the proof.
\end{proof}

\subsection{Completion of the Proof}

We first show that the parameter conditions in Theorem~\ref{thm:reg-stoch} imply the concentration condition~\eqref{eq:stoch-app-scalar-condition}. Let
\[
q=1-c_E\bar\eta\lambda_0,
\qquad
\Delta_q=1-q,
\]
where $c_E>0$ is a sufficiently small universal constant, and choose $\delta_v$ to be a fixed positive fraction of $\delta$. Since $\kappa\leq\rho_b$ and $\bar\eta\asymp\nu$, we have $\Delta_q\asymp\nu\lambda_0$. Hence, it suffices to show that
\[
\nu\rho_b\log\frac{C}{\nu\lambda_0\delta}
\]
is sufficiently small.

Define
\[
\Lambda=\log\frac{C\rho_b}{\gamma\lambda_0\delta}\geq1.
\]
For a sufficiently small universal constant $c_\nu>0$, the parameter condition~\eqref{eq:reg-stoch-parameters} gives
\[
\nu\leq\nu_*:=
\frac{c_\nu\gamma}{\rho_b\Lambda}.
\]
Since the function $x\mapsto x\log(C/(x\lambda_0\delta))$ is increasing on $(0,\nu_*]$, we obtain
\[
\begin{aligned}
\nu\rho_b\log\frac{C}{\nu\lambda_0\delta}
&\leq
\frac{c_\nu\gamma}{\Lambda}
\log\frac{C\rho_b\Lambda}
{c_\nu\gamma\lambda_0\delta}\\
&\leq
Cc_\nu\bigl(1+\log(1/c_\nu)\bigr).
\end{aligned}
\]
Thus, choosing $c_\nu$ sufficiently small ensures that~\eqref{eq:stoch-app-scalar-condition} holds for all admissible $\nu$, including arbitrarily small values.

The parameter conditions also imply
\[
\nu\leq\frac{c\gamma}{\rho_b},
\qquad
T_*\asymp\nu^{-1},
\qquad
\bar\eta T_*\leq C,
\qquad
q^{-T_*}\leq C,
\qquad
\Delta_q\leq\frac12\min\{\nu,\gamma\}.
\]
These bounds will be used in the stopping argument and the convergence analysis below.

\begin{proof}
Apply Lemma~\ref{lem:det-app-init} with failure probability $\delta/2$. We condition on an initialization satisfying
\[
K(\theta^{(0)})\succeq\frac{3\lambda_0}{4}I,
\qquad
L_0:=L(\theta^{(0)})\leq S_0:=C\log\frac{2}{\delta}.
\]
Throughout the following analysis, all probabilities and expectations are taken over minibatch sampling, conditional on this initialization. We work with the auxiliary model until the final step of the proof. If $L_0=0$, all stochastic gradients vanish, and the conclusion follows immediately. Henceforth, assume that $L_0>0$.

Define
\[
S=\frac{C_SS_0}{\delta},
\qquad
\mathcal R=c_0\sqrt{m\lambda_0},
\qquad
q=1-c_E\bar\eta\lambda_0,
\]
where $C_S>0$ is a sufficiently large universal constant and $c_E>0$ is a sufficiently small universal constant, chosen consistently with the preceding lemmas. Under the assumptions of Theorem~\ref{thm:reg-stoch}, we have
\[
\bar\epsilon\geq C\sqrt S,
\qquad
m\geq\frac{CS_0}{\delta\lambda_0^2},
\qquad
\bar\eta T_*\leq C,
\qquad
q^{-T_*}\leq C,
\qquad
\Delta_q:=1-q\leq\frac12\min\{\nu,\gamma\}.
\]
In particular, the parameter conditions of Lemmas~\ref{lem:stoch-app-scalar-cap} and~\ref{lem:stoch-app-energy} are satisfied.

\paragraph{Control of the initial iterations.}
By the bias-corrected first-moment representation~\eqref{eq:det-app-s-history} and the global bound
$\E_t\|g^{(t)}\|^2\leq CL(\theta^{(t)})$, we obtain
\[
\E\|\widehat s^{(t)}\|^2\leq CM_t,
\qquad
M_t:=\max_{0\leq s\leq t}\E[L(\theta^{(s)})].
\]
Using the Lipschitz continuity of the residual vector and $P^{(t)}\preceq I$, we obtain
\[
M_{t+1}^{1/2}
\leq(1+C\bar\eta)M_t^{1/2}.
\]
Since $\bar\eta T_*\leq C$, it follows that $M_t\leq CL_0$ for all $0\leq t\leq T_*$. Moreover, Minkowski's inequality gives
\[
\begin{aligned}
\left(\mathbb{E}\left(
\sum_{s=0}^{T_*-1}
\|\theta^{(s+1)}-\theta^{(s)}\|
\right)^2\right)^{1/2}
&\leq C\bar\eta T_*\sqrt{L_0}\leq C\sqrt{L_0}.
\end{aligned}
\]
Using the Lipschitz continuity of the residual vector along the initial trajectory, together with $\bar\eta/\gamma\leq C$ and $\zeta\leq C$, we further obtain
\begin{equation}
\label{eq:stoch-app-startup}
\begin{aligned}
&\E\left[
\max_{0\leq s\leq T_*}L(\theta^{(s)})
\right]
+\E[\mathcal E^{(T_*)}]+
\E\left[
\left(
\sum_{s=0}^{T_*-1}
\|\theta^{(s+1)}-\theta^{(s)}\|
\right)^2
\right]
\leq CL_0.
\end{aligned}
\end{equation}

Define the $\mathcal F_{T_*}$-measurable event
\[
\Pi=
\left\{
\max_{0\leq s\leq T_*}L(\theta^{(s)})\leq S,
\quad
\sum_{s=0}^{T_*-1}
\|\theta^{(s+1)}-\theta^{(s)}\|
\leq\frac{\mathcal R}{4}
\right\}.
\]
Applying Markov's inequality to \eqref{eq:stoch-app-startup} yields
\[
\mathbb P(\Pi^c)
\leq\frac{CL_0}{S}
+\frac{CL_0}{\mathcal R^2}.
\]

\paragraph{Stopping argument and energy control.}
On the event $\Pi$, define the stopping time $\tau$ as the first $t\geq T_*$ at which any of the following conditions is violated:
\begin{equation}
\label{eq:stoch-app-stopping}
\mathcal E^{(t)}\leq Sq^{t-T_*},
\qquad
U^{(t-1)}\leq C_US,
\qquad\text{or}\qquad
\|\theta^{(t)}-\theta^{(0)}\|\leq \frac{\mathcal R}{2},
\end{equation}
where $C_U>0$ is a sufficiently large universal constant. We set $\tau=T_*$ on $\Pi^c$ and $\tau=\infty$ if none of these conditions is violated.
Specifically, we let
\[
\tau=\inf\left\{
t\ge T_*:
\mathcal E^{(t)}>Sq^{t-T_*}
\ \text{or}\ 
U^{(t-1)}>C_US
\ \text{or}\ 
\|\theta^{(t)}-\theta^{(0)}\|>\mathcal{R}/2
\right\},
\]
and \(\inf\varnothing=\infty\).
Since all the stopping criteria are $\mathcal F_t$-measurable, $\tau$ is a stopping time. Define the predictable indicator
\[
I^{(t)}=\mathbf1_\Pi\mathbf1_{\{t<\tau\}},
\qquad t\geq T_*.
\]
On the event $\{I^{(t)}=1\}$, the global comparison in \eqref{eq:stoch-app-comparison} gives
\[
L(\theta^{(t)})
\leq2\mathcal E^{(t)}
\leq2Sq^{t-T_*},
\]
and the Gram matrix lower bound remains valid. Applying Lemma~\ref{lem:stoch-app-energy} with the loss bound $2S$, we obtain
\begin{equation}
\label{eq:stoch-app-stopped-drift}
\mathcal E^{(t)}
-\E_t[\mathcal E^{(t+1)}]
\geq\mathcal D^{(t)}
\geq(1-q)\mathcal E^{(t)}
\qquad\text{on }\{I^{(t)}=1\}.
\end{equation}

We first control the probability of violating the second-moment bound. 
Let $\delta_v=c_v\delta$, where $c_v>0$ is a sufficiently small universal constant. Applying Lemma~\ref{lem:stoch-app-scalar-cap} with loss bound $2S$ and failure probability $\delta_v$ to the predictable indicator

\[
\widetilde I^{(s)}
=
\begin{cases}
\mathbf1_{\{\max_{0\leq j\leq s}
L(\theta^{(j)})\leq S\}},
& s<T_*,\\
I^{(s)},&s\geq T_*.
\end{cases}
\]
By construction,
\[
\widetilde I^{(s)}L(\theta^{(s)})
\leq2Sq^{(s-T_*)_+}.
\]
On $\Pi$, all terms up to time $\tau-1$ are retained in the masked moving average. Consequently, at any finite stopping time $\tau$, the masked average coincides with $U^{(\tau-1)}$, including the case $\tau=T_*$. Choosing $C_U$ sufficiently large to accommodate the constant in Lemma~\ref{lem:stoch-app-scalar-cap}, we obtain
\[
\mathbb P\left(
\Pi,\ \tau<\infty,\
U^{(\tau-1)}>C_US
\right)
\leq \delta_v.
\]

Next, we control the probability of violating the energy bound. By \eqref{eq:stoch-app-stopped-drift}, the stopped process
\[
\mathbf1_\Pi
q^{-((t\wedge\tau)-T_*)}
\mathcal E^{(t\wedge\tau)}
\]
is a nonnegative supermartingale. Applying the maximal inequality in Lemma~\ref{lem:aux-doob}, together with the initial estimate \eqref{eq:stoch-app-startup}, we have
\[
\mathbb P\left(
\Pi,\ \tau<\infty,\
\mathcal E^{(\tau)}>Sq^{\tau-T_*}
\right)
\leq\frac{CL_0}{S}.
\]

Finally, we establish a contraction estimate for the expected energy before the stopping time. Define
\[
A_t=\E[I^{(t)}\mathcal E^{(t)}].
\]
Since $\mathcal E^{(t+1)}\geq0$ and
$I^{(t+1)}\leq I^{(t)}$, the conditional descent estimate \eqref{eq:stoch-app-stopped-drift} implies
\begin{equation}
\label{eq:stoch-app-killed-energy}
\begin{aligned}
\Delta_t:=A_t-A_{t+1}
&\geq\E[I^{(t)}\mathcal D^{(t)}]\\
&\geq(1-q)A_t.
\end{aligned}
\end{equation}
Consequently,
\[
A_t\leq q^{t-T_*}A_{T_*}
\leq Cq^{t-T_*}L_0,
\qquad t\geq T_*,
\]
where the last inequality follows from \eqref{eq:stoch-app-startup}.

\paragraph{Control of parameter movement.}
We decompose the virtual iterate increment into its predictable and martingale components:
\[
Z^{(t+1)}-Z^{(t)}
=W^{(t)}+\xi^{(t+1)},
\]
where
\[
W^{(t)}
=\bar\eta\E_t[u_t(g^{(t)})],
\qquad
\E_t[\xi^{(t+1)}]=0.
\]
The conditional movement bound in Lemma~\ref{lem:stoch-app-energy} yields
\[
\E[I^{(t)}\|W^{(t)}\|^2]
\leq C\bar\eta\Delta_t.
\]
The telescoping argument used in the deterministic proof also applies to the sequence $A_t$, giving
\[
\sqrt{\Delta_t}
\leq\frac{2}{\sqrt{1-q}}
\left(\sqrt{A_t}-\sqrt{A_{t+1}}\right).
\]
Consequently, Minkowski's inequality and
\eqref{eq:stoch-app-killed-energy} imply
\[
\begin{aligned}
\E\left[
\left(
\sum_{t\geq T_*}I^{(t)}
\|W^{(t)}\|
\right)^2
\right]
&\leq\frac{C\bar\eta L_0}{1-q}\leq\frac{CL_0}{\lambda_0}.
\end{aligned}
\]

For the martingale increments, Lemma~\ref{lem:stoch-app-energy} further gives
\[
\begin{aligned}
\sum_{t\geq T_*}
\E[I^{(t)}\|\xi^{(t+1)}\|^2]
&\leq
\frac{C\bar\eta^2(\rho_b-1)}{\gamma}
\sum_{t\geq T_*}
\E[I^{(t)}\|G^{(t)}\|^2]\\
&\leq
\frac{C\bar\eta(\rho_b-1)}{\gamma}L_0\\
&\leq CL_0.
\end{aligned}
\]
Applying the martingale maximal inequality in Lemma~\ref{lem:aux-doob} to the stopped martingale, and combining the resulting bound with the initial movement estimate and the global comparison
$\|\theta^{(t)}-Z^{(t)}\|^2\leq C\mathcal E^{(t)}$, we obtain
\begin{equation}
\label{eq:stoch-app-stopped-path}
\E\left[
\mathbf1_\Pi
\sup_{t\geq T_*}
\|Z^{(t\wedge\tau)}-\theta^{(0)}\|^2
\right]
\leq\frac{CL_0}{\lambda_0}.
\end{equation}
The preceding estimates can first be established over finite horizons and then extended to the infinite horizon. In particular, the stopped sums include the update from $\tau-1$ to $\tau$.

We next bound the probability of leaving the region of Gram matrix stability. Consider a finite stopping time $\tau$ at which the parameter movement condition is violated while the energy condition remains satisfied. By the global comparison in \eqref{eq:stoch-app-comparison} and the network width assumption,
\[
\|\theta^{(\tau)}-Z^{(\tau)}\|
\leq C\sqrt S
\leq\frac{\mathcal R}{4}.
\]
Therefore, such an exit necessarily implies
\[
\|Z^{(\tau)}-\theta^{(0)}\|
>\frac{\mathcal R}{4}.
\]
Applying Markov's inequality to
\eqref{eq:stoch-app-stopped-path}, the probability of this event is bounded by $\frac{CL_0}{\lambda_0\mathcal R^2}.$

Combining the preceding estimates for the initialization phase, energy, second-moment history, and parameter movement, we define the event
\[
\mathcal G_0=\Pi\cap\{\tau=\infty\}.
\]
Its failure probability satisfies
\[
\begin{aligned}
\mathbb P(\mathcal G_0^c)
&\leq
\frac{CL_0}{S}
+\frac{CL_0}{\mathcal R^2}
+\frac{CL_0}{\lambda_0\mathcal R^2}
+\delta_v\\
&\leq\frac{\delta}{4},
\end{aligned}
\]
where the last inequality follows by choosing $C_S$ and the universal constant in the network width requirement sufficiently large, with $\delta_v=c_v\delta$ for a sufficiently small constant $c_v>0$.

On $\mathcal G_0$, the stopping time is infinite, so $I^{(t)}=1$ for all $t\geq T_*$. Hence, the energy estimate \eqref{eq:stoch-app-killed-energy}, together with $L(\theta^{(t)})\leq2\mathcal E^{(t)}$, yields
\begin{equation}
\label{eq:stoch-app-restricted-loss}
\E\left[
L(\theta^{(t)})\mathbf1_{\mathcal G_0}
\right]
\leq Cq^tL_0,
\qquad t\geq T_*.
\end{equation}
For $t<T_*$, the same bound follows from
\eqref{eq:stoch-app-startup} and $q^{-T_*}\leq C$. Thus,
\eqref{eq:stoch-app-restricted-loss} holds for all $t\geq0$.

\paragraph{Return to the original network.}
It remains to control the movement of individual neurons, particularly their output coefficients, to establish agreement between the auxiliary and original Adam trajectories. Define the predictable indicator
\[
\widehat I^{(s)}
=
\begin{cases}
1, & s<T_*,\\
I^{(s)}, & s\geq T_*.
\end{cases}
\]
By the initial-phase estimate~\eqref{eq:stoch-app-startup} and the stopped energy bound~\eqref{eq:stoch-app-killed-energy}, we have
\[
\E\left[
\widehat I^{(s)}L(\theta^{(s)})
\right]
\leq CL_0q^s,
\qquad s\geq0.
\]
Together with $\E_s[Y^{(s)}]=2L(\theta^{(s)})$, this implies
\[
\left(\E\left\|
\widehat I^{(s)}\sqrt{Y^{(s)}}
\right\|^2\right)^{1/2}
\leq C\sqrt{L_0}\,q^{s/2}.
\]
Define
\[
\mathcal T
=\bar\eta
\sum_{t=0}^{\infty}
\sum_{s=0}^t
\frac{\gamma\beta_1^{t-s}}
{1-\beta_1^{t+1}}
\widehat I^{(s)}\sqrt{Y^{(s)}}.
\]
Applying the geometric-weight estimate from the proof of Lemma~\ref{lem:det-app-moments}, together with Minkowski's inequality, yields
\[
\begin{aligned}
\left(\E\|\mathcal T\|^2\right)^{1/2}
&\leq
C\bar\eta\sqrt{L_0}
\sum_{t=0}^{\infty}q^{t/2}\\
&\leq
\frac{C\sqrt{L_0}}{\lambda_0}.
\end{aligned}
\]

On the event $\mathcal G_0$, all the indicators $\widehat I^{(s)}$ equal one. Therefore, the minibatch block-gradient bound in \eqref{eq:stoch-app-batch-blocks}, the bias-corrected first-moment representation~\eqref{eq:det-app-s-history}, and $P^{(t)}\preceq I$ imply
\[
\sum_{t=0}^{\infty}
\max_{1\leq r\leq m}
\|\theta_r^{(t+1)}-\theta_r^{(t)}\|
\leq
\frac{C\mathcal T}{\sqrt m}
\qquad\text{on }\mathcal G_0.
\]
For a sufficiently large universal constant $C_T>0$, define
\[
\mathcal G
=
\mathcal G_0
\cap
\left\{
\mathcal T
\leq
C_T\sqrt{\frac{L_0}{\delta\lambda_0^2}}
\right\}.
\]
By Markov's inequality and the preceding second-moment bound on $\mathcal T$, we obtain
\[
\mathbb P(\mathcal G)\geq1-\frac{\delta}{2}.
\]
On $\mathcal G$, the cumulative neuron-wise movement satisfies
\begin{equation}
\label{eq3}
    \begin{aligned}
\sum_{t=0}^{\infty}
\max_{1\leq r\leq m}
\|\theta_r^{(t+1)}-\theta_r^{(t)}\|
&\leq
C\sqrt{\frac{L_0}{m\delta\lambda_0^2}}\\
&\leq\frac12,
\end{aligned}
\end{equation}
where the last inequality follows from the prescribed network width condition.

Restoring the auxiliary notation, we conclude that
\[
|a_{\dagger,r}^{(t)}|\leq\frac32,
\qquad
\text{for all }r\text{ and }t
\quad\text{on }\mathcal G.
\]
Hence, the auxiliary iterates remain in $\mathcal S$ throughout training. By the trajectory agreement established in Subsection~\ref{subsec:stoch-auxiliary-network}, the auxiliary and original Adam algorithms therefore generate identical iterates on $\mathcal G$:
\[
\theta_\dagger^{(t)}=\theta^{(t)},
\qquad
L^\dagger(\theta_\dagger^{(t)})
=L(\theta^{(t)}),
\qquad t\geq0.
\]
Since $\mathcal G\subseteq\mathcal G_0$, the restricted expected-loss bound~\eqref{eq:stoch-app-restricted-loss} for the auxiliary process transfers to the original Adam iterates:
\[
\begin{aligned}
\E\left[
L(\theta^{(t)})\mathbf1_{\mathcal G}
\right]
&=
\E\left[
L^\dagger(\theta_\dagger^{(t)})
\mathbf1_{\mathcal G}
\right]\\
&\leq
Cq^tL(\theta^{(0)}),
\qquad t\geq0.
\end{aligned}
\]
Recalling that all probabilities and expectations are conditional on the fixed initialization, we obtain \eqref{eq:reg-stoch-rate}. Combining the initialization and sampling probabilities, the joint success probability is at least
\[
\left(1-\frac{\delta}{2}\right)^2
\geq1-\delta.
\]

We then verify the parameter requirements. The initialization estimate requires
\[
m\gtrsim\lambda_0^{-1}\log\frac{n}{\delta},
\]
up to universal constants, while the preceding trajectory control requires
\[
m\gtrsim\frac{S_0}{\delta\lambda_0^2},
\qquad
\bar\epsilon\gtrsim\sqrt{\frac{S_0}{\delta}}.
\]
Since $S_0\asymp\log(2/\delta)$, these conditions are ensured by the width and damping requirements in Theorem~\ref{thm:reg-stoch}.

Finally, \eqref{eq3} implies that, almost surely on $\mathcal G$,
\[
\sum_{t=0}^{\infty}
\max_{1\leq r\leq m}
\left(
|a_r^{(t+1)}-a_r^{(t)}|
+\|w_r^{(t+1)}-w_r^{(t)}\|
\right)<\infty.
\]
Since $m$ is finite, the complete parameter sequence converges to a finite limit $\theta^{\star}$ on $\mathcal G$.
Moreover, by Fatou's lemma, continuity of $L$, and the restricted expected-loss bound, we have
\[
\begin{aligned}
\mathbb E\left[
L(\theta^{\star})\bm{1}_{\mathcal G}
\mid\theta^{(0)}
\right]
&\leq
\liminf_{t\to\infty}
\mathbb E\left[
L(\theta^{(t)})\bm{1}_{\mathcal G}
\mid\theta^{(0)}
\right]
=0.
\end{aligned}
\]
Thus, $L(\theta^{\star})=0$ almost surely on $\mathcal G$, which completes the proof.

\end{proof}

\newpage

\section{Proof of Theorem~\ref{thm:pinn-det}}
\label{app:pinn-det-proof}

We use the PINN residual formulation and the associated Gram matrix defined in Section~\ref{sec:pinn}. We first establish the differential feature representations, initialization bounds, and regularity properties required for the PINN analysis. We then extend the moment bounds and Lyapunov-based descent analysis from the deterministic regression analysis to the PINN setting. Finally, we use these bounds to verify the required local conditions and complete the convergence proof by induction.

\subsection{Differential Features and Positive Definiteness of the Limiting Gram Matrix}
\label{app:pinn-gram-proof}
Let $Q=\operatorname{diag}(I_d,0)\in\R^{(d+1)\times(d+1)}$. For any $w\in\R^{d+1}$ and input $x\in\R^d$, define
\[
s=w^\top\widetilde x,
\qquad
h(w)=\frac12w^\top Qw.
\]
The PDE feature and its first two derivatives with respect to $w$ are given by
\begin{equation}
\label{eq:pinn-app-feature-derivatives}
\begin{aligned}
\phi_{\rm P}(w;x)
&=h(w)\sigma''(s),\\
\nabla_w\phi_{\rm P}(w;x)
&=Qw\,\sigma''(s)
+h(w)\sigma'''(s)\widetilde x,\\
\nabla_w^2\phi_{\rm P}(w;x)
&=Q\sigma''(s)
+\bigl(Qw\widetilde x^\top+\widetilde xw^\top Q\bigr)
\sigma'''(s)
+h(w)\sigma''''(s)
\widetilde x\widetilde x^\top.
\end{aligned}
\end{equation}
For the boundary feature $\phi_{\rm B}(w;z)=\sigma(w^\top\widetilde z)$, we have
\[
\nabla_w\phi_{\rm B}(w;z)
=\sigma'(w^\top\widetilde z)\widetilde z,
\qquad
\nabla_w^2\phi_{\rm B}(w;z)
=\sigma''(w^\top\widetilde z)
\widetilde z\widetilde z^\top.
\]
Since both layers are trained, we also use the identities
\begin{equation}
\label{eq:pinn-app-neuron-derivatives}
\nabla_{(a,w)}(a\phi)
=\begin{pmatrix}
\phi\\
a\nabla_w\phi
\end{pmatrix},
\qquad
\nabla_{(a,w)}^2(a\phi)
=\begin{pmatrix}
0&(\nabla_w\phi)^\top\\
\nabla_w\phi&a\nabla_w^2\phi
\end{pmatrix}.
\end{equation}
Since $\sigma=\tanh$ has bounded derivatives up to order four, the PDE feature and its first two derivatives are bounded in norm by $C(1+\|w\|^2)$. The corresponding boundary feature bounds are of order one. Consequently, the first- and second-order derivatives of the single-neuron contributions satisfy the same bounds when $|a|\leq2$. These constants do not depend on the input dimension $d$, since $\|Q\|=1$ and all augmented inputs have norm at most one.

For each $w\in\R^{d+1}$, define the stacked feature vector
\begin{equation}
\label{eq1}
    \Phi(w)=
\begin{pmatrix}
\bigl(\phi_{\rm P}(w;x_i)/\sqrt{n_{\rm PDE}}\bigr)_{i=1}^{n_{\rm PDE}}\\
\bigl(\sqrt{\alpha_{\rm BC}/n_{\rm BC}}\,
\phi_{\rm B}(w;z_j)\bigr)_{j=1}^{n_{\rm BC}}
\end{pmatrix}
\in\R^N.
\end{equation}
Denote its Jacobian with respect to $w$ by
$\nabla_w\Phi(w)\in\R^{N\times(d+1)}$.
Then the limiting Gram matrix admits the representation
\begin{equation}
\label{eq:pinn-kernel}
K_\infty
=\E_w\left[
\Phi(w)\Phi(w)^\top
+\nabla_w\Phi(w)\nabla_w\Phi(w)^\top
\right],
\end{equation}
where $w\sim\mathcal N(0,I_{d+1})$, since $(a_r^{(0)})^2 = 1$.

\begin{proof}[Proof of Lemma~\ref{lem:pinn-positivity}]
Since the remaining contribution to $K_\infty$ is positive semidefinite, it suffices to establish the positive definiteness of $\E[\Phi(w)\Phi(w)^\top]$.

Suppose that a coefficient vector $(c_1,\ldots,c_{n_{\rm PDE}},
d_1,\ldots,d_{n_{\rm BC}})\in\R^N$ belongs to the null space of $\E_w[\Phi(w)\Phi(w)^\top]$. 
Then the corresponding linear combination of the features vanishes almost surely under the Gaussian distribution. Since the features are continuous in $w$ and the Gaussian distribution has full support on $\R^{d+1}$, this identity holds for every $w\in\R^{d+1}$. Absorbing the positive row normalization factors into the coefficients, we obtain
\[
\sum_{i=1}^{n_{\rm PDE}}
c_i\frac{w^\top Qw}{2}
\sigma''(w^\top\widetilde x_i)
+
\sum_{j=1}^{n_{\rm BC}}
d_j\sigma(w^\top\widetilde z_j)
=0,
\qquad \forall w\in\R^{d+1},
\]
where $Q=\operatorname{diag}(I_d,0)$.

We first show that all PDE coefficients vanish. Since the interior collocation points are pairwise distinct, there exists a unit vector $e\in\R^d$ such that $e^\top x_1,\ldots,e^\top x_{n_{\rm PDE}}$ are pairwise distinct. Such a vector can be chosen outside the finite union of proper hyperplanes
\[
\bigcup_{i<j}
\left\{
e\in\R^d:
e^\top(x_i-x_j)=0
\right\}.
\]
Fix any $k\in\{1,\ldots,n_{\rm PDE}\}$ and, for $A>0$ and $s\in\R$, define
\[
w_A(s)
=
\sqrt{2}
\begin{pmatrix}
Ae\\
s-Ae^\top x_k
\end{pmatrix}
\in\R^{d+1}.
\]
By construction,
\[
w_A(s)^\top\widetilde x_i
=Ae^\top(x_i-x_k)+s,
\qquad
\frac{w_A(s)^\top Qw_A(s)}{2}=A^2.
\]
Substituting $w=w_A(s)$ into the preceding identity and dividing by $A^2$, we obtain
\[
\begin{aligned}
&\sum_{i=1}^{n_{\rm PDE}}
c_i\sigma''\bigl(Ae^\top(x_i-x_k)+s\bigr)+
\frac{1}{A^2}
\sum_{j=1}^{n_{\rm BC}}
d_j\sigma\bigl(Ae^\top(z_j-x_k)+s\bigr)
=0.
\end{aligned}
\]
Since $\sigma=\tanh$ is bounded and $\sigma''(s)\to0$ as $|s|\to\infty$, letting $A\to\infty$ yields
\[
c_k\sigma''(s)=0,
\qquad \forall s\in\R.
\]
As $\sigma''$ is not identically zero, we conclude that $c_k=0$. Because $k$ was arbitrary, all PDE coefficients vanish.

It remains to show that the boundary coefficients vanish. The preceding identity now reduces to
\[
\sum_{j=1}^{n_{\rm BC}}
d_j\sigma(w^\top\widetilde z_j)=0,
\qquad \forall w\in\R^{d+1}.
\]
Since the boundary collocation points are pairwise distinct, we may similarly choose a unit vector $e\in\R^d$ such that their projections onto $e$ are pairwise distinct. Fix $k\in\{1,\ldots,n_{\rm BC}\}$ and define
\[
w_A(s)
=
\sqrt{2}
\begin{pmatrix}
Ae\\
s-Ae^\top z_k
\end{pmatrix}.
\]
Substituting this choice into the boundary identity gives
\[
\sum_{j=1}^{n_{\rm BC}}
d_j\sigma\bigl(Ae^\top(z_j-z_k)+s\bigr)=0.
\]
Differentiating with respect to $s$, we obtain
\[
\sum_{j=1}^{n_{\rm BC}}
d_j\sigma'\bigl(Ae^\top(z_j-z_k)+s\bigr)=0.
\]
For any fixed $s$, letting $A\to\infty$ and using
$\sigma'(s)\to0$ as $|s|\to\infty$ yields
\[
d_k\sigma'(s)=0.
\]
Since $\sigma'(0)=1$, we conclude that $d_k=0$. As $k$ was arbitrary, all boundary coefficients vanish. Therefore, the original coefficient vector is zero, proving that $\E[\Phi(w)\Phi(w)^\top]$ and hence $K_\infty$ are positive definite.
\end{proof}

\subsection{Key Lemmas}

\begin{lemma}[Initialization]
\label{lem:pinn-det-init}
Let $S_0=Cd^2\log(2/\delta)$ and $0 < \delta < 1$, where $C>0$ is a sufficiently large universal constant. Suppose that
\[
m\gtrsim\frac{\ell_m}{\lambda_0}
\log\frac{N}{\delta},
\]
where $\ell_m$ is defined in~\eqref{eq_l_m}.
Then, with probability at least $1-\delta$ over the random initialization, we have
\begin{equation}
\label{eq:pinn-app-init}
\begin{gathered}
K(\theta^{(0)})\succeq\frac{3\lambda_0}{4}I,
\qquad
L(\theta^{(0)})\leq S_0,\\
\max_{1\leq r\leq m}
\left(1+\|w_r^{(0)}\|^2\right)^2
\leq C\ell_m,\\
\frac1m\sum_{r=1}^m
\left(1+\|w_r^{(0)}\|^2\right)^2
\leq Cd^2.
\end{gathered}
\end{equation}
\end{lemma}

\begin{proof}
For $w\sim\mathcal N(0,I_{d+1})$, the Gaussian moment function and Markov's inequality yield
\[
\mathbb P(\|w\|>B)
\leq 2^{(d+1)/2}\exp\left(-\frac{B^2}{4}\right).
\]
Choose
\[
B^2=C_B\left(
d+\log\frac{C_*md^2}{\delta\lambda_0}
\right),
\]
where $C_B$ and $C_*$ are sufficiently large universal constants. By the definition of $\ell_m$, we have
\[
(1+B^2)^2\leq C\ell_m.
\]
Moreover, a union bound gives
\[
\mathbb P\left(
\max_{1\leq r\leq m}\|w_r^{(0)}\|>B
\right)
\leq\frac{\delta}{4}.
\]

Since the output weights are initialized from the Rademacher distribution, the Gram matrix at initialization admits the decomposition
\[
K(\theta^{(0)})
=\frac1m\sum_{r=1}^m Z(w_r^{(0)}),
\]
where
\[
Z(w)
=\Phi(w)\Phi(w)^\top
+\nabla_w\Phi(w)\nabla_w\Phi(w)^\top.
\]
The differential feature bounds established above, together with the normalization of the residual rows, imply
\[
0\preceq Z(w),
\qquad
\|Z(w)\|
\leq\operatorname{trace}(Z(w))
\leq C(1+\|w\|^2)^2.
\]
Consequently, the Gaussian moment bounds yield
\[
\E\|Z(w)\|^2\leq Cd^4.
\]
In particular, these estimates do not introduce an additional dependence on the number of collocation points $N$.

To apply the matrix Chernoff inequality, define the truncated matrix
\[
Z^B(w)
=Z(w)\mathbf1_{\{\|w\|\leq B\}}.
\]
By Cauchy--Schwarz and the Gaussian tail bound,
\[
\begin{aligned}
\|K_\infty-\E[Z^B(w)]\|
&\leq
\E\left[
\|Z(w)\|\mathbf1_{\{\|w\|>B\}}
\right]\\
&\leq
\bigl(\E\|Z(w)\|^2\bigr)^{1/2}
\mathbb P(\|w\|>B)^{1/2}\\
&\leq
Cd^2\mathbb P(\|w\|>B)^{1/2}\\
&\leq\frac{\lambda_0}{16},
\end{aligned}
\]
where the last inequality follows by choosing the constants in $B$ sufficiently large. Since $K_\infty\succeq\lambda_0 I$, we obtain
\[
\E[Z^B(w)]\succeq\frac{15\lambda_0}{16}I.
\]
Moreover, the truncation ensures that
\[
0\preceq Z^B(w)\preceq C\ell_m I.
\]
By applying Lemma~\ref{lem:aux-matrix-chernoff} to the independent truncated matrices $Z^B(w_r^{(0)})$, with relative deviation $1/5$, we have
\[
\begin{aligned}
&\mathbb P\left(
\lambda_{\min}\left(
\frac1m\sum_{r=1}^m Z^B(w_r^{(0)})
\right)
<\frac{3\lambda_0}{4}
\right)\\
&\qquad\leq
N\exp\left(
-\frac{cm\lambda_0}{\ell_m}
\right)
\leq\frac{\delta}{4},
\end{aligned}
\]
where the last inequality follows from the given network width condition.

Finally, since $Z(w)\succeq Z^B(w)$ for every $w$, the same lower bound holds for the original Gram matrix $K(\theta^{(0)})$. We emphasize that the truncated matrices remain independent, as the concentration argument does not condition on the event controlling the maximum of the initialized weights.

Next, we establish the bound on the averaged Gaussian weights. Define the truncated random variables
\[
T_r
=\left(1+\|w_r^{(0)}\|^2\right)^2
\mathbf1_{\{\|w_r^{(0)}\|\leq B\}},
\qquad r=1,\ldots,m.
\]
Gaussian moment bounds imply
\[
\E[T_r]\leq Cd^2,
\qquad
\operatorname{Var}(T_r)\leq Cd^4,
\]
while the truncation ensures that
$0\leq T_r\leq C\ell_m$.
Applying the Bernstein inequality in
Lemma~\ref{lem:aux-martingale-bernstein}
to the independent centered variables
$T_r-\E[T_r]$, we obtain
\[
\mathbb P\left(
\frac1m\sum_{r=1}^m T_r>Cd^2
\right)
\leq
\exp\left(-\frac{cmd^2}{\ell_m}\right)
\leq\frac{\delta}{4}.
\]
Here, we used $\ell_m\geq d^2$ and the prescribed width condition, which implies $m\gtrsim\frac{\ell_m}{d^2} \log\frac{C}{\delta},$ since $\lambda_0\leq\alpha_{\rm BC}$, the boundary penalty is fixed, and $N\geq2$. On the event $\max_r\|w_r^{(0)}\|\leq B$, the truncation is inactive, so
\[
T_r=\left(1+\|w_r^{(0)}\|^2\right)^2
\]
for every $r$.

Finally, we bound the initial loss. Fix the hidden weights satisfying the averaged bound established above, and define the normalized prediction vector
\[
F^{(0)}
=\frac1{\sqrt m}\sum_{r=1}^m
a_r^{(0)}\Phi(w_r^{(0)}),
\]
where $\Phi(\cdot)$ is defined in \eqref{eq1}.
Since the output weights are independent Rademacher random variables, conditioning on the hidden weights gives
\[
\E_a\|F^{(0)}\|^2
=\frac1m\sum_{r=1}^m
\|\Phi(w_r^{(0)})\|^2
\leq Cd^2,
\]
where $\E_a$ denotes expectation over the output weights $(a_1,a_2,\cdots,a_m)$.

Changing a single output weight $a_r^{(0)}$ changes $\|F^{(0)}\|$ by at most $\frac{2}{\sqrt m}\|\Phi(w_r^{(0)})\|$. The sum of the squares of these bounded-difference constants is therefore at most $Cd^2$. Applying Lemma~\ref{lem:aux-bounded-differences}, together with
$\E_a\|F^{(0)}\|\leq Cd$, we have
\[
\|F^{(0)}\|
\leq Cd\sqrt{\log\frac{2}{\delta}}
\]
with conditional probability at least $1-\delta/4$.

Let $y_{\rm PINN}$ denote the normalized target vector corresponding to the PDE source term and boundary data, then
\[
\|y_{\rm PINN}\|^2
=
\frac1{n_{\rm PDE}}\sum_{i=1}^{n_{\rm PDE}}f(x_i)^2
+
\frac{\alpha_{\rm BC}}{n_{\rm BC}}
\sum_{j=1}^{n_{\rm BC}}\varphi(z_j)^2
\leq1+\alpha_{\rm BC}.
\]
Consequently,
\[
\begin{aligned}
L(\theta^{(0)})
&=\frac12\|F^{(0)}-y_{\rm PINN}\|^2\\
&\leq
\frac12\left(
\|F^{(0)}\|+\|y_{\rm PINN}\|
\right)^2\\
&\leq Cd^2\log\frac{2}{\delta}
=S_0.
\end{aligned}
\]
Combining the probability bounds for the Gaussian initialization, the initial Gram matrix, and the initial loss by a union bound completes the proof.
\end{proof}

\begin{lemma}[Regularity and Gram matrix stability]
\label{lem:pinn-det-geometry}
On the initialization event~\eqref{eq:pinn-app-init}, define the convex set
\[
\mathcal S_{\rm P}
=\left\{
\theta:\ |a_r|\leq2,\quad
\|w_r-w_r^{(0)}\|\leq1,\quad
1\leq r\leq m
\right\}.
\]
Then, for any $\theta,\vartheta\in\mathcal S_{\rm P}$, we have
\begin{align}
\|J(\theta)\|^2
&\leq Cd^2,
&
\|J(\theta)-J(\vartheta)\|
&\leq C\sqrt{\frac{\ell_m}{m}}
\|\theta-\vartheta\|,
\label{eq:pinn-app-jacobian}\\
\|\nabla_{\theta_r}L(\theta)\|
&\leq C_g\sqrt{\frac{\ell_m L(\theta)}{m}},
&
\|\nabla^2L(\theta)\|
&\leq Cd^2+
C\sqrt{\frac{\ell_m}{m}}\|R(\theta)\|,
\label{eq:pinn-app-regularity}
\end{align}
where the block-gradient bound holds for all $1\leq r\leq m$.

Furthermore, let
\[
\mathcal R=c_0\sqrt{\frac{m\lambda_0}{\ell_m}},
\]
where $c_0>0$ is a sufficiently small universal constant. Then, every $\theta\in\mathcal S_{\rm P}$ satisfying
$\|\theta-\theta^{(0)}\|\leq\mathcal R$ obeys
\begin{equation}
\label{eq:pinn-app-local-kernel}
K(\theta)\succeq\frac{\lambda_0}{4}I,
\qquad
\|\nabla L(\theta)\|^2
\geq\frac{\lambda_0}{2}L(\theta).
\end{equation}
\end{lemma}

\begin{proof}
On $\mathcal S_{\rm P}$, the feature derivatives associated with neuron $r$ are bounded by $C(1+\|w_r^{(0)}\|^2)$. Accounting for the network scaling $m^{-1/2}$ and the normalization of the residual rows, we obtain
\[
\begin{aligned}
\|J(\theta)\|^2
&\leq\|J(\theta)\|_{\rm F}^2\\
&\leq\frac{C}{m}\sum_{r=1}^m
(1+\|w_r^{(0)}\|^2)^2
\leq Cd^2.
\end{aligned}
\]
Since $\mathcal S_{\rm P}$ is convex, integrating the neuron-wise Hessian bounds along the segment joining $\theta$ and $\vartheta$ and summing over the residual rows yields
\[
\begin{aligned}
\|J(\theta)-J(\vartheta)\|^2
&\leq\frac{C}{m}\sum_{r=1}^m
(1+\|w_r^{(0)}\|^2)^2
\|\theta_r-\vartheta_r\|^2\\
&\leq\frac{C\ell_m}{m}
\|\theta-\vartheta\|^2.
\end{aligned}
\]

Let $J_r(\theta)$ denote the column block of $J(\theta)$ corresponding to neuron $r$. Since
$\nabla_{\theta_r}L(\theta)=J_r(\theta)^\top R(\theta)$ and
\[
\|J_r(\theta)\|
\leq\frac{C(1+\|w_r^{(0)}\|^2)}{\sqrt m},
\]
the neuron-wise gradient bound follows. Moreover,
\[
\nabla^2L(\theta)
=J(\theta)^\top J(\theta)
+\sum_{i=1}^N R_i(\theta)\nabla^2R_i(\theta).
\]
The second term is block diagonal with respect to the neuron parameters. By applying Cauchy--Schwarz within the two residual groups, each block has operator norm at most
\[
\frac{C(1+\|w_r^{(0)}\|^2)}{\sqrt m}
\|R(\theta)\|.
\]
Combining this estimate with \eqref{eq:pinn-app-jacobian} establishes the Hessian bound in \eqref{eq:pinn-app-regularity}.

Finally, for any unit vector $z\in\R^N$, the initialization bound and the Jacobian Lipschitz estimate imply
\[
\begin{aligned}
\|J(\theta)^\top z\|
&\geq\|J(\theta^{(0)})^\top z\|
-\|J(\theta)-J(\theta^{(0)})\|\\
&\geq\sqrt{\frac{3\lambda_0}{4}}
-C\sqrt{\frac{\ell_m}{m}}
\|\theta-\theta^{(0)}\|\\
&\geq\frac{\sqrt{\lambda_0}}{2},
\end{aligned}
\]
provided that $c_0$ is sufficiently small. This yields
$K(\theta)\succeq\lambda_0 I/4$. The gradient inequality then follows from
$\nabla L(\theta)=J(\theta)^\top R(\theta)$ and
$\|R(\theta)\|^2=2L(\theta)$.
\end{proof}

Let $\gamma=1-\beta_1$, $L_0=L(\theta^{(0)})$, and $P^{(t)}=\epsilon D^{(t)}$, as in the deterministic regression analysis. The bias-corrected moment representations remain unchanged. The following lemma extends the moment bounds and Lyapunov descent estimates in Lemmas~\ref{lem:det-app-moments} and~\ref{lem:det-app-energy} to the PINN setting, using the regularity properties established in Lemma~\ref{lem:pinn-det-geometry}. The required smoothness and Gram matrix conditions will be verified in the final induction argument.

\begin{lemma}[Moment bounds and Lyapunov descent for PINNs]
\label{lem:pinn-det-adam}
Work on the initialization event~\eqref{eq:pinn-app-init}. Let $H_{\rm P}\geq\max\{1,\lambda_0\}$, and suppose that
\[
0<\bar\eta\leq\frac{\gamma}{8H_{\rm P}},
\qquad
q=1-\frac{\bar\eta\lambda_0}{32}.
\]
Then $q\in(0,1)$ and $1-q\leq\gamma/2$.

Suppose that $\theta^{(0)},\ldots,\theta^{(T)}\in\mathcal S_{\rm P}$. If $L(\theta^{(k)})\leq S$ for all $0\leq k\leq T$, then
\begin{equation}
\label{eq:pinn-det-metric}
P^{(t)}\succeq
\left(
1+\frac{C_g\sqrt{\ell_m S}}{\bar\epsilon}
\right)^{-1}I,
\qquad 0\leq t\leq T.
\end{equation}
If, in addition,
\[
L(\theta^{(k)})\leq q^kL_0,
\qquad 0\leq k\leq T,
\]
then
\begin{align}
\|\widehat s_r^{(t)}\|
&\leq
2C_g\sqrt{\frac{\ell_m L_0}{m}}\,q^{t/2},
\qquad 0\leq t\leq T,
\label{eq:pinn-det-moment}\\
\sum_{t=0}^T
\max_{1\leq r\leq m}
\|\theta_r^{(t+1)}-\theta_r^{(t)}\|
&\leq
\frac{128C_g}{\lambda_0}
\sqrt{\frac{\ell_m L_0}{m}}.
\label{eq:pinn-det-neuron-movement}
\end{align}
In particular, the latter bound includes the update from iteration $T$ to $T+1$.

Next, define the tracking error and Lyapunov function by
\[
e^{(t)}=\widehat s^{(t)}-g^{(t)},
\qquad
\mathcal E^{(t)}
=L(\theta^{(t)})
+\frac{4\bar\eta}{\gamma}\|e^{(t)}\|^2,
\qquad
Q^{(t)}=(g^{(t)})^\top P^{(t)}g^{(t)}.
\]
For any iteration $t$ satisfying
\[
\sup_{\theta\in[\theta^{(t)},\theta^{(t+1)}]}
\|\nabla^2L(\theta)\|\leq H_{\rm P},
\qquad
K(\theta^{(t)})\succeq\frac{\lambda_0}{4}I,
\qquad
\frac12I\preceq P^{(t)}\preceq I,
\]
the following estimates hold:
\begin{align}
\Delta^{(t)}
:=\mathcal E^{(t)}-\mathcal E^{(t+1)}
&\geq
\frac{\bar\eta}{4}Q^{(t)}
+\bar\eta\|e^{(t)}\|^2,
\label{eq:pinn-det-dissipation}\\
\mathcal E^{(t+1)}
&\leq q\mathcal E^{(t)},
\qquad
\|\theta^{(t+1)}-\theta^{(t)}\|^2
\leq8\bar\eta\Delta^{(t)}.
\label{eq:pinn-det-energy-step}
\end{align}
Moreover, if these smoothness, Gram matrix, and preconditioner bounds hold for all $0\leq t\leq T$, then
\begin{equation}
\label{eq:pinn-det-total-movement}
\sum_{t=0}^T
\|\theta^{(t+1)}-\theta^{(t)}\|
\leq32\sqrt{\frac{L_0}{\lambda_0}}.
\end{equation}
\end{lemma}

\begin{proof}
Since $\bar\eta\lambda_0\leq\gamma/8$, the stated properties of $q$ follow immediately. By the neuron-wise gradient bound in Lemma~\ref{lem:pinn-det-geometry} and the convex combination representation~\eqref{eq:det-app-v-history}, we have
\[
\widehat v_j^{(t)}
\leq\frac{C_g^2\ell_m S}{m}.
\]
Together with $\epsilon=\bar\epsilon/\sqrt m$, this establishes \eqref{eq:pinn-det-metric} for any $\beta_2\in[0,1)$. Applying the geometric-weight argument from Lemma~\ref{lem:det-app-moments}, with $C_g$ replaced by $C_g\sqrt{\ell_m}$, gives \eqref{eq:pinn-det-moment} and \eqref{eq:pinn-det-neuron-movement}. Since this argument uses only the gradients through iteration $T$ and $P^{(t)}\preceq I$, it also controls the candidate update at iteration $T+1$.

For the Lyapunov estimates, the proof of Lemma~\ref{lem:det-app-energy} requires smoothness only along the update segment and the Gram matrix lower bound only at its starting point. Under these conditions, we obtain
\[
\begin{aligned}
L(\theta^{(t+1)})-L(\theta^{(t)})
&\leq-\frac{\bar\eta}{2}Q^{(t)}
+2\bar\eta\|e^{(t)}\|^2,\\
\|e^{(t+1)}\|^2
&\leq\beta_1\|e^{(t)}\|^2
+\frac{2\beta_1^2H_{\rm P}^2\bar\eta^2}{\gamma}
\bigl(Q^{(t)}+\|e^{(t)}\|^2\bigr).
\end{aligned}
\]
The second bound follows from the exact bias-corrected first-moment recurrence; when $\beta_1=0$, the tracking error vanishes. Multiplying this inequality by $4\bar\eta/\gamma$ and combining it with the loss descent estimate establishes \eqref{eq:pinn-det-dissipation}, since $H_{\rm P}\bar\eta/\gamma\leq1/8$.

Moreover, the Gram matrix and preconditioner lower bounds imply
$Q^{(t)}\geq\lambda_0L(\theta^{(t)})/4$. Consequently,
\[
\begin{aligned}
\Delta^{(t)}
&\geq\frac{\bar\eta\lambda_0}{16}L(\theta^{(t)})
+\bar\eta\|e^{(t)}\|^2\\
&\geq\frac{\bar\eta\lambda_0}{32}\mathcal E^{(t)},
\end{aligned}
\]
where the last inequality uses $\bar\eta\lambda_0\leq\gamma/8$. Finally, the squared-step and telescoping arguments in Lemma~\ref{lem:det-app-energy} yield \eqref{eq:pinn-det-energy-step} and \eqref{eq:pinn-det-total-movement}, respectively, using $\mathcal E^{(0)}=L_0$ from exact bias correction.
\end{proof}

\subsection{Completion of the Proof}

\begin{proof}[Proof of Theorem~\ref{thm:pinn-det}]
We work on the initialization event established in Lemma~\ref{lem:pinn-det-init}. If $L_0=0$, all residuals and gradients vanish, and the Adam iterates remain at initialization. Thus, the conclusion holds trivially. Henceforth, assume that $L_0>0$.

Let $H_{\rm P}=C_Hd^2$, where $C_H\geq\max\{1,\alpha_{\rm BC}\}$ is a universal constant. In particular, $H_{\rm P}\geq\max\{1,\lambda_0\}$. The assumptions of Theorem~\ref{thm:pinn-det} ensure that
\begin{equation}
\label{eq:pinn-det-bootstrap}
\bar\epsilon\geq C_g\sqrt{\ell_m S_0},
\qquad
0<\bar\eta\leq\frac{\gamma}{8H_{\rm P}},
\qquad
m\geq C_{\rm w}\frac{\ell_m S_0}{\lambda_0^2},
\end{equation}
where $C_{\rm w}>0$ is a sufficiently large universal constant. Define
\[
q=1-\frac{\bar\eta\lambda_0}{32},
\qquad
\mathcal R=c_0\sqrt{\frac{m\lambda_0}{\ell_m}},
\]
as in Lemmas~\ref{lem:pinn-det-adam} and~\ref{lem:pinn-det-geometry}, respectively.

We prove by induction that, for every integer $t\geq0$,
\begin{equation}
\label{eq:pinn-det-induction}
\begin{gathered}
\mathcal E^{(t)}\leq q^tL_0,
\qquad
\max_{1\leq r\leq m}
\|\theta_r^{(t)}-\theta_r^{(0)}\|
\leq\frac12,\\
\|\theta^{(t)}-\theta^{(0)}\|
\leq\frac{\mathcal R}{2}.
\end{gathered}
\end{equation}
These assertions hold at $t=0$, since $\mathcal E^{(0)}=L_0$ and the parameter differences vanish. Suppose that they hold for all $0\leq t\leq T$. Then the iterates $\theta^{(0)},\ldots,\theta^{(T)}$ belong to $\mathcal S_{\rm P}$ and satisfy
\[
L(\theta^{(k)})\leq q^kL_0,
\qquad 0\leq k\leq T.
\]
By Lemma~\ref{lem:pinn-det-adam} and the damping condition in \eqref{eq:pinn-det-bootstrap}, we obtain
\[
P^{(T)}\succeq\frac12 I.
\]
Moreover, the Gram matrix lower bound in Lemma~\ref{lem:pinn-det-geometry} gives
\[
K(\theta^{(T)})\succeq\frac{\lambda_0}{4}I.
\]

We first verify the regularity conditions along the candidate update segment before applying the Lyapunov descent estimates. By \eqref{eq:pinn-det-neuron-movement} and \eqref{eq:pinn-det-bootstrap},
\[
\begin{aligned}
\max_r\|\theta_r^{(T+1)}-\theta_r^{(0)}\|
&\leq\sum_{t=0}^T\max_r
\|\theta_r^{(t+1)}-\theta_r^{(t)}\|\\
&\leq\frac{128C_g}{\lambda_0}
\sqrt{\frac{\ell_m L_0}{m}}
\leq\frac12.
\end{aligned}
\]
Hence, $\theta^{(T+1)}\in\mathcal S_{\rm P}$, and convexity ensures that the entire update segment lies in $\mathcal S_{\rm P}$.

The Jacobian bound in \eqref{eq:pinn-app-jacobian}, together with the induction hypothesis, gives
$\|g^{(k)}\|\leq Cd\sqrt{L_0}$ for all $k\leq T$. Using the convex combination representation~\eqref{eq:det-app-s-history} and $P^{(T)}\preceq I$, we obtain
\[
\|\theta^{(T+1)}-\theta^{(T)}\|
\leq\bar\eta\|\widehat s^{(T)}\|
\leq C\bar\eta d\sqrt{L_0}.
\]
Since $\|J(\theta)\|\leq Cd$ throughout the segment, the fundamental theorem of calculus gives
\[
\begin{aligned}
\sup_{\theta\in[\theta^{(T)},\theta^{(T+1)}]}
\|R(\theta)\|& \leq\sqrt{2L_0}
+Cd\|\theta^{(T+1)}-\theta^{(T)}\|\\
&\leq C(1+\bar\eta d^2)\sqrt{L_0}
\leq C\sqrt{L_0},
\end{aligned}
\]
where the last inequality follows from $\bar\eta\leq\gamma/(8H_{\rm P})$ and $H_{\rm P}=C_Hd^2$.

Consequently, the Hessian bound in \eqref{eq:pinn-app-regularity} implies
\[
\begin{aligned}
&\sup_{\theta\in[\theta^{(T)},\theta^{(T+1)}]}
\|\nabla^2L(\theta)\|\\
&\quad\leq Cd^2
+C\sqrt{\frac{\ell_m L_0}{m}}\\
&\quad\leq Cd^2+C\lambda_0
\leq H_{\rm P},
\end{aligned}
\]
where we used $L_0\leq S_0$, the width condition in \eqref{eq:pinn-det-bootstrap}, and $\lambda_0\leq\alpha_{\rm BC}$. Choosing $C_H$ sufficiently large therefore verifies the segment smoothness condition in Lemma~\ref{lem:pinn-det-adam}.

We can now apply Lemma~\ref{lem:pinn-det-adam} at iteration $T$ to obtain
\[
\mathcal E^{(T+1)}\leq q^{T+1}L_0.
\]
Moreover, the cumulative parameter movement bound gives
\[
\begin{aligned}
\|\theta^{(T+1)}-\theta^{(0)}\|
&\leq32\sqrt{\frac{L_0}{\lambda_0}}\\
&\leq\frac{c_0}{2}
\sqrt{\frac{m\lambda_0}{\ell_m}}
=\frac{\mathcal R}{2},
\end{aligned}
\]
where the second inequality follows from \eqref{eq:pinn-det-bootstrap} by choosing $C_{\rm w}$ sufficiently large. This completes the induction.

Since $L(\theta^{(t)})\leq\mathcal E^{(t)}$, we conclude that
\[
L(\theta^{(t)})
\leq\left(1-\frac{\bar\eta\lambda_0}{32}\right)^tL_0,
\]
establishing the claimed linear convergence rate with $c=1/32$. The Gram matrix lower bound holds uniformly along the trajectory by Lemma~\ref{lem:pinn-det-geometry}.

Furthermore, letting $T\to\infty$ in \eqref{eq:pinn-det-total-movement} shows that the parameter trajectory has finite total length and hence converges to a limit $\theta^\star$. By continuity of $L$ and the convergence $L(\theta^{(t)})\to0$, we have $L(\theta^\star)=0$.

Substituting $S_0=Cd^2\log(2/\delta)$ into \eqref{eq:pinn-det-bootstrap} recovers the width, damping, and stepsize conditions in Theorem~\ref{thm:pinn-det}, up to constants depending only on the fixed boundary penalty $\alpha_{\rm BC}$. The remaining width condition guarantees the initialization event in Lemma~\ref{lem:pinn-det-init}, which holds with probability at least $1-\delta$.

Finally, the finite-length bound~\eqref{eq:pinn-det-total-movement} on the trajectory implies that $\{\theta^{(t)}\}$ converges to a finite limit $\theta^{\star}$. By the continuity of $L$ and \eqref{eq:pinn-det-rate}, we have $L(\theta^{\star})=0$.
\end{proof}

\newpage

\section{Proof of Theorem~\ref{thm:pinn-stoch}}
\label{app:pinn-stoch-proof}

We consider the empirical PINN loss and the independent PDE and boundary minibatches defined in the main text, under the fixed-penalty convention of Section~\ref{sec:pinn}. As in the stochastic regression analysis, let
\[
\gamma=1-\beta_1,\qquad
\nu=1-\beta_2,\qquad
T_*=1+\left\lceil\frac{\log 2}{-\log\beta_2}\right\rceil.
\]
The proof adapts the stochastic regression argument by establishing global regularity for an auxiliary PINN, deriving the required minibatch moment bounds and conditional Lyapunov descent, and applying a stopping-time argument to show that the auxiliary and original Adam trajectories coincide.

Recall that $\ell_m$ is defined in~\eqref{eq_l_m}. Since $d\geq1$ and $C_*$ depends only on the fixed boundary penalty $\alpha_{\rm BC}$,
\[
\ell_m\asymp
d^2+\log^2\frac{md^2}{\delta\lambda_0}.
\]
For brevity, we use $\ell_m$ throughout the proof, while Theorem~\ref{thm:pinn-stoch} states the dependence on these parameters explicitly.

\subsection{An Auxiliary PINN with Global Residual Regularity}

We work on the initialization event of Lemma~\ref{lem:pinn-det-init}, with failure probability $\delta/2$. Replacing $\delta$ by $\delta/2$ affects only the constants in the initialization bounds. Throughout this appendix, constants may depend on the fixed boundary penalty $\alpha_{\rm BC}$ and the cutoff functions, but not on other problem or optimizer parameters.

We use the output-weight cutoff $\psi$ introduced in Appendix~\ref{app:reg-stoch-proof}. Unlike in the regression setting, we also introduce a cutoff for the hidden weights, since the derivatives of the PDE features can grow quadratically with the hidden-weight norm. Let $\chi_{\rm cut}:[0,\infty)\to[0,1]$ be a fixed smooth function satisfying
$\chi_{\rm cut}(s)=1$ for $s\leq1$ and
$\chi_{\rm cut}(s)=0$ for $s\geq4$.
Define the auxiliary network by
\begin{equation}
\label{eq:pinn-stoch-app-aux-network}
\begin{gathered}
w_r^\dagger
=w_r^{(0)}
+\chi_{\rm cut}(\|w_r-w_r^{(0)}\|^2)
(w_r-w_r^{(0)}),\\
u_\theta^\dagger(x)
=\frac1{\sqrt m}\sum_{r=1}^m
\psi(a_r)\sigma((w_r^\dagger)^\top\widetilde x).
\end{gathered}
\end{equation}
The corresponding unnormalized residuals are
\begin{equation}
\label{eq:residual}
\begin{aligned}
r_i^{{\rm PDE},\dagger}(\theta)
&=\Delta u_\theta^\dagger(x_i)-f(x_i),\\
r_j^{{\rm BC},\dagger}(\theta)
&=u_\theta^\dagger(z_j)-\varphi(z_j).
\end{aligned}
\end{equation}
The map defining $w_r^\dagger$, together with its first two derivatives, is uniformly bounded independently of $d$. We define $R^\dagger$, $L^\dagger$, $J^\dagger$, and $K^\dagger$ as in~\eqref{eq:pinn-residual}, using the same residual normalization and boundary penalty.

\begin{lemma}[Properties of the auxiliary PINN]
\label{lem:pinn-stoch-app-auxiliary}
On the initialization event of Lemma~\ref{lem:pinn-det-init}, the following properties hold.
\begin{enumerate}
\item The auxiliary and original residual maps, together with their Jacobians, coincide on the set $\mathcal S_{\rm P}$ defined in Lemma~\ref{lem:pinn-det-geometry}. In particular,
\[
L^\dagger(\theta)=L(\theta),
\qquad
K^\dagger(\theta)=K(\theta),
\qquad \theta\in\mathcal S_{\rm P}.
\]

\item For any parameter vectors $\theta,\vartheta$, the auxiliary residual map satisfies
\begin{equation}
\label{eq:pinn-stoch-app-aux-regularity}
\begin{gathered}
\|J^\dagger(\theta)\|\leq Cd,
\qquad
\|R^\dagger(\theta)-R^\dagger(\vartheta)\|
\leq Cd\|\theta-\vartheta\|,\\
\|J^\dagger(\theta)-J^\dagger(\vartheta)\|
\leq C\sqrt{\frac{\ell_m}{m}}\|\theta-\vartheta\|,\\
\|\nabla^2L^\dagger(\theta)\|
\leq Cd^2+
C\sqrt{\frac{\ell_m}{m}}\|R^\dagger(\theta)\|.
\end{gathered}
\end{equation}
Moreover, the unnormalized PDE and boundary residuals, defined in \eqref{eq:residual}, satisfy
\begin{equation}
\label{eq:pinn-stoch-app-row-bounds}
\begin{gathered}
\|\nabla r_i^{{\rm PDE},\dagger}(\theta)\|^2\leq Cd^2,
\qquad
\|\nabla r_j^{{\rm BC},\dagger}(\theta)\|^2\leq C,\\
\|\nabla_{\theta_r}r_i^{{\rm PDE},\dagger}(\theta)\|
\leq\frac{C(1+\|w_r^{(0)}\|^2)}{\sqrt m},
\qquad
\|\nabla_{\theta_r}r_j^{{\rm BC},\dagger}(\theta)\|
\leq\frac{C}{\sqrt m}.
\end{gathered}
\end{equation}

\item Let
\[
\mathcal R=c_0\sqrt{\frac{m\lambda_0}{\ell_m}},
\]
where $c_0>0$ is sufficiently small. Then every parameter vector $\theta$ satisfying
$\|\theta-\theta^{(0)}\|\leq\mathcal R$ obeys
\begin{equation}
\label{eq:pinn-stoch-app-aux-kernel}
K^\dagger(\theta)\succeq\frac{\lambda_0}{4}I,
\qquad
\|\nabla L^\dagger(\theta)\|^2
\geq\frac{\lambda_0}{2}L^\dagger(\theta).
\end{equation}
\end{enumerate}
\end{lemma}

\begin{proof}
When $\|w_r-w_r^{(0)}\|\leq1$, the hidden-weight cutoff map coincides with the identity, as does its derivative. Likewise, $\psi(a)=a$ and $\psi'(a)=1$ for $|a|\leq2$. Since both cutoffs are independent of the spatial variable, we have
\[
\Delta u_\theta^\dagger(x)
=\frac1{\sqrt m}\sum_{r=1}^m
\psi(a_r)\phi_{\rm P}(w_r^\dagger;x).
\]
Differentiating with respect to the network parameters shows that the residual maps and their Jacobians on $\mathcal S_{\rm P}$ coincide.

For arbitrary parameters, $\|w_r^\dagger-w_r^{(0)}\|\leq2$. By the feature derivative formulas in~\eqref{eq:pinn-app-feature-derivatives} and the chain rule, each PDE neuron contribution and its first two parameter derivatives are bounded by
$C(1+\|w_r^{(0)}\|^2)/\sqrt m$.
The corresponding boundary bounds are $C/\sqrt m$.
These estimates rely on dimension-independent bounds for the derivatives of the cutoff map. Indeed, for the map
$z\mapsto\chi_{\rm cut}(\|z\|^2)z$, the Jacobian is
\[
\chi_{\rm cut}(\|z\|^2)I
+2\chi_{\rm cut}'(\|z\|^2)zz^\top,
\]
whose nonconstant cutoff terms are supported on $\|z\|\leq2$. The second derivative is uniformly bounded for the same reason.

The block-wise arguments in the proof of Lemma~\ref{lem:pinn-det-geometry} now apply globally. The initialization bounds
\[
\frac1m\sum_{r=1}^m
(1+\|w_r^{(0)}\|^2)^2\leq Cd^2,
\qquad
\max_r(1+\|w_r^{(0)}\|^2)^2\leq C\ell_m
\]
yield the Jacobian bound and its Lipschitz estimate, respectively. The residual Lipschitz bound follows by integration. Moreover,
\[
\nabla^2L^\dagger
=(J^\dagger)^\top J^\dagger
+\sum_{i=1}^N R_i^\dagger\nabla^2R_i^\dagger.
\]
The second term is block diagonal across neurons, with each block bounded by
$C\sqrt{\ell_m/m}\|R^\dagger\|$, establishing the Hessian estimate. Summing the squared neuron-wise gradient bounds gives~\eqref{eq:pinn-stoch-app-row-bounds}.

Finally, since
$J^\dagger(\theta^{(0)})=J(\theta^{(0)})$,
the global Jacobian Lipschitz estimate allows us to apply the singular-value perturbation argument of Lemma~\ref{lem:pinn-det-geometry} throughout the ball
$\|\theta-\theta^{(0)}\|\leq\mathcal R$.
This proves~\eqref{eq:pinn-stoch-app-aux-kernel}.
\end{proof}

We apply Adam to the auxiliary loss $L^\dagger$ with the same initialization, optimizer parameters, and minibatch sequences as the original algorithm. The trajectory-agreement argument from Appendix~\ref{app:reg-stoch-proof} applies with $\mathcal S_{\rm P}$ (defined in Lemma~\ref{lem:pinn-det-geometry}) in place of $\mathcal S$. As long as the auxiliary iterates remain in $\mathcal S_{\rm P}$, the two algorithms have identical gradients, moment estimates, and parameter updates. For the remainder of the proof, we suppress the superscript $\dagger$ on all auxiliary quantities. In particular, $G^{(t)}=\nabla L(\theta^{(t)})$ denotes the full gradient of the auxiliary loss. We establish agreement with the original trajectory at the end of the proof.

\subsection{Key Lemmas}

Define the finite-population correction factors
\[
\chi_{\rm PDE}
=\frac{n_{\rm PDE}-b_{\rm PDE}}
{b_{\rm PDE}(n_{\rm PDE}-1)},
\qquad
\chi_{\rm BC}
=\frac{n_{\rm BC}-b_{\rm BC}}
{b_{\rm BC}(n_{\rm BC}-1)},
\]
with each factor defined to be zero when the corresponding population size is one. Recall that
\[
\rho_b
=1+\frac{1}{\lambda_0}
\max\{d^2\chi_{\rm PDE},
\alpha_{\rm BC}\chi_{\rm BC}\},
\]
and define
\[
\kappa
=\max\left\{
\frac{n_{\rm PDE}}{b_{\rm PDE}},
\frac{n_{\rm BC}}{b_{\rm BC}}
\right\}.
\]
As in the stochastic regression analysis, let $\mathcal F_t$ denote the sigma-algebra generated by the initialization and all PDE and boundary minibatches sampled before iteration $t$, and write
\[
\E_t[\cdot]=\E[\cdot\mid\mathcal F_t].
\]
The following lemma provides bounds valid for all auxiliary parameter vectors, with only the relative variance bound requiring the local Gram matrix lower bound.

\begin{lemma}[Two-group minibatch moment bounds]
\label{lem:pinn-stoch-app-sampling}
Define the sampled weighted squared residual by
\begin{equation}
\label{eq:pinn-stoch-app-scalar-input}
Y^{(t)}
=\frac1{b_{\rm PDE}}
\sum_{i\in B_{\rm PDE}^{(t)}}
\bigl(r_i^{\rm PDE}(\theta^{(t)})\bigr)^2
+\frac{\alpha_{\rm BC}}{b_{\rm BC}}
\sum_{j\in B_{\rm BC}^{(t)}}
\bigl(r_j^{\rm BC}(\theta^{(t)})\bigr)^2.
\end{equation}
Then $\kappa\leq C\rho_b$, and the stochastic gradient satisfies
\begin{equation}
\label{eq:pinn-stoch-app-sampling-bounds}
\begin{gathered}
\E_t[g^{(t)}]=G^{(t)},
\qquad
\|g^{(t)}\|^2\leq Cd^2Y^{(t)},
\qquad
\|g_r^{(t)}\|^2
\leq\frac{C\ell_m}{m}Y^{(t)},\\
\E_t\|g^{(t)}\|^2
\leq Cd^2L(\theta^{(t)}),
\qquad
\E_t\|g^{(t)}-G^{(t)}\|^2
\leq C\lambda_0(\rho_b-1)L(\theta^{(t)}).
\end{gathered}
\end{equation}
In particular, if
$K(\theta^{(t)})\succeq\lambda_0 I/4$, then
\[
\E_t\|g^{(t)}-G^{(t)}\|^2
\leq C(\rho_b-1)\|G^{(t)}\|^2.
\]

The sampled squared residual also satisfies
\begin{equation}
\label{eq:pinn-stoch-app-scalar-moments}
\E_t[Y^{(t)}]=2L(\theta^{(t)}),
\qquad
0\leq Y^{(t)}\leq2\kappa L(\theta^{(t)}),
\qquad
\operatorname{Var}_t(Y^{(t)})
\leq4\kappa L(\theta^{(t)})^2.
\end{equation}
Define its bias-corrected exponential moving average by
\[
U^{(t)}
=\frac{1-\beta_2}{1-\beta_2^{t+1}}
\sum_{s=0}^t\beta_2^{t-s}Y^{(s)}.
\]
Then the Adam second-moment estimates satisfy
\begin{equation}
\label{eq:pinn-stoch-app-coordinate-trace}
\widehat v_j^{(t)}
\leq \frac{C\ell_m}{m} U^{(t)}
\quad\text{for every }j,
\qquad
\sum_j\widehat v_j^{(t)}
\leq Cd^2U^{(t)}.
\end{equation}
\end{lemma}

\begin{proof}
The uniform-subset variance formula in
Lemma~\ref{lem:stoch-app-sampling} applies separately to the
PDE and boundary groups. At $\theta^{(t)}$, define
\[
G_i^{\rm PDE}
=r_i^{\rm PDE}\nabla r_i^{\rm PDE},
\qquad
G_j^{\rm BC}
=r_j^{\rm BC}\nabla r_j^{\rm BC},
\]
and let $\overline G^{\rm PDE}$ and $\overline G^{\rm BC}$
denote their respective population means.
By independence of the two minibatches,
\begin{align*}
\E_t\|g^{(t)}-G^{(t)}\|^2
&=\chi_{\rm PDE}\left(
\frac1{n_{\rm PDE}}\sum_i\|G_i^{\rm PDE}\|^2
-\|\overline G^{\rm PDE}\|^2\right)\\
&\quad+\alpha_{\rm BC}^2\chi_{\rm BC}\left(
\frac1{n_{\rm BC}}\sum_j\|G_j^{\rm BC}\|^2
-\|\overline G^{\rm BC}\|^2\right).
\end{align*}
The boundary contribution contains $\alpha_{\rm BC}^2$,
one factor of which is absorbed into the weighted loss.
Using the row-gradient bounds in
\eqref{eq:pinn-stoch-app-row-bounds}, we obtain
\[
\E_t\|g^{(t)}-G^{(t)}\|^2
\leq C\lambda_0(\rho_b-1)L(\theta^{(t)}).
\]
The relative variance bound then follows from the local
Gram matrix lower bound~\eqref{eq:pinn-stoch-app-aux-kernel}.

Applying Cauchy--Schwarz within each sampled group yields
the neuron-wise estimate
\[
\|g_r^{(t)}\|^2
\leq\frac{C}{m}
\left(1+\|w_r^{(0)}\|^2\right)^2Y^{(t)}.
\]
Taking the maximum over neurons or summing the squared
bounds gives the pointwise gradient estimates in
\eqref{eq:pinn-stoch-app-sampling-bounds}.
Taking conditional expectations gives the corresponding second-moment bound. Finally, averaging the coordinate-wise and summed gradient bounds with the nonnegative bias-correction weights establishes
\eqref{eq:pinn-stoch-app-coordinate-trace}.

For the scalar moment bounds, we apply the same finite-population variance formula to the nonnegative populations $(r_i^{\rm PDE})^2$ and $(r_j^{\rm BC})^2$. For any nonnegative population of size $n$ with mean $\mu$, its population variance is bounded by $(n-1)\mu^2$. Since $(n-1)\chi=n/b-1$, the variance of its uniformly sampled mean is at most $(n/b-1)\mu^2$. Applying this bound separately to the PDE and boundary groups, accounting for the boundary penalty and using independence, establishes the variance bound in~\eqref{eq:pinn-stoch-app-scalar-moments}. The expectation identity follows from unbiased sampling, while the almost-sure bound follows from nonnegativity.

% Finally, the diagonal estimates in~\eqref{eq:pinn-app-diagonals} imply
% \[
% \lambda_0\leq\frac{Cd^2}{n_{\rm PDE}},
% \qquad
% \lambda_0\leq\frac{\alpha_{\rm BC}}{n_{\rm BC}}.
% \]
% For each group with population size $n\geq 1$, the identity
% $n/b=1+(n-1)\chi$ therefore gives
% \[
% \frac{n}{b}\leq C\rho_b.
% \]
% For a singleton group, $n/b=1$. Taking the maximum over the two groups yields $\kappa\leq C\rho_b$, completing the proof.
\end{proof}

\paragraph{Uniform control of the scalar history.}
The proof of Lemma~\ref{lem:stoch-app-scalar-cap} uses only the three scalar moment bounds in~\eqref{eq:stoch-app-scalar-moments}. Since their two-group counterparts in~\eqref{eq:pinn-stoch-app-scalar-moments} have the same form, the lemma applies directly to the present $Y^{(t)}$, $U^{(t)}$, and $\kappa$, with the same predictable masking argument under the condition
\[
\nu\kappa\log\frac{C}{\Delta_q\delta_v}\leq c.
\]
No union bound over parameter coordinates is needed, since~\eqref{eq:pinn-stoch-app-coordinate-trace} controls all coordinatewise second moments and their sum through the single scalar process $U^{(t)}$.

\paragraph{Conditional Lyapunov descent.}
We continue to work with the auxiliary process, with $G^{(t)}=\nabla L(\theta^{(t)})$ and $P^{(t)}=\epsilon D^{(t)}$. We use the quantities $d^{(t)}$, $k^{(t)}$, and $Z^{(t)}$ from~\eqref{eq:stoch-app-virtual}, now defined using the PINN iterates. In particular,
\[
\begin{gathered}
Z^{(t)}
=\theta^{(t)}
-\bar\eta k^{(t)}P^{(t-1)}
\widehat s^{(t-1)},\\
\zeta=C_0\frac{\bar\eta\nu}{\gamma^2},
\qquad
\mathcal E^{(t)}
=L(Z^{(t)})+\zeta\|\widehat s^{(t-1)}\|^2.
\end{gathered}
\]
Here $C_0>0$ is a sufficiently large constant depending only on the fixed boundary penalty and cutoff functions. We also set $H_{\rm P}=C_Hd^2$, where $C_H>0$ is chosen sufficiently large to ensure that $H_{\rm P}\geq\max\{1,\lambda_0\}$.
The scalar history $U^{(t)}$ is defined in Lemma~\ref{lem:pinn-stoch-app-sampling}.

\begin{lemma}[Stochastic Lyapunov descent for PINNs]
\label{lem:pinn-stoch-app-energy}
Suppose that $t\geq T_*$ and
\[
K(\theta^{(t)})\succeq\frac{\lambda_0}{4}I,
\qquad
L(\theta^{(t)})\leq S,
\qquad
U^{(t-1)}\leq C_US.
\]
Assume further that
\[
m\geq\frac{C\ell_m S}{\lambda_0^2},
\qquad
\bar\epsilon\geq C\sqrt{\ell_m S},
\qquad
0<\nu\leq\frac{c\gamma}{\rho_b},
\qquad
0<\bar\eta H_{\rm P}\leq c\nu.
\]
Then the Lyapunov function satisfies
\begin{equation}
\label{eq:pinn-stoch-app-drift}
\begin{aligned}
\mathcal E^{(t)}-\E_t[\mathcal E^{(t+1)}]
&\geq\mathcal D^{(t)}\\
&:=c\bar\eta\|G^{(t)}\|^2
+c\gamma\zeta\|\widehat s^{(t-1)}\|^2\\
&\geq c\bar\eta\lambda_0\mathcal E^{(t)}.
\end{aligned}
\end{equation}
Moreover, let $\Delta Z^{(t)}=Z^{(t+1)}-Z^{(t)}$, then
\begin{equation}
\label{eq:pinn-stoch-app-speeds}
\begin{aligned}
\|\E_t[\Delta Z^{(t)}]\|^2
&\leq C\bar\eta\mathcal D^{(t)},\\
\E_t\|\Delta Z^{(t)}-\E_t[\Delta Z^{(t)}]\|^2
&\leq\frac{C\bar\eta^2(\rho_b-1)}{\gamma}
\|G^{(t)}\|^2.
\end{aligned}
\end{equation}
Finally, the following comparison bounds hold globally for the auxiliary process
\begin{equation}
\label{eq:pinn-stoch-app-comparison}
L(\theta^{(t)})\leq2\mathcal E^{(t)},
\qquad
\|\theta^{(t)}-Z^{(t)}\|^2
\leq\frac{C}{H_{\rm P}}\mathcal E^{(t)}.
\end{equation}
These bounds do not require the local Gram matrix or loss bounds.
\end{lemma}

\begin{proof}
We adapt the proof of Lemma~\ref{lem:stoch-app-energy} by establishing the required smoothness bounds at the scale $H_{\rm P}$. This step is necessary because the auxiliary PINN loss has a residual-dependent Hessian bound rather than a uniform bound of order $H_{\rm P}$.

Condition on $\mathcal F_t$ and write $h=\widehat s^{(t-1)}$. The historical momentum bound used in the proof of Lemma~\ref{lem:stoch-app-energy}, together with the trace bound in~\eqref{eq:pinn-stoch-app-coordinate-trace}, yields
\[
\|h\|^2
\leq\frac{C\gamma}{\nu}
\sum_j\widehat v_j^{(t-1)}
\leq\frac{C\gamma}{\nu}H_{\rm P}S.
\]
Since $k^{(t)}\leq\beta_1/\gamma$, it follows that
\[
\sqrt{H_{\rm P}}\|Z^{(t)}-\theta^{(t)}\|
\leq C\frac{\bar\eta H_{\rm P}}{\sqrt{\gamma\nu}}\sqrt S
\leq C\sqrt S.
\]

For any realization of the current minibatches, the exact virtual update~\eqref{eq:stoch-app-exact-update} satisfies
\[
\|\Delta Z^{(t)}\|
\leq\bar\eta\left(
2\|g^{(t)}\|+\frac{\beta_1}{\gamma}\|h\|
\right).
\]
Using $\|g^{(t)}\|^2\leq CH_{\rm P}Y^{(t)}$ and
$Y^{(t)}\leq2\kappa L(\theta^{(t)})$, we obtain
\[
\sqrt{H_{\rm P}}\|\Delta Z^{(t)}\|
\leq C\left(
\nu\sqrt{\kappa}+\sqrt{\frac{\nu}{\gamma}}
\right)\sqrt S
\leq C\sqrt S,
\]
where the last inequality uses $\kappa\lesssim\rho_b$ and
$\nu\rho_b\lesssim\gamma$.

By the global Lipschitz continuity of the auxiliary residual map, these estimates imply that $\|R\|\leq C\sqrt S$ throughout both segments
$[\theta^{(t)},Z^{(t)}]$ and
$[Z^{(t)},Z^{(t+1)}]$.
Consequently, the residual-dependent Hessian bound gives
\[
\sup\|\nabla^2L\|
\leq CH_{\rm P}
+C\sqrt{\frac{\ell_m S}{m}}
\leq CH_{\rm P}+C\lambda_0
\leq CH_{\rm P}
\]
on both segments, including any candidate update that exits the prescribed region. In particular, no Gram matrix lower bound is needed at the virtual iterates or at $\theta^{(t+1)}$.

To transfer the remaining Adam estimates, consider the rescaling
\[
\begin{gathered}
\widetilde L=\frac{L}{H_{\rm P}},
\qquad
\widetilde g=\frac{g}{H_{\rm P}},
\qquad
\widetilde s=\frac{s}{H_{\rm P}},
\qquad
\widetilde v=\frac{v}{H_{\rm P}^2},\\
\widetilde\epsilon=\frac{\epsilon}{H_{\rm P}},
\qquad
\widetilde{\bar\eta}=H_{\rm P}\bar\eta.
\end{gathered}
\]
The raw stepsize $\eta$ remains unchanged, and these substitutions preserve the parameter iterates, normalized preconditioners, and virtual iterates. The rescaled residual map $R/\sqrt{H_{\rm P}}$ is globally Lipschitz, while the corresponding loss has a uniformly bounded Hessian on the two segments established above. The Gram matrix lower bound becomes $\lambda_0/(4H_{\rm P})$, and the relative variance factor remains $\rho_b$.

Under this rescaling, the coordinatewise gradient bounds involve the factor $\ell_m/H_{\rm P}$. In the proof of Lemma~\ref{lem:stoch-app-energy}, these bounds enter only through the preconditioner lower bound and the normalization-bias estimate. Expressed in the original variables, the latter becomes
\[
\begin{aligned}
\|\E_t[e_t(g^{(t)})]\|^2
&\leq\frac{C\nu\rho_b\ell_m S}
{\gamma\bar\epsilon^2}
\|G^{(t)}\|^2\\
&\leq\frac{C\nu\rho_b}{\gamma}
\|G^{(t)}\|^2,
\end{aligned}
\]
where $e_t$ is defined in~\eqref{eq:stoch-app-predictable-map}. The parameter conditions therefore provide the same uniform preconditioner lower bound and bias control as in the regression analysis.

The remaining steps of Lemma~\ref{lem:stoch-app-energy} rely only on the relative moment bounds, exact bias-correction identities, and smoothness along the two segments. Applying those arguments under the above rescaling establishes \eqref{eq:pinn-stoch-app-drift} and \eqref{eq:pinn-stoch-app-speeds}. Indeed, the rescaled Lyapunov coefficient satisfies $\widetilde\zeta=H_{\rm P}\zeta$, so the corresponding energy is exactly $\mathcal E^{(t)}/H_{\rm P}$.

Finally, global residual Lipschitz continuity, together with
$\zeta\geq C H_{\rm P}\bar\eta^2(k^{(t)})^2$, yields the comparison bounds~\eqref{eq:pinn-stoch-app-comparison}, as in Lemma~\ref{lem:stoch-app-energy}.
\end{proof}

\subsection{Completion of the Proof}
\label{end:pinn-stoch-appendix}

\begin{proof}[Proof of Theorem~\ref{thm:pinn-stoch}]
Apply Lemma~\ref{lem:pinn-det-init} with failure probability $\delta/2$, and fix an initialization in the resulting event. Define
\[
L_0:=L(\theta^{(0)})\leq S_0:=Cd^2\log\frac{2}{\delta},
\qquad
S:=\frac{C_SS_0}{\delta},
\qquad
\mathcal R:=c_0\sqrt{\frac{m\lambda_0}{\ell_m}},
\]
where $C_S>0$ is a sufficiently large constant. Throughout the remainder of the proof, all probabilities and expectations are conditional on this initialization.

If $L_0=0$, the iterates remain at initialization, and the conclusion follows immediately. We therefore assume $L_0>0$. The width and damping conditions of Theorem~\ref{thm:pinn-stoch} imply that
\[
m\geq\frac{C\ell_m S}{\lambda_0^2},
\qquad
\bar\epsilon\geq C\sqrt{\ell_m S}.
\]
We first analyze the auxiliary Adam process and then establish its agreement with the original trajectory.

\paragraph{Parameter conditions and initial iterations.}
Let $q=1-c_E\bar\eta\lambda_0$, where $c_E>0$ is sufficiently small. Since $H_{\rm P}\asymp d^2$ and $\bar\eta H_{\rm P}\asymp\nu$, the prescribed upper bound on $\nu$ implies that
\[
\begin{gathered}
\nu\rho_b\lesssim\gamma,
\qquad
\bar\eta H_{\rm P}T_*\lesssim1,
\qquad
q^{-T_*}\lesssim1,\\
1-q\leq\frac12\min\{\nu,\gamma\},
\qquad
\nu\kappa\log\frac{C}{(1-q)\delta}\leq c.
\end{gathered}
\]
The last inequality follows from the monotonicity argument in the proof of Theorem~\ref{thm:reg-stoch}, with $\lambda_0$ replaced by $\lambda_0/H_{\rm P}$ and $\kappa\lesssim\rho_b$. Thus, the scalar concentration lemma applies to every admissible $\nu>0$, including values below the prescribed upper bound.

To control the initial iterations, we use the global Lipschitz continuity of the auxiliary residual map and the conditional second-moment estimate
\[
\E_t\|g^{(t)}\|^2
\leq CH_{\rm P}L(\theta^{(t)}).
\]
Let $M_t=\max_{0\leq s\leq t}\E L(\theta^{(s)})$. Since the bias-corrected first-moment weights form a convex combination,
\[
\E\|\widehat s^{(t)}\|^2
\leq CH_{\rm P}M_t,
\qquad
M_{t+1}^{1/2}
\leq(1+C\bar\eta H_{\rm P})M_t^{1/2}.
\]
Using these bounds and $\bar\eta H_{\rm P}T_*\lesssim1$, we obtain
\begin{equation}
\label{eq:pinn-stoch-app-startup}
\begin{gathered}
\E\max_{0\leq s\leq T_*}L(\theta^{(s)})
+\E\mathcal E^{(T_*)}
\leq CL_0,\\
\E\left(
\sum_{s=0}^{T_*-1}
\|\theta^{(s+1)}-\theta^{(s)}\|
\right)^2
\leq\frac{CL_0}{H_{\rm P}}.
\end{gathered}
\end{equation}
Define the startup event
\[
\Pi=
\left\{
\begin{gathered}
\max_{0\leq s\leq T_*}L(\theta^{(s)})\leq S,\quad
\sum_{s=0}^{T_*-1}
\|\theta^{(s+1)}-\theta^{(s)}\|
\leq\frac{\mathcal R}{4}
\end{gathered}
\right\}.
\]
By Markov's inequality,
\[
\mathbb P(\Pi^c)
\leq\frac{CL_0}{S}
+\frac{CL_0}{H_{\rm P}\mathcal R^2}.
\]

\paragraph{Stopping and restricted energy.}
On the startup event $\Pi$, define $\tau$ as the first $t\geq T_*$ at which any of the following conditions is violated:
\[
\mathcal E^{(t)}\leq Sq^{t-T_*},
\qquad
U^{(t-1)}\leq C_US,
\qquad
\|\theta^{(t)}-\theta^{(0)}\|\leq\frac{\mathcal R}{2}.
\]
Specifically, we define
\[
\tau=\inf\left\{t\ge T_*:
\begin{array}{l}
\mathcal E^{(t)}>Sq^{t-T_*},\ \text{or}\\
U^{(t-1)}>C_US,\ \text{or}\\
\|\theta^{(t)}-\theta^{(0)}\|>R/2
\end{array}
\right\}.
\]
Set $\tau=T_*$ on $\Pi^c$, and define the predictable indicators
\[
I^{(t)}=\mathbf1_\Pi\mathbf1_{\{t<\tau\}},
\qquad t\geq T_*.
\]
On each event \(\{I^{(t)}=1\}\),
\[
L(\theta^{(t)})
\leq2\mathcal E^{(t)}
\leq2Sq^{t-T_*},
\]
and the local Gram matrix lower bound holds. Thus,
Lemma~\ref{lem:pinn-stoch-app-energy}, applied with $2S$ in place of $S$, yields the stopped drift inequality.

We use the scalar-history masking argument from the stochastic regression proof. For $t<T_*$, we retain a sampled residual only if the losses up to iteration $t$ are bounded by $S$. For $t\geq T_*$, we use the indicators $I^{(t)}$. The resulting predictable mask satisfies the loss bound $2Sq^{(t-T_*)_+}$. On $\Pi$, all terms before $\tau$ are retained in the masked moving average. Thus, Lemma~\ref{lem:stoch-app-scalar-cap} bounds the probability of violating the second-moment condition by $\delta_v$, where $\delta_v$ is a sufficiently small fixed fraction of $\delta$.

Moreover, the stopped process
\[
\mathbf1_\Pi
q^{-((t\wedge\tau)-T_*)}
\mathcal E^{(t\wedge\tau)}
\]
is a nonnegative supermartingale. Applying Lemma~\ref{lem:aux-doob} together with \eqref{eq:pinn-stoch-app-startup} bounds the probability of an energy exit by $CL_0/S$.

Finally, define the restricted expected energy
\[
A_t=\E[I^{(t)}\mathcal E^{(t)}].
\]
The conditional drift inequality and the monotonicity of $I^{(t)}$ imply
\begin{equation}
\label{eq:pinn-stoch-app-killed-energy}
\begin{aligned}
A_t-A_{t+1}
&\geq\E[I^{(t)}\mathcal D^{(t)}]\\
&\geq(1-q)A_t,
\qquad
A_t\leq Cq^{t-T_*}L_0.
\end{aligned}
\end{equation}
This argument uses only predictable indicators and does not condition on future survival.

\paragraph{Control of parameter movement.}
Decompose each virtual increment as
\[
\Delta Z^{(t)}=W^{(t)}+\xi^{(t+1)},
\qquad
W^{(t)}=\E_t[\Delta Z^{(t)}],
\]
where $\xi^{(t+1)}$ is a martingale difference.
Combining the increment bounds in~\eqref{eq:pinn-stoch-app-speeds} with~\eqref{eq:pinn-stoch-app-killed-energy}, and applying the telescoping and Minkowski arguments from the stochastic regression proof, we obtain
\[
\E\left(
\sum_{t\geq T_*}I^{(t)}\|W^{(t)}\|
\right)^2
\leq\frac{CL_0}{\lambda_0},
\]
and
\[
\begin{aligned}
\sum_{t\geq T_*}
\E[I^{(t)}\|\xi^{(t+1)}\|^2]
&\leq
\frac{C\bar\eta(\rho_b-1)}{\gamma}L_0\\
&\leq\frac{CL_0}{H_{\rm P}}.
\end{aligned}
\]
Combining these estimates with the startup bound
\eqref{eq:pinn-stoch-app-startup} and the martingale maximal
inequality gives
\begin{equation}
\label{eq:pinn-stoch-app-stopped-path}
\E\left[
\mathbf1_\Pi\sup_{t\geq T_*}
\|Z^{(t\wedge\tau)}-\theta^{(0)}\|^2
\right]
\leq\frac{CL_0}{\lambda_0}.
\end{equation}
Here we use $\lambda_0\leq H_{\rm P}$, and the stopped sums include the increment that causes an exit.

If a radius exit occurs without an energy exit, the comparison bound~\eqref{eq:pinn-stoch-app-comparison} gives
\[
\|\theta^{(\tau)}-Z^{(\tau)}\|^2
\leq\frac{CS}{H_{\rm P}}
\leq\frac{\mathcal R^2}{16},
\]
where the last inequality follows from the width condition.
Consequently, the probability of such an exit is bounded by
$CL_0/(\lambda_0\mathcal R^2)$.

Define the good sampling event
$\mathcal G_0=\Pi\cap\{\tau=\infty\}$.
Combining the bounds for startup failure, energy exit, second-moment exit, and radius exit,
\[
\begin{aligned}
\mathbb P(\mathcal G_0^c)
&\leq
\frac{CL_0}{S}
+\frac{CL_0}{H_{\rm P}\mathcal R^2}
+\frac{CL_0}{\lambda_0\mathcal R^2}
+\delta_v\\
&\leq\frac{\delta}{4},
\end{aligned}
\]
provided that the constants in $S$ and the width condition are chosen sufficiently large.

Since $\mathcal G_0\subseteq\{I^{(t)}=1\}$ for every $t\geq T_*$, the restricted energy estimate \eqref{eq:pinn-stoch-app-killed-energy} and the global loss
comparison imply
\[
\E\left[
L(\theta^{(t)})\mathbf1_{\mathcal G_0}
\right]
\leq Cq^tL_0,
\qquad t\geq T_*.
\]
For $t<T_*$, the same estimate follows from the startup bound and $q^{-T_*}\lesssim1$.

\paragraph{Agreement with the original PINN.}
To establish uniform neuron-wise containment, define the second mask by
\[
\widehat I^{(s)}
=
\begin{cases}
1, & s<T_*,\\
I^{(s)}, & s\geq T_*.
\end{cases}
\]
Then $\E[\widehat I^{(s)}L(\theta^{(s)})]\leq CL_0q^s$.
Define the scalar convolution
\[
\mathcal T
=\bar\eta\sum_{t\geq0}\sum_{s=0}^t
\frac{\gamma\beta_1^{t-s}}{1-\beta_1^{t+1}}
\widehat I^{(s)}\sqrt{Y^{(s)}}.
\]
Using the exact bias-corrected moment weights,
$\E_s[Y^{(s)}]=2L(\theta^{(s)})$,
and $1-q\lesssim\gamma$, the argument from the stochastic regression proof gives
\[
\E[\mathcal T^2]\leq\frac{CL_0}{\lambda_0^2},
\qquad
\sum_{t\geq0}\max_r
\|\theta_r^{(t+1)}-\theta_r^{(t)}\|
\leq C\sqrt{\frac{\ell_m}{m}}\,\mathcal T
\quad\text{on }\mathcal G_0.
\]

Now define
\[
\mathcal G
=\mathcal G_0\cap
\left\{
\mathcal T\leq
C_T\sqrt{\frac{L_0}{\delta\lambda_0^2}}
\right\}.
\]
By Markov's inequality, choosing $C_T$ sufficiently large ensures
$\mathbb P(\mathcal G)\geq1-\delta/2$.
On this event, the total neuron-wise parameter movement satisfies
\begin{equation}
\label{eq4}
    \sum_{t\geq0}\max_r
\|\theta_r^{(t+1)}-\theta_r^{(t)}\|
\leq
C\sqrt{\frac{\ell_mL_0}{m\delta\lambda_0^2}}
\leq\frac12,
\end{equation}
where the last inequality follows from the width condition.
Consequently, for every neuron $r$ and iteration $t$,
\[
|a_r^{(t)}|\leq\frac32,
\qquad
\|w_r^{(t)}-w_r^{(0)}\|\leq\frac12.
\]
Thus, the auxiliary trajectory remains in $\mathcal S_{\rm P}$, where the auxiliary and original PINN residuals and their parameter derivatives coincide. Since both algorithms start from the same parameters and zero moments and use identical minibatches, their gradients, moment estimates, and iterates coincide by induction. The restricted expected-loss bound therefore transfers to the original Adam iterates on $\mathcal G$, establishing~\eqref{eq:pinn-stoch-rate}.

Finally, substituting
$S_0=Cd^2\log(2/\delta)$
recovers the width and damping requirements in Theorem~\ref{thm:pinn-stoch}, while the remaining width term ensures the initialization event. Combining the initialization probability of at least $1-\delta/2$ with the conditional sampling probability of at least $1-\delta/2$ gives a joint success probability of at least $1-\delta$.

Moreover, \eqref{eq4} implies that, almost surely on $\mathcal G$,
\[
\sum_{t=0}^{\infty}
\max_{1\leq r\leq m}
\left(
|a_r^{(t+1)}-a_r^{(t)}|
+\|w_r^{(t+1)}-w_r^{(t)}\|
\right)<\infty.
\]
Since $m$ is finite, the complete parameter sequence converges to a finite limit $\theta^{\star}$ on $\mathcal G$.
Moreover, by Fatou's lemma, continuity of $L$, and the restricted expected-loss bound, we have
\[
\begin{aligned}
\mathbb E\left[
L(\theta^{\star})\bm{1}_{\mathcal G}
\mid\theta^{(0)}
\right]
&\leq
\liminf_{t\to\infty}
\mathbb E\left[
L(\theta^{(t)})\bm{1}_{\mathcal G}
\mid\theta^{(0)}
\right]
=0.
\end{aligned}
\]
Thus, $L(\theta^{\star})=0$ almost surely on $\mathcal G$, which completes the proof.

\end{proof}

\newpage

\section{Convergence Analysis of GD and SGD for PINNs}
\label{app:pinn-gd-sgd}
The following proposition establishes the GD and SGD convergence bounds reported in Table~\ref{tab:pinn-optimizer-comparison} under the assumptions of Section~\ref{sec:pinn}.

\begin{proposition}[GD and SGD for the empirical PINN loss]
\label{prop:pinn-gd-sgd}
Let $\delta\in(0,1)$ and consider the update
\[
\theta^{(t+1)}=\theta^{(t)}-h g^{(t)}.
\]
\begin{enumerate}[label=(\roman*)]
\item For GD, let $g^{(t)}=\nabla L(\theta^{(t)})$ and
$0<h\lesssim d^{-2}$. Under the width condition
\eqref{eq:pinn-det-width}, with probability at least
$1-\delta$ over initialization,
\[
L(\theta^{(t)})\leq(1-ch\lambda_0)^tL(\theta^{(0)}),
\qquad t\geq0.
\]

\item For SGD, let $g^{(t)}$ be the minibatch gradient defined in~\eqref{eq:pinn-minibatch-gradients}, and suppose that
$0<h\lesssim(d^2\rho_b)^{-1}$. Under the width condition
\eqref{eq:pinn-stoch-width}, with probability at least
$1-\delta/2$ over initialization, there exists a sampling event
$\mathcal G$ such that
\[
\begin{gathered}
\mathbb P(\mathcal G\mid\theta^{(0)})\geq1-\delta/2,\\
\E\!\left[L(\theta^{(t)})\mathbf1_{\mathcal G}
\mid\theta^{(0)}\right]
\leq(1-ch\lambda_0)^tL(\theta^{(0)}),
\qquad t\geq0.
\end{gathered}
\]
The joint success probability is at least $1-\delta$.
\end{enumerate}
Consequently, choosing $h\asymp d^{-2}$ for GD and $h\asymp(d^2\rho_b)^{-1}$ for SGD gives contraction gaps of order $\lambda_0/d^2$ and $\lambda_0/(d^2\rho_b)$, respectively.
\end{proposition}

\begin{proof}
Apply Lemma~\ref{lem:pinn-det-init} with failure probability $\delta$ for GD and $\delta/2$ for SGD. Fix an initialization in the corresponding event, and let
$L_0=L(\theta^{(0)})\leq S_0:=Cd^2\log(2/\delta)$ and
$\mathcal R=c_0\sqrt{m\lambda_0/\ell_m}$.
We use the auxiliary PINN from Lemma~\ref{lem:pinn-stoch-app-auxiliary} and suppress the superscript $\dagger$ throughout the proof. All probabilities and expectations below are conditional on the fixed initialization. We assume $L_0>0$, as the claim is immediate otherwise. The global auxiliary bounds give
\[
\|R(\theta^{(t+1)})\|
\leq(1+Chd^2\sqrt\kappa)\|R(\theta^{(t)})\|
\]
for every minibatch realization. 

\paragraph{One-step descent.}
Let $H_{\rm P}=C_Hd^2\geq\lambda_0$, as in Appendix~\ref{app:pinn-stoch-proof}. Suppose that $L(\theta^{(t)})\leq S$, $m\gtrsim\ell_m S/\lambda_0^2$, and $\theta^{(t)}$ lies in the kernel neighborhood. The auxiliary regularity and sampling lemmas imply that, for every minibatch realization,
\[
\sup_{\vartheta\in[\theta^{(t)},\,\theta^{(t)}-h g^{(t)}]}
\|R(\vartheta)\|
\lesssim(1+hH_{\rm P}\sqrt\kappa)\sqrt S
\lesssim\sqrt S,
\]
where the last inequality uses $\kappa\lesssim\rho_b$ and $hH_{\rm P}\rho_b\lesssim1$.
Hence,~\eqref{eq:pinn-stoch-app-aux-regularity} gives a Hessian bound of $CH_{\rm P}$ along the entire update segment, even if the candidate update leaves the kernel neighborhood. Writing $G^{(t)}=\nabla L(\theta^{(t)})$, unbiasedness and Lemma~\ref{lem:pinn-stoch-app-sampling} yield
\begin{equation}
\label{eq:pinn-gd-sgd-descent}
\begin{aligned}
\E_t L(\theta^{(t+1)})
&\leq L(\theta^{(t)})-h\|G^{(t)}\|^2
       +CH_{\rm P}h^2\rho_b\|G^{(t)}\|^2\\
&\leq L(\theta^{(t)})-\frac h2\|G^{(t)}\|^2\\
&\leq qL(\theta^{(t)}),
\qquad q:=1-ch\lambda_0\in(0,1).
\end{aligned}
\end{equation}
For GD, we use full batches, so $\rho_b=\kappa=1$ and no conditional expectation is needed.

We first consider GD. Set $S=L_0$. Applying the descent and square-root telescoping arguments from the proof of Lemma~\ref{lem:pinn-det-adam} to the update $-hG^{(t)}$, we obtain
\[
\sum_t h\|G^{(t)}\|\lesssim\sqrt{\frac{L_0}{\lambda_0}},
\qquad
\sum_t\max_r\|\theta_r^{(t+1)}-\theta_r^{(t)}\|
\lesssim\frac1{\lambda_0}\sqrt{\frac{\ell_mL_0}{m}}.
\]
These bounds hold for every finite trajectory segment up to and including a candidate exit step. The second bound follows from the block-gradient estimate in Lemma~\ref{lem:pinn-stoch-app-sampling}, with full batches, and the geometric decay of $\sqrt{L(\theta^{(t)})}$.
The width condition~\eqref{eq:pinn-det-width} ensures that the two bounds are at most $\mathcal R/2$ and $1/2$, respectively. Thus, by induction, the iterates remain in the Gram matrix neighborhood and in $\mathcal S_{\rm P}$. The auxiliary and original GD trajectories therefore coincide, proving (i).

We next consider SGD. Set $S=C_SS_0/\delta$ and define
\[
\tau=\inf\left\{t\geq0:
L(\theta^{(t)})>Sq^t\ \text{or}\
\|\theta^{(t)}-\theta^{(0)}\|>\mathcal R/2\right\},
\qquad
I^{(t)}=\mathbf1_{\{t<\tau\}}.
\]
The width condition ensures that
$m\gtrsim\ell_m S/\lambda_0^2$.
By~\eqref{eq:pinn-gd-sgd-descent},
$q^{-(t\wedge\tau)}L(\theta^{(t\wedge\tau)})$
is a nonnegative supermartingale.
Define
$A_t=\E[I^{(t)}L(\theta^{(t)})]$ and
$\Delta_t=A_t-A_{t+1}$. Then
\[
A_t\leq q^tL_0,\qquad
\Delta_t\geq\frac h2\E[I^{(t)}\|G^{(t)}\|^2],
\qquad
\Delta_t\geq(1-q)A_t.
\]
As in Appendix~\ref{app:pinn-stoch-proof}, we decompose the update into its predictable and martingale increments,
$-hG^{(t)}$ and
$-h(g^{(t)}-G^{(t)})$, respectively.
Here, no startup period is needed.
The key telescoping estimate is
\[
\sqrt{\Delta_t}\leq\frac{2}{\sqrt{1-q}}
\bigl(\sqrt{A_t}-\sqrt{A_{t+1}}\bigr).
\]
Applying Minkowski's inequality and the martingale maximal inequality in Lemma~\ref{lem:aux-doob}, first over finite horizons and then passing to the limit by monotone convergence, we obtain
\[
\begin{gathered}
\E\!\left[\left(\sum_{t\geq0}I^{(t)}h\|G^{(t)}\|\right)^2\right]
\lesssim\frac{hL_0}{1-q}\lesssim\frac{L_0}{\lambda_0},\\
\sum_{t\geq0}\E\!\left[I^{(t)}h^2\|g^{(t)}-G^{(t)}\|^2\right]
\lesssim h(\rho_b-1)L_0\lesssim\frac{L_0}{H_{\rm P}},\\
\E\sup_{t\geq0}\|\theta^{(t\wedge\tau)}-\theta^{(0)}\|^2
\lesssim\frac{L_0}{\lambda_0}.
\end{gathered}
\]
The stopped sums include the increment that causes an exit. The maximal inequality for the loss supermartingale and Markov's inequality for the parameter movement therefore give
\[
\mathbb P(\tau<\infty)
\lesssim\frac{L_0}{S}+\frac{L_0}{\lambda_0\mathcal R^2}
\leq\frac\delta4.
\]

To establish neuron-wise containment, define
$\mathcal T=h\sum_{t\geq0}I^{(t)}\sqrt{Y^{(t)}}$.
By~\eqref{eq:pinn-stoch-app-scalar-moments}
and Minkowski's inequality,
\[
\E[\mathcal T^2]
\leq2h^2\left(\sum_{t\geq0}\sqrt{A_t}\right)^2
\lesssim\frac{L_0}{\lambda_0^2}.
\]
On $\{\tau=\infty\}$, the total neuron-wise parameter movement is bounded by
$C\sqrt{\ell_m/m}\,\mathcal T$.
Define
\[
\mathcal G
=
\{\tau=\infty\}
\cap
\left\{
\mathcal T\leq
C_T\sqrt{\frac{L_0}{\delta\lambda_0^2}}
\right\}.
\]
By Markov's inequality, choosing $C_T$ sufficiently large gives
$\mathbb P(\mathcal G)\geq1-\delta/2$.
On $\mathcal G$, the width condition ensures that the total neuron-wise movement is at most $1/2$.
Thus, the auxiliary iterates remain in $\mathcal S_{\rm P}$, and the original and auxiliary SGD trajectories coincide.
Since $I^{(t)}=1$ on $\mathcal G$ for every $t\geq0$, the bound $A_t\leq q^tL_0$ establishes (ii).
\end{proof}

\end{document}